%% file: main.tex
\newif\ifFORM
\FORMfalse

\newif\iftr
\newif\ifcnf
\newif\ifconf

\trfalse
\cnftrue
\conftrue

\trtrue
\cnffalse
\conffalse
 
\newif\ifnohl
\nohlfalse

\ifcnf
\documentclass[10pt,journal,compsoc]{IEEEtran}
\else
\documentclass[10pt,journal,compsoc]{IEEEtran_custom}
\fi

\IEEEoverridecommandlockouts

\include{mac-includes}

\include{mac-math}

\usepackage{cite}

\usepackage{fontawesome}
\usepackage{pifont}
\usepackage{textcomp}

\usepackage{xcolor}
\usepackage{soul}
\usepackage{dblfloatfix}

\usepackage{colortbl}
\usepackage[hidelinks]{hyperref}

\usepackage{xurl} 

\renewcommand{\comment}[1]{\ignorespaces}

\usepackage{physics}
\usepackage{tabularx}
\usepackage{amsmath}
\usepackage{mathtools} 

\newcommand{\leonard}[1]{\textcolor{purple}{[Leonard: #1]}}

\nohltrue

\ifnohl

\renewcommand{\rowcolor}[1]{}
\renewcommand{\marginpar}[1]{}

\fi

\newcolumntype{y}{>{}l}

\IEEEaftertitletext{\vspace{-2\baselineskip}}

\newif\ifHL
\HLfalse

\newcommand{\faY}[0]{\faBatteryFull}
\newcommand{\faH}[0]{\faBatteryHalf}
\newcommand{\faN}[0]{\faTimes}

\usepackage{moresize}

\usepackage{longtable}
\usepackage{pdflscape}
\usepackage{array}
\newcolumntype{P}[1]{>{\raggedright\arraybackslash}p{#1}}

\newcommand{\noAnswer}{\textcolor{lightgray}{\faQuestionCircle}}

\def\genbox#1#2#3#4#5#6{
    \leavevmode\raise#4bp\hbox to#5bp{\vrule height#5bp depth0bp width0bp
    \pdfliteral{q .5 w \csname #2COLOR\endcsname\space RG
                       \csname #3PDF\endcsname{#5}{#6} S Q
             \ifx1#1 q \csname #2COLOR\endcsname\space rg 
                       \csname #3PDF\endcsname{#5}{#6} f Q\fi}\hss}}

\def\trianbox   #1#2{\genbox{#1}{#2}  {trian}    {0.5}   {6}    {3}}

\begin{document}

\title{Performance Foundations of Parallel \& Distributed Reasoning Language Models}

\author{Maciej Besta$^{\dagger *}$, Leonard Schmidt$^{*}$, Lara Nonino, Robert Gerstenberger, Pierre Pang, \\ Patrik Okanovic, Ales Kubicek, Tiancheng Chen, Baraq Lipshitz, Torsten Hoefler\\
\vspace{0.5em}{ETH Zurich \quad {\small $^\dagger$\textit{Corresponding author} \quad $^*$\textit{Core contributions}}}}

\IEEEtitleabstractindextext{%
\begin{abstract}
Reinforcement Learning with Verifiable Rewards (RLVR) and other RL-style post-training paradigms have been used for aligning large language models (LLMs) with reasoning standards. The resulting recent \emph{Reasoning Language Models} (RLMs) such as DeepSeek-R1, o3, and Kimi k1.5 show that such RL-style post-training (``RL-for-LLMs'') can substantially improve chain-of-thought reasoning, long-horizon planning, and self-correction. However, the computational footprint of these systems is massive: state-of-the-art RLM training requires millions of GPU-hours and tightly coupled multi-model pipelines that stress modern hardware far beyond classical supervised LLM training. This makes RLM training as much a parallel and distributed systems problem as an algorithmic one. In this work, to facilitate developing RLMs that are simultaneously high-performance, scalable, and cost-effective, we first systematize the RL-for-LLM paradigm and provide a compute-centric analysis of prominent post-training algorithmic frameworks: Proximal Policy Optimization (PPO), Group Relative Policy Optimization (GRPO), as well as their variants. Second, we develop a taxonomy of \textit{intra-} and \textit{inter-model} parallelism strategies for RL-for-LLMs, covering both traditional techniques (data, tensor, pipeline, sequence, context, and expert parallelism) as well as novel forms of parallelism and optimization techniques for multi-model RLM training, for example disaggregated placement, stage fusion, hybrid parallelism, and asynchronous execution. We harness the work--depth model of parallel computing to make our taxonomy and its insights rigorous and portable. Finally, we analyze existing RLM frameworks and we distill practical guidelines and outline open research directions for building scalable, fast, and cost-effective RLMs.
\end{abstract}

\iftr
\fi
}

\maketitle
\IEEEdisplaynontitleabstractindextext
\IEEEpeerreviewmaketitle

\iftr
\else
{\vspace{-1.0em}\noindent \textbf{An extended version: \url{??}}\vspace{1em}}
\fi

\input{intro}
\input{background}
\input{intra}
\input{inter}
\iftr\input{appendix-schemes}\fi
\input{designs}

\input{opportunities}

\input{conclusion}

\section*{Acknowledgements}

{\small We thank Hussein Harake, Colin McMurtrie, Mark Klein, Angelo Mangili, and the whole CSCS team granting access to the Ault and Alps machines, and for their excellent technical support.
We gratefully acknowledge Polish high-performance computing infrastructure PLGrid (HPC Center: ACK Cyfronet AGH) for providing compute facilities within computational grant no.~PLG/2026/019437; we also thank Łukasz Flis and the whole Cyfronet team for their excellent technical support.
We thank Timo Schneider for help with infrastructure at SPCL.
This project received funding from the European Research Council (Project PSAP, No.~101002047), and the European High-Performance Computing Joint Undertaking (JU) under grant agreement No.~955513 (MAELSTROM). This project was supported by the ETH Future Computing Laboratory (EFCL), financed by a donation from Huawei Technologies. 
%
%
We acknowledge the Swiss AI Initiative for the computational grant.
A language model served as an editorial tool for this manuscript, while ideas and content are original work of the authors.}


\bibliographystyle{IEEEtranS}
\bibliography{references.complete}

\appendices

\input{appendix-pipeline}
\input{appendix-derivations}

\ifcnf

\vspace{-3.5em}
\begin{IEEEbiographynophoto}{\ssmall Maciej Besta}
\ssmall
is a researcher at ETH Zurich. He works on understanding and accelerating
large-scale irregular computations, such as graph streaming, graph neural
networks, or graph databases, at all levels of the computing stack.
\end{IEEEbiographynophoto}
\vspace{-4em}
\begin{IEEEbiographynophoto}{\ssmall Torsten Hoefler}
\ssmall
is a Professor at ETH Zurich, where he leads the Scalable Parallel Computing
Lab. His research aims at understanding performance of parallel computing
systems ranging from parallel computer architecture through parallel programming
to parallel algorithms.
\end{IEEEbiographynophoto}

\fi

\end{document}

%% file: mac-includes.tex
\newif\ifsq     

\newif\ifsqCAP
\newif\ifsqVS
\newif\ifsqEN
\newif\ifsqTIT

\newcommand{\ignore}[1]{}

\sqtrue
\sqCAPfalse
\sqENtrue
\sqVStrue
\sqTITtrue

\sqfalse
\sqCAPfalse
\sqENfalse
\sqVSfalse
\sqTITfalse

\usepackage{etex}
\usepackage{balance}
\usepackage{epstopdf}
\usepackage{placeins}
\usepackage{comment}

\usepackage{graphicx}
\usepackage{float}
\usepackage{multirow}
\usepackage{rotating}
\usepackage{makecell}
\usepackage{tabulary}
\usepackage{parcolumns}
\usepackage{tikz}
\usetikzlibrary{tikzmark}

\usepackage{xpatch}
\expandafter\xpatchcmd
\csname pgfk@/tikz/every picture/.@cmd\endcsname
{\thepage}{\arabic{page}}{}{}

\tikzstyle{comment} = [draw, fill=blue!70, text=white, text width=3cm, minimum height=1cm, rounded corners, align=left, font=\scriptsize]
\tikzstyle{background_alg} = [draw, fill=blue!20, opacity=0.4, inner sep=4pt, rounded corners=2pt]

\usetikzlibrary{shapes}
\usetikzlibrary{plotmarks}
\usetikzlibrary{calc, fit}

\usepackage{enumitem}

\usepackage{amsthm}
\usepackage{amsmath,amssymb,amsfonts}
\usepackage{mathtools,mathrsfs}

\usepackage{soul}
\usepackage{fontawesome}
\usepackage{pifont}
\usepackage{textcomp}
\usepackage{booktabs}
\usepackage{url}
\usepackage{pbox}
\usepackage[normalem]{ulem}
\usepackage[10pt]{moresize}

\ifsqCAP
\usepackage[font={normalfont, scriptsize}]{caption}
\usepackage[font={normalfont, scriptsize}]{subcaption}
\else
\usepackage[font={normalfont, small}]{caption}
\usepackage[font={normalfont, small}]{subcaption}
\fi

\newcommand{\vspaceSQ}[1]{\ifsqVS\vspace{#1}\fi}
\newcommand{\enlargeSQ}[1]{\ifsqEN\enlargethispage{\baselineskip}\fi}

\ifsqTIT
\usepackage[compact]{titlesec}
\titlespacing*{\section}{0pt}{3pt}{-1pt}
\titlespacing*{\subsection}{0pt}{0pt}{-3pt}
\titlespacing*{\subsubsection}{0pt}{2pt}{1pt}
\fi

\usepackage{xcolor}
\definecolor{darkgrey}{RGB}{70,70,70}
\definecolor{lightgrey}{RGB}{200,200,200}
\definecolor{lyellow}{RGB}{255,255,100}
\definecolor{llyellow}{RGB}{250,250,180}
\definecolor{lgreen}{RGB}{144,238,144}
\definecolor{raphael_comments}{RGB}{13, 145, 24}

\usepackage[customcolors]{hf-tikz}
\hfsetbordercolor{white}
\hfsetfillcolor{vlgray}

\definecolor{vlgray}{rgb}{0.77 0.77 0.77}
\definecolor{ablack}{rgb}{0.2 0.2 0.2}
\definecolor{vllgray}{rgb}{0.9 0.9 0.9}
\definecolor{bblue}{rgb}{0.7 0.7 0.99}

\usepackage{colortbl}

\usepackage{inconsolata}
\usepackage{listings}

\ifsq
\else
\fi

\newcommand{\maciej}[1]{\textcolor{blue}{[Maciej: #1]}}

\definecolor{hlL}{rgb}{0.8 0.8 0.99}

\newcounter{highlight}

\newcounter{hlLR}

\newcounter{hlLIR}

\newcounter{hlLIIR}

\newcounter{Ahighlight}

\usepackage{scalerel,stackengine}
\stackMath
\newcommand\rwh[1]{%
\savestack{\tmpbox}{\stretchto{%
  \scaleto{%
        \scalerel*[\widthof{\ensuremath{#1}}]{\kern-.6pt\bigwedge\kern-.6pt}%
                  {\rule[-\textheight/2]{1ex}{\textheight}}
                              }{\textheight}%
}{0.5ex}}%
\stackon[1pt]{#1}{\tmpbox}%
}

\usepackage[hang,flushmargin]{footmisc}

\renewcommand{\epsilon}{\ensuremath\varepsilon}

\renewcommand{\phi}{\ensuremath{\varphi}}

\usepackage[linesnumbered,ruled]{algorithm2e}
\usepackage{multicol}
\SetKwComment{Comm}{$\triangleright$\ }{}
\SetAlFnt{\scriptsize}
\SetAlCapFnt{\scriptsize}
\SetAlCapNameFnt{\scriptsize}
\SetKwInOut{Input}{Input}
\SetKwInOut{Output}{Output}

\makeatletter
\NewDocumentCommand{\LeftComment}{s m}{%
\Statex \IfBooleanF{#1}{\hspace*{\ALG@thistlm}}\(\triangleright\) #2}
\makeatother

%% file: intro.tex
\section{Introduction}

\begin{figure*}[t]
    \centering
    \includegraphics[width=1.0\textwidth]{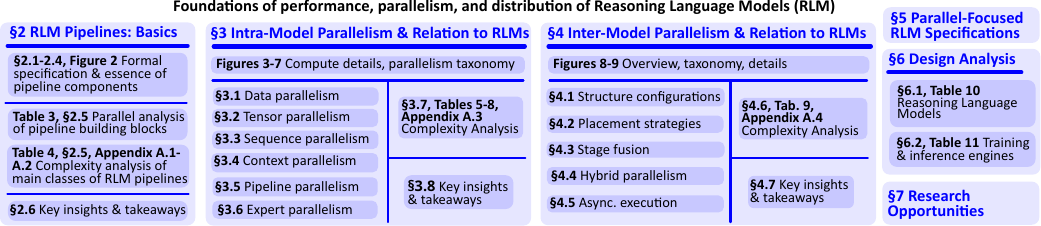}
    \vspace{-1.5em}
    \caption{\textbf{Overview of the contributions and analyses in this work.}}
    \label{fig:summary}
    \vspaceSQ{-2.5em}
\end{figure*}

Reinforcement learning (RL) is now the computationally dominant post-training paradigm for aligning large language models (LLMs) with reasoning standards. After pretraining and supervised fine-tuning (SFT), RL optimizes a policy against learned reward models or other feedback signals to improve task-specific performance beyond next-token prediction~\cite{ouyang2022training, bai2022helpful}.
The resulting frontier \emph{Reasoning Language Models} (RLMs), such as DeepSeek-R1~\cite{guo2025deepseekr1}, OpenAI's o3~\cite{openai2025o3o4}, and Kimi k1.5~\cite{du2025kimi}, demonstrate that carefully designed preference learning and policy optimization over Chain-of-Thought trajectories can produce substantially stronger reasoning, long-horizon planning, and self-correction than SFT alone. A key paradigm behind this trend is \emph{RL with Verifiable Rewards} (RLVR), in which reward signals are obtained from automatically checkable outcomes, such as exact-answer verification in mathematics, unit tests in code, or other task-specific verifiers~\cite{wen2025rlvr, hu2025openreasonerzero, mroueh2025reinforcement}.

The computational footprint of RLMs is enormous~\cite{liu2025deepseekv3}. For example, OpenAI reports that in developing o3 it scaled RL by an \textit{additional order of magnitude} in both train-time RL compute and inference-time reasoning compute, while still observing clear gains from further scaling~\cite{openai2024o3}. More broadly, cost analyses of frontier model training indicate that end-to-end training runs for leading systems already cost tens to hundreds of millions of US dollars in compute alone, and they may reach more than \$1B by 2027~\cite{cottier2024frontiercost}.


In contrast to classical supervised LLM training, RLM pipelines~\cite{besta2025reasoning} couple several large models and stages. A standard Proximal Policy Optimization (PPO)-style setup maintains four distinct LLMs (actor (policy), reward, critic (value), and a reference model) and repeatedly executes them across three stages (Generation, Assessment\footnote{While the Assessment stage is often called ``Inference'' in the literature,
we use the term ``Assessment'' because it more precisely describes its
functional role. Indeed, inference is not specific to this stage:
Generation also performs inference, albeit autoregressively.}, Training) with complex data dependencies. 
Recent reasoning-oriented systems often use critic-free Group Relative Policy Optimization (GRPO)-style training in practice, largely because removing the critic reduces memory pressure and simplifies large-scale rollout training~\cite{shao2024deepseekmath, guo2025deepseekr1}. Nevertheless, PPO-style actor--critic training remains a key and relevant design point as its learned value baseline reduces gradient variance and stabilizes policy updates, providing more faithful token-level credit assignment than critic-free alternatives such as GRPO~\cite{de2025learning, wang2026sppo}. PPO has been used or revived in reasoning-focused RL systems such as VAPO~\cite{yue2025vapo}. Overall, both PPO and GRPO as well as their variants are crucial RLM training frameworks.

Systems used for RL post-training, for each ReaL~\cite{mei2025real}, RLHFuse~\cite{zhong2024optimizing}, and OpenRLHF~\cite{hu2025openrlhf}, have shown that naively reusing supervised-training parallelization strategies for RLMs yields suboptimal GPU utilization. Instead, the RLM workflow demands joint reasoning about model placement, inter-model scheduling, and heterogeneous workload characteristics. This combination of massive scale and multi-model RL pipelines makes RLMs fundamentally a \emph{parallel and distributed systems} problem.


To facilitate the development of next-generation scalable and cost-effective RLMs, we conduct an in-depth investigation into the foundations of their parallel and distributed computational characteristics; a roadmap of the paper can be found in Figure~\ref{fig:summary}. First, we systematize the RL-for-LLM paradigm and provide a compute-centric analysis of several prominent post-training frameworks (\textbf{contribution~1}), including PPO-style~\cite{zheng2023secrets} and GRPO-like methods~\cite{shao2024deepseekmath}. We also consider preference-gradient approaches such as Direct Preference Optimization (DPO)~\cite{xiao2024comprehensive} -- while these methods do not use RL, they have been shown to also enhance reasoning capabilities~\cite{wang2026vpo, lai2024step, wang2024self, xu2025full}. 
%

Next, we develop a taxonomy of parallelism for RLMs (\textbf{contribution~2}). The first part focuses on \textbf{intra-model} parallelism, instantiated by LLM-based policies, critics, rewards, and references. For this, we analyze traditional forms of parallelism and how they interact with the RL pipeline, for instance which RL stage benefits most from which form of intra-model parallelism. Here, we consider data parallelism (DP)~\cite{li2020pytorch}, tensor parallelism (TP, also referred to as operator parallelism~\cite{shoeybi2019megatron, ben2019demystifying}, pipeline parallelism (PP)~\cite{huang2019gpipe, narayanan2021efficient}, sequence parallelism (SP)~\cite{korthikanti2022reducing}, context parallelism (CP)~\cite{liu2023ring}, and expert parallelism (EP)~\cite{lepikhin2020gshard}).

The second part of our taxonomy targets \textbf{inter-model parallelism}: opportunities to run {different models and RL stages} concurrently, and system-level optimizations that reduce end-to-end iteration time by reshaping how these components interact.
We show that the available forms of inter-model parallelism are largely determined by a small set of concrete design choices:
\emph{model structure configurations}, which can {eliminate} redundant model executions (e.g., critic--reward merging), 
\emph{model placement strategies}, which enable {component-level concurrency} by co-locating models on the same devices or disaggregating them onto separate device groups/services;
\emph{stage and sub-task fusion}, which introduces {stage-level overlap} (e.g., overlapping long-tailed generation with reward/value evaluation, or pipelining actor and critic updates) to remove serialization between stages;
\emph{hybrid intra-model configurations across components}, where different models and stages use different mixtures of intra-model parallelism; and
\emph{asynchronous execution}, which decouples rollout generation from learning via bounded staleness.
Our taxonomy provides a common performance-oriented language for formulating, comparing, and implementing new optimizations for large-scale RLM training.

To ensure that all the insights are portable across different architectures, we use work/depth/memory complexities for all the parts of the taxonomy (\textbf{contribution~4}).

We then use the taxonomy to conduct a systematic analysis of existing RLM models and frameworks and analyze them through the lens of the parallelism and optimization strategies they implement (\textbf{contribution~5}). Concretely, we examine general-purpose libraries such as TRL~\cite{vonwerra2022trl}, ColossalChat~\cite{colossalchat-medium}, DeepSpeed-Chat~\cite{yao2023deepspeed}, OpenRLHF~\cite{hu2025openrlhf}, HybridFlow~\cite{sheng2024hybridflow}, and NeMo RL~\cite{shen2024nemoaligner,nvidia2026nemorl}, as well as more specialized systems including ReaL~\cite{mei2025real}, RLHFuse~\cite{zhong2024optimizing}, FlexRLHF~\cite{xiao2025flexrlhf}, StreamRL~\cite{zhong2025streamrl}, Asynchronous RLHF~\cite{noukhovitch2025asynchronous}, AReaL~\cite{fu2025areal}, and Pipe-RLHF~\cite{xu2025piperlhf}. For each framework, we catalogue which intra- and inter-model techniques are supported. This enables practitioners to make informed choices about which framework best matches their needs, and it reveals systematic gaps where our taxonomy suggests additional optimizations (for example, underexplored stage-fusion and placement policies for multi-model PPO-style pipelines). In this way, our study not only describes the current ecosystem but also points to concrete directions for enhancing existing RL frameworks and designing new ones.

\subsection{Complementary Analyses \& Related Work}


There exist general studies on the blueprint for RLMs~\cite{besta2025reasoning} and broad surveys on LLM-enhanced RL~\cite{wang2025reinforcment, cao2025survey}. Other works provide comprehensive analyses of specific components of the RL pipeline~\cite{zheng2023secrets, wang2024secrets, xiao2024comprehensive}.

Some surveys detail techniques for resource-efficient LLMs, covering algorithms, model compression, and systems~\cite{ding2024efficiency, bai2024beyond, xu2025resource}. More broadly,  \cite{ben2019demystifying} offers a foundational concurrency analysis of parallel and distributed deep learning. However, these works do not focus on the unique demands of RLM pipelines.

Closer to our work, Liu et al.~explore acceleration techniques for deep RL~\cite{liu2024acceleration} and Chen et al.~survey the effects of scaling LLM's reasoning capabilities~\cite{chen2025survey}. However, these works do not explain the underlying performance foundations of the parallel and distributed systems that enable RLMs to scale. Specifically, the latter focuses on how reasoning capabilities emerge with scale, not on the concurrency and system bottlenecks of the RLM training loop. Similarly, the former analyzes acceleration for general deep RL, not the specific parallel architectures required by modern RLM pipelines.

We complement these studies by providing the first in-depth analysis to demystify the performance foundations of parallel and distributed RLMs. 














\subsection{Analysis of Parallel Algorithms}

We use formal models for reasoning about parallelism. Specifically, we use \textit{the work-depth (WD) analysis}, an established approach for bounding runtimes of parallel algorithms. The \textbf{work} ($W$) of an algorithm is the total number of operations and the \textbf{depth} ($D$) is defined as the longest sequential chain of execution in the algorithm, and it forms the lower bound on the algorithm execution time~\cite{blelloch1996parallel, bilardi2011models}. One usually wants to minimize depth while preventing work from increasing too much.
Finally, \textbf{Memory} accounts for model parameters, gradients, optimizer state (Adam first and second order moments), and stored activation tensors during a single training iteration.

%% file: background.tex
\input{table-notation}

\begin{figure*}[t]
    \centering
    \includegraphics[width=1.0\textwidth]{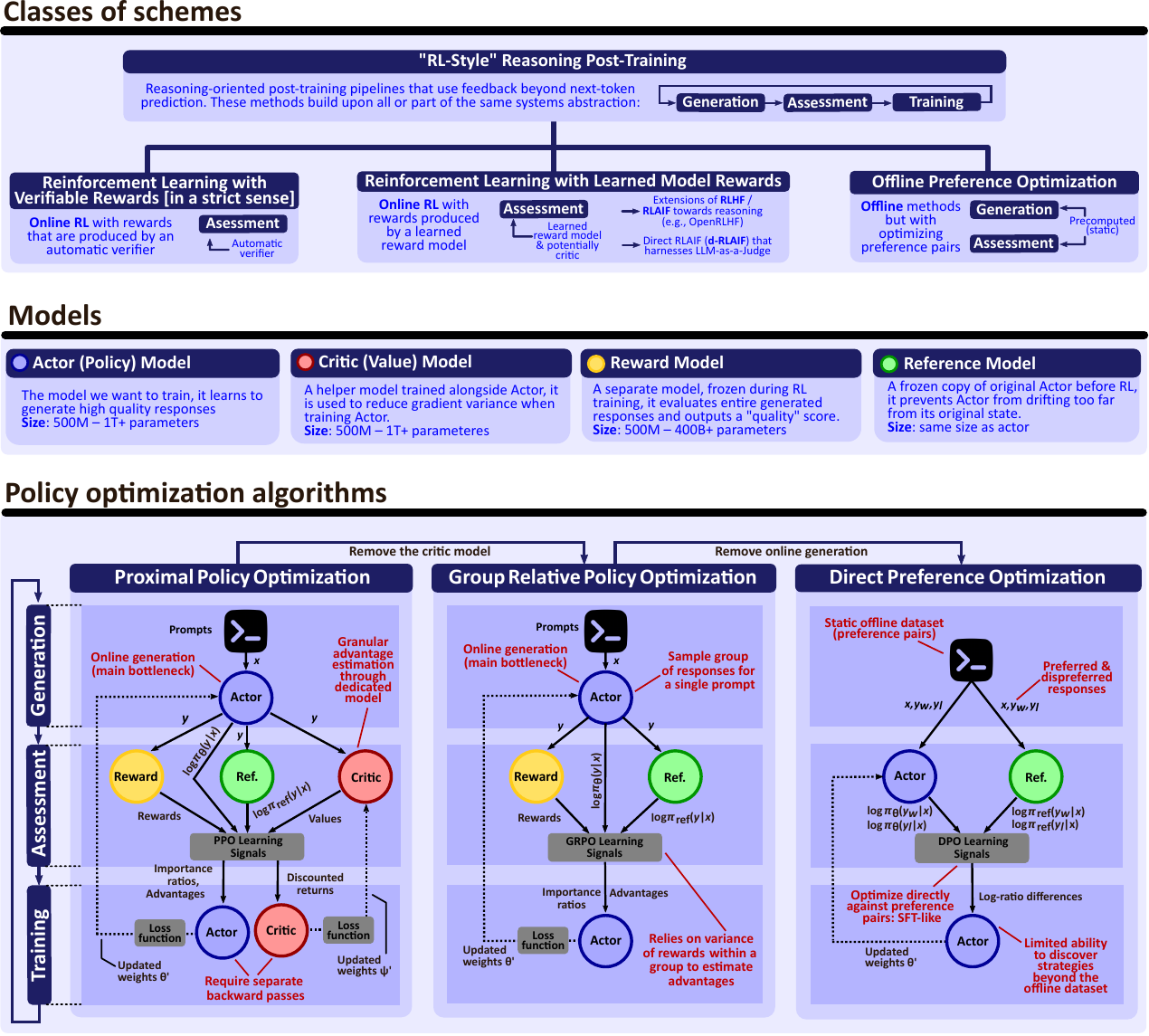}
    \vspace{-1.5em}
    \caption{\textbf{Overview of policy optimization algorithms and used models.} \textit{No AI was used to conceive or to draw the figure.}}
    \label{fig:overview}
    \vspaceSQ{-1.5em}
\end{figure*}

\section{Overview \& Foundations of RLMs}

We first overview basic RLM post-training concepts.
%


\subsection{RL Post-Training Pipeline: Overview}

The essence of RL for LLM can be captured as a simple three-stage loop that iterates continuously: \emph{Generation} $\rightarrow$ \emph{Assessment} $\rightarrow$ \emph{Training} $\rightarrow$ \emph{Generation} $\rightarrow \dots$. These stages operate over a \emph{batch} of input prompts, denoted by $X = \{x_1, x_2, \dots, x_B\}$, where $B$ is the batch size. Starting from a language model that has typically been pre-trained and then fine-tuned via supervised learning (SFT), this loop further improves this model (referred to as the \emph{policy model} $\pi_\theta$) through RL-based optimization. In each iteration, the \emph{Generation} stage produces candidate responses to input prompts using the current policy $\pi_\theta$. Then, \emph{Assessment} evaluates those responses and produces feedback (which can be both scalar and vector). Finally, \emph{Training} uses the resulting feedback to update the policy $\pi_\theta$. Both Assessment and Training stages may invoke one or more auxiliary models that provide the feedback signals used to improve the policy $\pi_\theta$, i.e., the \emph{reward model} $R_\varphi$, the \emph{critic model} $V_\psi$, and the \emph{reference model} $\pi_{\mathrm{ref}}$. This iterative process continues until convergence criteria are met or a predetermined number of training loops are completed, with the updated $\pi_\theta$ model from each training stage serving as the starting point for the next cycle.
We overview this process in Figure~\ref{fig:overview}; additional mathematical details are in Appendix~\ref{sec:app:pipeline} and in Table~\ref{tab:notation}.

\subsection{Terminology: RLVR, RLHF, RLAIF \& Others}

We use \emph{\textbf{RL-style reasoning post-training}} as the umbrella term for reasoning-oriented post-training pipelines that use feedback beyond next-token prediction. This term is broader than strict online RL: it includes \textbf{(1)} RLVR, where online RL algorithms such as PPO and GRPO-style methods use verifiers as rewards; \textbf{(2)} online RL-based post-training with model-based assessment, including RL from Human Feedback (RLHF), RL from AI Feedback (RLAIF), direct RLAIF (d-RLAIF), and learned reward methods; and \textbf{(3)} offline preference-optimization methods such as DPO, which are not RL in their basic form but have also been used to enhance reasoning. This broader umbrella is also reflected in systems practice: frameworks originally named after RLHF are now used for broader reasoning post-training. For example, OpenRLHF~\cite{hu2025openrlhf} supports PPO- and GRPO-style training, while RLHFuse~\cite{zhong2024optimizing} optimizes the same staged workflow used by both RLHF and RLVR systems.

\textbf{Strict RLVR} denotes online RL with rewards produced by an automatic verifier rather than a learned reward model. Examples include exact-answer matching in mathematics, unit-test execution or compilation in code, theorem proving, and other programmatic correctness checks~\cite{shao2024deepseekmath,guo2025deepseekr1}.

A second class uses \textbf{model-based assessment}. In classical RLHF, rewards are derived from human preference data, typically through a learned reward model~\cite{ouyang2022training}. In canonical RLAIF, human preference labels are replaced by AI-generated preferences or critiques, which are then used to train a reward model; Constitutional AI is a canonical example~\cite{bai2022constitutional,lee2024rlaif}. A closely related variant is \emph{direct RLAIF} (d-RLAIF), where an LLM directly evaluates policy outputs and provides rewards during RL, without training a separate reward model~\cite{lee2024rlaif}. Thus, LLM-as-a-judge reward schemes can naturally be viewed as instances of d-RLAIF rather than a separate feedback class. Neither RLAIF nor d-RLAIF is strict RLVR because their Assessment signal is an AI-generated judgment rather than an externally verifiable correctness check.

\textit{Learned assessment models are nevertheless important for reasoning}. Process-supervised reward models improve mathematical reasoning~\cite{lightman2024let}; Math-Shepherd trains a process reward model and uses step-level PPO to improve GSM8K and MATH performance~\cite{wang2024math}; there are further similar examples~\cite{le2022coderl,dai2024process,gorinski2023automatic}. These examples motivate retaining reward-model terms in our complexity analysis. When Assessment is implemented by a learned outcome or process reward model, it requires LLM forward passes; when process-level feedback is used, the evaluator may be invoked at many reasoning steps. Such learned-assessment pipelines can therefore be more compute-heavy than lightweight exact-match RLVR, even though both instantiate the same Assessment role.

A third class consists of \textbf{offline preference-optimization methods}. DPO is not RL in its basic form: it inherits the preference-learning objective of RLHF but optimizes static preference pairs without online rollout generation or explicit reward-model inference~\cite{rafailov2023direct}. Yet, DPO-style methods have also been adapted to reasoning, for example through iterative reasoning preference optimization and step-wise DPO~\cite{wang2026vpo,lai2024step,wang2024self,xu2025full}. We include them because they share parts of the same post-training staged dataflow with policy/reference forward passes and policy backpropagation.

Overall, all these methods build upon all or part of the same systems abstraction: \emph{Generation} $\rightarrow$ \emph{Assessment} $\rightarrow$ \emph{Training}. Online RL methods such as RLHF, RLAIF/d-RLAIF, and RLVR execute the full loop: the policy samples trajectories, Assessment maps them to rewards, model judgments, verifier outcomes, advantages, or other learning signals, and Training updates the policy and potentially critic. Offline preference methods such as DPO remove online Generation and explicit reward Assessment from the inner loop, but retain the policy/reference comparison and Training components. Our analysis therefore targets the broad class of RL-style reasoning post-training systems, where the Assessment stage may be instantiated by a learned reward model, an LLM evaluator, an automatic verifier, or an offline preference signal.

\subsection{Auxiliary Models in the RL-LLM Pipeline}

We now investigate the functional and execution characteristics of models used in the RL pipeline, see Table~\ref{tab:model-taxonomy} and Figure~\ref{fig:overview} (top). While conceptually distinct, in practical settings, they share similar LLM-based architectures.

\input{table-models.tex}

\subsubsection{Reward Model}

The {reward model} is a function $R_\varphi : \mathcal{X} \times \mathcal{Y} \to \mathbb{R}$ that maps a prompt–response pair $(x, y)$ to a scalar reward $r$, i.e., $R_\varphi(x, y) = r$, representing overall response quality, helpfulness, or alignment with desired behavior. These rewards are typically learned from human preference data and serve as the principal alignment signal for the policy. During the {Assessment} stage, the reward model performs a \emph{single forward pass per full sequence} to produce $r_b^{(i)}$. It is a frozen model (no gradients computed). Moreover, as scalar rewards are usually single-token, its execution is usually non‑autoregressive, which makes its evaluation considerably cheaper than autoregressive policy generation.

There are two types of reward models: \textbf{outcome-based} and \textbf{process-based} reward models ($R^\text{proc}_\varphi$). The former is a standard mechanism that assigns a single scalar score $r$ to a \textit{completed sequence}. Process-based reward models  provide feedback at \textit{intermediate steps} of reasoning. Formally, it maps a sequence to a vector of rewards. While this provides a richer supervision signal, it significantly increases computational cost (as it may require multiple forward passes or Monte Carlo Tree Search (MCTS) integration) and is difficult to train~\cite{guo2025deepseekr1, besta2025reasoning}.

\subsubsection{Critic Model}

The {critic model} is a function defined as $V_\psi : \mathcal{X} \times \mathcal{Y}_{\le t} \to \mathbb{R}$. It estimates the expected future reward from a partial sequence, i.e., $V_\psi(x, y_{<t}) = v_t$. Given a prompt $x$ and prefix $y_{<t}$, the critic predicts $v_t$ as an estimate of the return that the policy can expect from that point onward, i.e., $V_\psi(x, y_{<t}) \approx \mathbb{E}_{\pi_\theta}\!\left[ \sum_{\tau = t}^{T} \gamma^{\tau-t} r_\tau \mid x, y_{<t} \right]$. When $r$ comes from a learned reward model, we have $r_T = R_\varphi(x, y_{1:T})$ and $r_t = 0$ for $t < T, \gamma =  1$. Unlike the reward model, the critic provides \emph{token-level} information, producing value estimates for all positions $t$ in a single forward pass over the full sequence. During the {Assessment} stage, these predictions are combined with observed rewards to compute advantage estimates $A_t$. These advantage estimates are then used in {Training} to guide gradient updates of the policy model. The critic runs a full forward pass per sampled sequence and a backward pass for its parameter update, often in parallel with the policy gradient computation.


Unlike reward models, the critic is \textit{trainable}. The reward model $R_\varphi$ acts as a fixed proxy for task quality; keeping it fixed during policy optimization provides a stationary reward objective. The critic $V_\psi$, however, estimates the expected return under the \emph{current} policy. As $\pi_\theta$ changes, the distribution of rollouts and their expected returns changes as well. Therefore, PPO-style methods update $V_\psi$ after each rollout batch by minimizing a value-regression loss $\frac{1}{|\mathcal{B}|} \sum_{(x,y)\in\mathcal{B}}\sum_t \left(V_\psi(x,y_{<t})-\hat{G}_t\right)^2$, where $\hat{G}_t$ is a Monte-Carlo or GAE-style return target computed from the rewards collected on that batch. In this way, the critic tracks the return function induced by the evolving policy.

\subsubsection{Reference Model}

The {reference model} $\pi_{\mathrm{ref}}$ is a frozen policy $\pi_{\mathrm{ref}} : \mathcal{X} \times \mathcal{Y} \to [0,1]$ that serves as a fixed baseline to stabilize optimization. It is typically a snapshot of the policy after SFT and before any RL. During the {Assessment} stage, it provides per-token log-probabilities $\log \pi_{\mathrm{ref}}(y_t \mid x, y_{<t})$ used to measure deviation from the current policy. During the {Training} stage, these outputs are reused for regularization, e.g., through a KL-divergence penalty $D_{\mathrm{KL}}(\pi_\theta \| \pi_{\mathrm{ref}})$ or preference deltas in DPO. Since it remains frozen, $\pi_{\mathrm{ref}}$ requires only forward passes, but it can still be expensive to evaluate at scale because it performs a full-sequence teacher-forced forward pass over every rollout.

\subsection{Auxiliary Models vs.~Algorithmic RL Frameworks}

Depending on what auxiliary models are used, there are \textbf{PPO-like methods}, \textbf{GRPO-like methods}, and \textbf{DPO-like methods}, with distinct computational trade-offs. PPO~\cite{schulman2017proximal} uses all four models, requiring both forward and backward passes for $\pi_\theta$ and $V_\psi$, and additional forward-only evaluation for $R_\varphi$ and $\pi_{\mathrm{ref}}$. GRPO~\cite{guo2025deepseekr1, liu2025deepseekv3, shao2024deepseekmath} omits the critic, saving backward passes but still requiring reward inference. DPO~\cite{rafailov2023direct} relies solely on two models ($\pi_\theta$ and $\pi_{\mathrm{ref}}$), resulting in an offline teacher-forced forward/backward pipeline (without online Generation or explicit reward-model inference) that is computationally lightweight and easier to scale; however, it also takes as input the pre-computed preference dataset. Thus, the combination of these models not only defines the algorithmic behavior of the RL method but also determines its computational profile, parallelism potential, and system-level design.

\subsubsection{Variants and Emerging Frameworks}

There are numerous variants of PPO, GRPO, and DPO-like methods, we now summarize most important ones, focusing on their computational characteristics.

\textbf{Within the PPO family}, VAPO~\cite{yue2025vapo} retains the value model
but improves value-based training efficiency and stability through value
pretraining and decoupled GAE, so its compute remains recognizably
actor--critic rather than critic-free. RTO~\cite{zhong2025rto} shifts
supervision toward token-level rewards and advantages, increasing per-token
bookkeeping during training. Safe-RLHF~\cite{dai2024safe} extends PPO-like
RLHF with a separate cost model and a Lagrangian safety constraint, thereby
adding another assessment model while making helpfulness--harmlessness trade-offs
explicit. Other PPO-style trust-region variants include divergence-based
DPPO~\cite{qi2026dppo}, which replaces ratio clipping with direct divergence
estimates while preserving the basic online actor--critic profile, and
mirror-descent formulations such as MDPO~\cite{tomar2021mdpo}, which largely
preserve the actor--critic compute structure while changing
the policy update from PPO's clipped-ratio surrogate to a Bregman- or
KL-regularized mirror-descent step around the previous policy.

\textbf{Within the GRPO family}, DAPO~\cite{yu2025dapo} improves sample and FLOP
efficiency through asymmetric clipping, dynamic sampling, token-level losses, and
better handling of overlong trajectories, while pruning-based extensions such as
DPPO~\cite{zhen2026dppo} reduce wasted decoding and training work by pruning
unpromising trajectories with unbiased correction. Other GRPO-like extensions
such as Graph-GRPO~\cite{cang2026graphgrpo}, RiskPO~\cite{ren2025riskpo},
GBMPO~\cite{yuan2026gbmpo}, MicroCoder-GRPO~\cite{li2026microcoder}, and
PIPO~\cite{wang2026pipo} mostly preserve the same critic-free online compute
profile while changing the regularizer, feedback signal, or application setting.
Related critic-free online methods such as RLOO~\cite{ahmadian2024rloo} and
ReMax~\cite{li2024remax} also avoid value-model training, but replace learned
critics with leave-one-out, sample-based, or greedy-response baselines rather
than GRPO's group-relative normalization.

\textbf{DPO-style variants} typically preserve the offline, SFT-like compute
profile in which the policy is updated by batched teacher-forced passes over
preference data. IPO~\cite{gheshlaghiazar2024psi} changes the pairwise objective
without materially changing this profile; ORPO~\cite{hong2024orpo} combines the
SFT term with an odds-ratio preference penalty and removes the explicit reference
model; SimPO~\cite{meng2024simpo} also removes the reference model and thus
reduces memory and forward cost; KTO~\cite{ethayarajh2024kto} uses binary
desirability signals instead of paired preferences; and AlphaDPO~\cite{
wu2025alphadpo} introduces adaptive margins while keeping the basic offline
scaling behavior. Reward-aware preference objectives~\cite{sun2025reward}
partially reintroduce reward information into this family: they add quality-aware
signals to the preference loss without returning to the full online
actor--critic loop.

Finally, \textbf{emerging likelihood-oriented methods} such as MaxRL~\cite{tajwar2026maxrl} target reasoning-heavy settings with compute-indexed sampling objectives that interpolate between standard RL and
maximum-likelihood optimization as additional sampling compute is allocated.
Overall, these variants do not change the main systems distinction: online
methods are dominated by rollout generation, critic-free methods remove
\(V_\psi\)-dependent costs, and offline preference methods trade exploration for
cheaper batched training.

\subsection{Computational Analysis of RL-LLM}
\label{sec:comp-anal-frameworks}

We now analyze computational complexities of RL-LLMs. Mathematical derivations are detailed in Appendix~\ref{sec:app:derivations}.


\input{table-building-blocks}

We first derive the computational costs of \textbf{building blocks}, see Table~\ref{tab:building-blocks}. Namely, each harnessed model comes with forward and backward pass costs (work, depth, and memory consumption). These costs are similar or identical (in the asymptotic sense) across different models, because all models are based on the transformer architecture.


\input{table-framework-taxonomy}

Next, we obtain the \textbf{complexities for PPO, GRPO, and DPO}, see Table~\ref{tab:framework-complexity}. Overall, the work--depth expressions make the ordering precise. Among the online methods, PPO is the most expensive because its per-iteration work contains all major terms, and its training memory also includes both trainable models. GRPO removes the critic \(V_\psi\), so it eliminates the additional \(\Theta(BK(S+T)[C_f^{\text{tok}}(V_\psi)+C_b^{\text{tok}}(V_\psi)])\) work and the corresponding \(\Theta(|V_\psi|)\) model-state memory, but it retains the same dominant online bottleneck as PPO, namely the autoregressive generation depth \(D_{\mathrm{gen}}=\Theta(T\cdot D_f(\pi_\theta))\).
By contrast, DPO removes online rollout generation and explicit reward/critic evaluation from the inner loop, leaving only batched forward/backward passes over preference pairs. Thus, unlike PPO and GRPO, DPO has no \(T\)-step autoregressive term in its inner-loop depth, which explains why its systems profile is much closer to standard SFT. The trade-off is algorithmic: because DPO optimizes over a fixed preference dataset rather than fresh online rollouts, it cannot directly explore behavior outside that dataset.

\iftr
Later in Section~\ref{sec:alg-specs}, we also present detailed algorithmic \textbf{parallelism-focused specifications} (Algorithms~\ref{alg:generation-common}, \ref{alg:ppo}, \ref{alg:grpo}, and~\ref{alg:dpo}) that explicitly annotate most relevant parallel execution.
\fi

\subsection{Key Insights \& Takeaways}

We summarize key takeaways.

\noindent
\trianbox1{cblue} 
\textbf{Autoregressive generation imposes an irreducible per-rollout
dependency.}
For PPO and GRPO, each rollout is generated token by token, yielding a
per-trajectory depth of
$O\!\left(TD_f(\pi_\theta;S+T)\right)$.
Parallelizing Assessment or Training cannot remove this sequential
dependency within an individual trajectory. Consequently, efficient
decoding kernels remain
central to reducing rollout latency and increasing generation throughput.

\noindent
\trianbox1{cblue} 
\textbf{Generation is not necessarily the end-to-end system bottleneck.}
The sequential depth of an individual rollout does not imply that the
entire Generation stage must dominate iteration time. Through inter-stage
fusion, completed samples or rollout subbatches can be streamed into
Assessment while longer trajectories or later rollout waves are still
being generated. This overlap is useful when completion lengths are
skewed or when finite generation capacity forces the rollout batch to be
processed in multiple waves. It trades additional memory (e.g., through live KV-cache and
communication buffers) for reduced pipeline
idle time.

\noindent
\trianbox1{cblue} \textbf{Auxiliary models define the systems profile.}
PPO uses \(\{\pi_{\mathrm{ref}},R_\phi,V_\psi\}\), GRPO removes \(V_\psi\), and DPO removes both \(R_\phi\) and \(V_\psi\) from the inner loop. Hence the PPO--GRPO gap is precisely the critic cost $\Theta\!\left(BK(S+T)[C_f^{\text{tok}}(V_\psi)+C_b^{\text{tok}}(V_\psi)]\right)$ plus critic model-state memory. The GRPO--DPO gap is more structural: DPO also removes online rollout generation from the optimization loop, eliminating the \(T\)-step generation depth term.

\noindent
\trianbox1{cblue} \textbf{Frozen models are placement opportunities.}
The reward and reference models are forward-only. They require no gradients or optimizer state, so they can be replicated, quantized, offloaded, or served as independent inference services. By contrast, trainable models \(\pi_\theta\) and \(V_\psi\) dominate memory through activations, gradients, and optimizer state.

\noindent
\trianbox1{cblue} \textbf{Reward granularity is a compute--credit-assignment trade-off.}
Outcome rewards are cheap: one scalar per completion. Process rewards can improve reasoning supervision but may require step-level annotations, verifier calls, or search, turning Assessment from one batched forward pass into a much heavier verification workload.

\noindent
\trianbox1{cblue} \textbf{DPO is SFT-like, but not free.}
DPO still performs {\small $W_{\mathrm{DPO}} = \Theta\!\left(B_{\mathrm{pairs}}(S+T)[C_f^{\text{tok}}(\pi_\theta)+C_f^{\text{tok}}(\pi_{\mathrm{ref}})+C_b^{\text{tok}}(\pi_\theta)]\right)$}, but these are teacher-forced passes over static preference pairs. Its systems advantage is therefore the removal of online generation and reward/critic inference; its algorithmic limitation is that it cannot directly explore beyond the support of the preference dataset.

%% file: table-notation.tex
\begin{table*}[ht!]
\footnotesize
    \centering
    \begin{tabular}{@{}ll@{}} 
        \toprule
        \multicolumn{2}{l}{\textbf{Models \& training signals used in the RL pipeline}} \\
        \midrule
        $\pi_\theta$ & Policy (Actor) model: the trainable LLM being optimized. \\
        $\pi_{\mathrm{ref}}$ & Reference model: the frozen baseline policy (usually a checkpoint right after SFT). \\
        $R_\varphi$ & Reward model: an assessment tool that maps prompt--response pairs to scalar rewards. \\
        $V_\psi$ & Critic model (Value function): an assessment tool that estimates expected future returns from the reward model. \\
        $r_b^{(i)}, A_{b,t}^{(i)}, \hat{G}_{b,t}^{(i)}$ & Reward for the $i$-th candidate of prompt $b$, advantage at token step $t$, and discounted return estimated at token step $t$. \\
        \midrule
        \multicolumn{2}{l}{\textbf{Data \& Dimensions}} \\
        \midrule
        $\mathcal{X}, \mathcal{Y}$ & Space of input prompts and token sequences, respectively\\
        $X = \{x_1, \dots, x_B\}$ & Batch of $B$ input prompts; $X \subset \mathcal{X}$. \\
        $y = (y_1, \dots, y_T)$ & A response sequence consisting of $T$ tokens ($y \in \mathcal{Y}$). \\
        $K$ & Number of candidate responses generated per prompt (in the Generation stage). \\
        $S, T$ & The average prompt length and the average response length. \\
        \midrule
        \multicolumn{2}{l}{\textbf{LLM \& system design}} \\
        \midrule
        $N, L, V, B$ & Number of parallel devices, number of layers, vocabulary size, batch size, respectively. \\
        $d, d_{\text{ff}}, h, d_k = d / h$ & Hidden dimension, the intermediate FFN dimension, number of parallel heads, and per-head dimension, respectively. \\
        $E, E_a, d_e$ & The number of experts per layer, the number of active experts per token, and the expert hidden dimension, respectively. \\
        \bottomrule
    \end{tabular}
    \vspace{-0.5em}
    \caption{\textbf{Overview of basic mathematical notation and concepts used in the paper.} Notation specific to the complexity analyses is detailed separately in Table~\ref{tab:building-blocks}.}
    \label{tab:notation}
\end{table*}

%% file: table-models.tex
\begin{table}[h!]
\centering
\scriptsize
\setlength{\tabcolsep}{3pt} 
\begin{tabular*}{\columnwidth}{l l l p{3.9cm} }
\toprule
\textbf{Model} & 
\textbf{State} & 
\textbf{Used in} & 
\textbf{System design remarks} \\
\midrule
\makecell[tl]{\textbf{Policy}\\ \textbf{(Actor)} $\pi_\theta$} & 
\textcolor{red}{Trainable} & 
\makecell[tl]{Generation,\\ Training} & 
\textbf{Bottleneck:} Sequential decoding dominates wall-clock time. \\
\midrule

\makecell[tl]{\textbf{Reward} $R_\varphi$\\ (Outcome)} & 
\textcolor{blue}{Frozen} & 
Assessment & 
Efficient parallel evaluation; no backward pass. Easy to offload. \\
\midrule

\makecell[tl]{\textbf{Reward} $R_\varphi^{\text{pr}}$\\ (Process)} & 
\textcolor{blue}{Frozen} & 
Assessment & 
High overhead; sequential decoding if using sequential verification (e.g., MCTS). \\
\midrule

\makecell[tl]{\textbf{Critic}\\ \textbf{(Value)} $V_\psi$} & 
\textcolor{red}{Trainable} & 
\makecell[tl]{Assessment,\\ Training} & 
Doubles training compute (fwd+bwd). High memory pressure (optimizer states). \\
\midrule

\textbf{Reference} $\pi_{\text{ref}}$ & 
\textcolor{blue}{Frozen} & 
Assessment & 
Required for KL/DPO. Forward-only; often quantized to save memory. \\

\bottomrule
\end{tabular*}
\vspace{-1em}
\caption{\textbf{Overview of models used in the RLM pipeline.}}
\label{tab:model-taxonomy}
\end{table}

%% file: table-building-blocks.tex
\begin{table*}[hbtp]
\centering
\scriptsize


\begingroup
\scriptsize
\setlength{\tabcolsep}{3pt}
\renewcommand{\arraystretch}{0.9}

\begin{tabular*}{\textwidth}{@{\extracolsep{\fill}} l c p{8.4cm} p{0.5\textwidth} }
\toprule
\textbf{Cat.} & \textbf{Symbol} & \textbf{Meaning and remarks} & \textbf{Mathematical expression} \\
\midrule

\multirow[t]{2}{*}{\textbf{Work}}
& $C_f^{\text{tok}}(M)$
& \textbf{Forward-pass FLOPs \underline{per token} for model $M$ on a context of length $S+T$ (input prompt length + response length).}
  The reference policy $\pi_{\mathrm{ref}}$ is architecturally identical to the actor $\pi_\theta$.
  Unembedding FLOPs per token are $2V d_\pi$ (policy only). The reward model $R_\varphi$ adds a projection head applied to the final token only, contributing $\tfrac{2 d_R}{S+T}$ FLOPs per token, whereas the value model $V_\psi$ applies a projection head to every token, contributing $2 d_V$ FLOPs per token.
& \parbox[t]{\linewidth}{%
  \vspace{-2pt}
  \(
  \begin{alignedat}{3}
    &C_f^{\text{tok}}(\pi_\theta)         & = & \ 2L_\pi\left[4d_\pi^2 + 2d_\pi d_{\pi\mathrm{ff}} + (S+T)d_\pi\right] + 2Vd_\pi \\
    &C_f^{\text{tok}}(\pi_{\mathrm{ref}}) & = & \ C_f^{\text{tok}}(\pi_\theta) \\
    &C_f^{\text{tok}}(R_\varphi)          & = & \ 2L_R\left[4d_R^2 + 2d_R d_{R\mathrm{ff}} + (S+T)d_R\right] + \tfrac{2d_R}{S+T} \\
    &C_f^{\text{tok}}(V_\psi)             & = & \ 2L_V \left[4d_V^2 + 2d_V d_{V\mathrm{ff}} + (S+T)d_V \right] + 2d_V \\
    \end{alignedat}
  \)
}\\

& $C_b^{\text{tok}}(M)$
& \textbf{Backward-pass FLOPs \underline{per token} for model $M$ on a context of length $S+T$ (input prompt length + response length).}
& \parbox[t]{\linewidth}{%
  \vspace{-2pt}
  \(
  \begin{aligned}
  C_b^{\text{tok}}(M) \ \ \approx 2 \cdot C_f^{\text{tok}}(M)
  \end{aligned}
  \)
}\\

& $C_{\text{gen}}^{\text{roll}}(\pi_\theta)$
& \textbf{Autoregressive generation FLOPs \underline{per rollout}} for generating $T$ tokens (response length) from a context of length $S$ (input prompt length) with KV caching (policy only). It is decomposed into prefill and decode. Prefill FLOPs are for processing the prompt of length $S$ once and initializing the KV cache before autoregressive decoding begins. Prefill attention is assumed to be dense; as it is causal only half of the full MM product are needed, giving $\approx S^2 d_\pi$ FLOPs for both $Q K^T$ and for $P V$; $O(S^2)$ and $O(S d)$ account for lower-order contributions (respectively -- softmax and scaling by $1/\sqrt{d_h}$ as well as RoPE and layer norms). Decode FLOPs are for generating the remaining $T-1$ tokens autoregressively with KV caching. At decode step $t$, the model attends to context length $S+t$.
& \parbox[t]{\linewidth}{%
  \vspace{-2pt}
  \(
  \begin{aligned}
  C_{\text{gen}}^{\text{roll}}(\pi_\theta) &= C_{\text{pf}}^{\text{roll}}(\pi_\theta)+C_{\text{dec}}^{\text{roll}}(\pi_\theta)\\
  C_{\text{pf}}^{\text{roll}}(\pi_\theta) &= 2L_\pi S\!\left(4d_\pi^2 + 2d_\pi d_{\pi \mathrm{ff}} + S d_\pi\right) \\ &+ L_\pi \left( O(S^2) + O(S d_\pi) \right) + 2V d_\pi\\
  C_{\text{dec}}^{\text{roll}}(\pi_\theta) &= \sum_{t=1}^{T-1}\!\Bigl[2L_\pi\!\bigl(4d_\pi^2 + 2d_\pi d_{\pi \mathrm{ff}} + 2(S+t)d_\pi\bigr) + 2V d_\pi\!\Bigr] \\ &+ L_\pi \left( O(S+t) + O(d_\pi) \right)
  \end{aligned}
  \)
}\\
\midrule

\multirow[t]{3}{*}{\textbf{Depth}}
& $D_f(M)$
& \textbf{Forward-pass depth} of model $M$ on a context of length $S+T$, defined as the length of the critical path assuming unbounded parallelism. The policy $\pi_\theta$ includes an additional unembedding (output projection) step, which contributes a $\log d_\pi$ term to the depth. For the reward model $R_\varphi$ and the critic $V_\psi$, the additional projection head contributes an extra $\log d$ term to the depth.
& \parbox[t]{\linewidth}{%
  \vspace{-2pt}
  \(
  \begin{alignedat}{3}
    &D_f(\pi_\theta)          & = & \ O\left(L_\pi\left[\log d_\pi + \log(S+T)\right] + \log d_\pi\right) \\
    &D_f(\pi_{\mathrm{ref}})  & = & \ D_f(\pi_\theta) \\
    &D_f(R_\varphi)           & = & \ O\left(L_R\left[\log d_R + \log(S+T)\right] + \log d_R\right) \\
    &D_f(V_\psi)              & = & \ O\left(L_V\left[\log d_V + \log(S+T)\right] + \log d_V\right) 
    \end{alignedat}         
  \)
}\\

& $D_f^{\mathrm{TP}}\!(M)$
& \textbf{Forward-pass depth} of model $M$ \textbf{with tensor parallelism} degree $P_t$ on a context of length $S+T$.
& \parbox[t]{\linewidth}{%
  \vspace{-2pt}
  \(
  \begin{aligned}
  D_f^{\mathrm{TP}}\!(M) \! &=  O \left(L_M \left[\log \! \tfrac{d_M}{P_t} + \log(S+T)\right] + \log \tfrac{d_M}{P_t}\right)
  \end{aligned}
  \)
}\\

& $D_b(M)$
& \textbf{Backward-pass depth} of model $M$ on a context of length $S+T$. It equals the depth of the forward pass for the corresponding model $M$ (which entails computing the gradients with respect to activations), plus the additional term $O(\log (B (S+T)))$ coming from computing the gradients with respect to model weights, where one has to accumulate gradients across batch (as these two branches can overlap, the expression given is a conservative bound).
& \parbox[t]{\linewidth}{%
  \vspace{-2pt}
  \(
  \begin{aligned}
  D_b(M) \ \ &= D_f(M) + O(\log (B (S+T)))
  \end{aligned}
  \)
}\\

& $D_{\text{gen}}(\pi_\theta)$
& \textbf{Autoregressive generation depth} for generating $T$ tokens given a prompt of length $S$. Prefill contributes one forward-pass depth (over context of length $S$); decode contributes one
sequential forward-pass depth per generated token (i.e., over context of length $S+t$ for the $t$-th token).
& \parbox[t]{\linewidth}{%
  \vspace{-2pt}
  \(
  \begin{aligned}
  D_\text{gen}(\pi_\theta) &= D_f(\pi_\theta;S) + \sum_{t=1}^{T-1} D_f(\pi_\theta;S+t) \\
  & = O\!\bigl(T \cdot D_f(\pi_\theta;S+T)\bigr) 
  \end{aligned}
  \)
}\\
\midrule

\vspace{0.5em}\multirow[t]{2}{*}{\textbf{Memory}}
& $\lvert M\rvert$
& \textbf{Model parameter count} of $M$ (weights and embeddings). It always includes the parameters from query, key, value and output projections ($4d^2$), two FFN projections ($2dd_{ff}$), and the final logit computation ($Vd$). For reward and critic models, there are also $d$ parameters in the final head that outputs the score(s).
& \parbox[t]{\linewidth}{%
  \vspace{-5pt}
  \(
  \begin{alignedat}{3}
    &\lvert \pi_\theta \rvert \ \ \ & = & \ L_\pi (4d_\pi^2 + 2d_\pi d_{\pi \mathrm{ff}}) + Vd_\pi, \,\lvert \pi_\text{ref} \rvert = \lvert \pi_\theta \rvert \\[-2pt]
    &\lvert \pi_\text{ac} \rvert & = & \ L_\text{s} (4d_\text{ac}^2 + 2d_\text{ac} d_{\text{ac},\mathrm{ff}}) + Vd_\text{ac} + d_\text{ac} \\[-2pt]
    &\lvert R_\varphi \rvert & = & \ L_R (4d_R^2 + 2d_R d_{R\mathrm{ff}}) + Vd_R +d_R \\[-2pt]
    &\lvert V_\psi \rvert & = & \ L_V (4d_V^2 + 2d_V d_{V\mathrm{ff}}) + Vd_V + d_V \\
    \end{alignedat}         
  \)
}\\

& $M_{\mathrm{Act}}$
& \textbf{Training activation memory} per processed token for model $M$. Under the
leading-order activation model used throughout the paper, activations required
for backpropagation scale with the number of layers and hidden width. Constant
factors from Q/K/V tensors, FFN intermediates, normalization, residual, and
other temporary tensors are suppressed. This assumes no activation
checkpointing.
& $M_{\mathrm{Act}}=\Theta(L_M d_M)$
\\

& $M_{\mathrm{Inf}}$
& \textbf{Forward-only inference-buffer memory} per processed token for model $M$.
Because intermediate layer buffers can be reused across layers and the
attention matrix is assumed not to be fully materialized, the leading-order
buffer scales with hidden width rather than layer count.
& $M_{\mathrm{Inf}}(M)=\Theta(d_M)$
\\

& $M_{\mathrm{KV}}$
& \textbf{KV-cache memory} required for one rollout of prompt length $S$ and response length $T$ during generation. If all $BK$ rollouts are generated concurrently, the peak KV-cache memory is $BK \cdot M_{\mathrm{KV}}$; if rollouts are generated sequentially across the $K$ samples per prompt, the peak reduces accordingly but the generation depth increases.
& \parbox[t]{\linewidth}{%
  \vspace{-2pt}
  \(
  \begin{aligned}
  M_{\mathrm{KV}} &= 2(S+T)L_\pi d_\pi
  \end{aligned}
  \)
}\\
\bottomrule
\end{tabular*}
\endgroup
\vspace{-1.em}
\caption{
\textbf{Computational building blocks used across the paper.} We assume canonical multi-head attention and two-projection FFN. For computing FLOP costs, an addition and a multiplication count as two separate FLOPs (i.e., a dot product of vectors of dimensionalities $d$ results in $d + (d-1) \approx 2 d$ FLOPs under this model). $C_f^{\text{tok}}$, $C_b^{\text{tok}}$, $C_{\text{gen}}^{\text{roll}}$ are the costs of (respectively) forward pass, backward pass, and autoregressive generation; defining them \underline{per token} and \underline{per rollout} effectively clarifies the notation for the subsequent analyses. The explicit depth expressions count the selected GEMM reduction chains and
parameter-gradient accumulation; they suppress the additional logarithmic
reductions from normalization, attention/vocabulary softmax, sampling, and
distributed collectives. For generation, we explicitly distinguish prefill and decode. Prefill processes the prompt once, while decode generates tokens sequentially with KV caching. This decomposition is important because rollout generation is the dominant online bottleneck and is not equivalent to a single teacher-forced forward pass over a sequence of length $S+T$.}
\label{tab:building-blocks}
\end{table*}

%% file: table-framework-taxonomy.tex
\begin{table*}[hbtp]
\centering
\scriptsize
\setlength{\tabcolsep}{0pt} 
\renewcommand{\arraystretch}{1.3} 

\begin{tabular*}{\textwidth}{@{\extracolsep{\fill}} l c c c }
\toprule
\textbf{Framework} & \textbf{PPO (Online)} & \textbf{GRPO (Online)} & \textbf{DPO (Offline)} \\
\midrule

\multicolumn{4}{l}{\textit{\textbf{Work (Total FLOPs)}}} \\
Generation 
  & $O(BK \cdot C_\text{gen}^{\text{roll}}(\pi_\theta))$ 
  & $O(BK \cdot C_\text{gen}^{\text{roll}}(\pi_\theta))$ 
  & $-$ \textsuperscript{\dag} \\

Assessment        
  & $O(BK(S+T) \cdot [C_f^{\text{tok}}(\pi_{\text{ref}}) + C_f^{\text{tok}}(R_\varphi) + C_f^{\text{tok}}(V_\psi)])$ 
  & $O(BK(S+T) \cdot [C_f^{\text{tok}}(\pi_{\text{ref}}) + C_f^{\text{tok}}(R_\varphi)])$ 
  & $O(B_{\text{pairs}}(S+T) \cdot [C_f^{\text{tok}}(\pi_{\text{ref}}) + C_f^{\text{tok}}(\pi_\theta)])$ \\

Training 
  & $O(BK(S+T) \cdot [C_{\text{tr}}^{\text{tok}}(\pi_\theta) + C_{\text{tr}}^{\text{tok}}(V_\psi)])$ 
  & $O(BK(S+T) \cdot C_{\text{tr}}^{\text{tok}}(\pi_\theta))$ 
  & $O(B_{\text{pairs}}(S+T) \cdot C_{\text{tr}}^{\text{tok}}(\pi_\theta))$ \\

\midrule
\multicolumn{4}{l}{\textit{\textbf{Depth (Critical Path Latency)}}} \\
Generation 
  & $O(D_\text{gen}(\pi_\theta))$ 
  & $O(D_\text{gen}(\pi_\theta))$ 
  & $-$ \\

Assessment
  & $O(\max\{D_f(\pi_{\text{ref}}),\, D_f(R_\varphi),\, D_f(V_\psi)\})$ \textsuperscript{\ddag} 
  & $O(\max\{D_f(\pi_{\text{ref}}),\, D_f(R_\varphi)\})$ \textsuperscript{\ddag} 
  & $O(\max\{D_f(\pi_\theta),\, D_f(\pi_{\text{ref}})\})$ \\

Training 
  & $O(\max\{D_{\text{tr}}(\pi_\theta),\, D_{\text{tr}}(V_\psi)\})$ 
  & $O(D_{\text{tr}}(\pi_\theta))$ 
  & $O(D_{\text{tr}}(\pi_\theta))$ \\

\midrule
\multicolumn{4}{l}{\textit{\textbf{Memory Cost}}} \\
Generation 
  & $O(|\pi_\theta| + BK \cdot M_{\text{KV}})$ 
  & $O(|\pi_\theta| + BK \cdot M_{\text{KV}})$ 
  & $-$ \\

Assessment 
  & $O(|\pi_{\text{ref}}| + |R_\varphi| + |V_\psi| + BK(S+T)M_{\text{Inf}})$ 
  & $O(|\pi_{\text{ref}}| + |R_\varphi| + BK(S+T)M_{\text{Inf}})$ 
  & $O(|\pi_\theta| + |\pi_{\text{ref}}| + B_{\text{pairs}}(S+T)M_{\text{Act}})$ \\

Training 
  & $O(|\pi_\theta| + |V_\psi| + BK(S+T) \cdot M_{\text{Act}})$ 
  & $O(|\pi_\theta| + BK(S+T) \cdot M_{\text{Act}})$ 
  & $O(|\pi_\theta| + B_{\text{pairs}}(S+T) \cdot M_{\text{Act}})$ \\

\bottomrule
\end{tabular*}
\vspace{-1em}
\caption{
    \textbf{Asymptotic Work, Depth, and Memory analysis of PPO, GRPO, and DPO.} $S$ is prompt length, $T$ is response length. We denote parameter counts by $|\cdot|$.
    \textbf{``tr'' subscript:} For conciseness, for training-stage model invocations, we define $C_{\text{tr}}^{\text{tok}}(M) := C_f^{\text{tok}}(M)+C_b^{\text{tok}}(M)\approx 3C_f^{\text{tok}}(M)$ and $D_{\mathrm{tr}}(M) := D_f(M)+D_b(M)$, because a parameter update requires a forward pass followed by backpropagation. 
    \textbf{Note on Work:} $C_f^{\text{tok}}$ and $C_b^{\text{tok}}$ are the cost of forward and backward passes \underline{per token} and $C_{\text{gen}}^{\text{roll}}$ is the autoregressive generation cost \underline{per rollout}; these are derived in Table~\ref{tab:building-blocks}. Work scales with total tokens $(S+T)$.
    \textbf{Note on Depth:} Generation is depth-bound by $T$ (sequential), while in Assessment/Training there is no outer linear \(T\)-step autoregressive chain.
    \textsuperscript{\dag} DPO is offline; generation occurs prior to the training loop.
    \textsuperscript{\ddag} Max depth assumes parallel (disaggregated) execution; sequential co-located execution is additive.
    $M_{\text{KV}}$ is Key-Value cache memory; $M_{\text{Act}}$ is activation memory (training); $M_{\text{Inf}}$ is inference buffer (assessment). Note that $\pi_{\text{ref}}$ never requires $M_{\text{Act}}$.
}
\label{tab:framework-complexity}
\end{table*}

%% file: intra.tex
\begin{figure*}[t]
    \centering
    \includegraphics[width=1.0\textwidth]{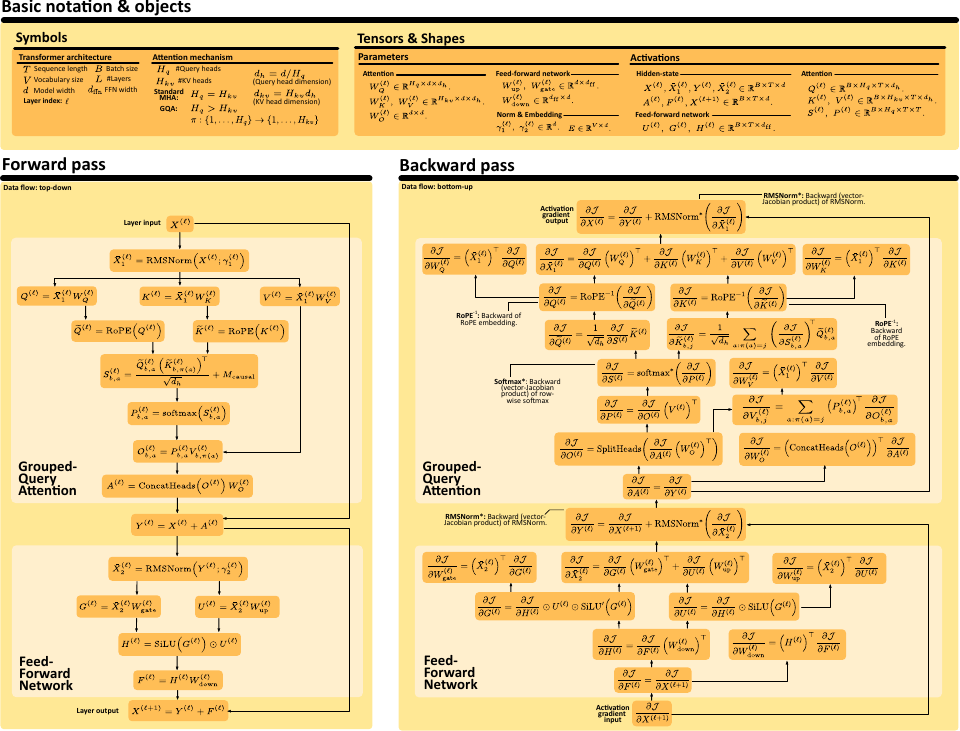}
    \vspace{-1.5em}
    \caption{\textbf{Mathematical notation and equations for the forward and backward passes of each decoder block.} Further details on computational aspects are in Figures~\ref{fig:taxo-intra-inference}-\ref{fig:taxo-intra-training} while parallelization details are in Figures~\ref{fig:taxo-intra-details-dpte}-\ref{fig:taxo-intra-details-scpf}. \textit{No AI was used to conceive or to draw the figure.}}
    \label{fig:taxo-intra-math}
    \vspaceSQ{-1.5em}
\end{figure*}

\begin{figure*}[hbtp]
    \centering
    \includegraphics[width=1.0\textwidth]{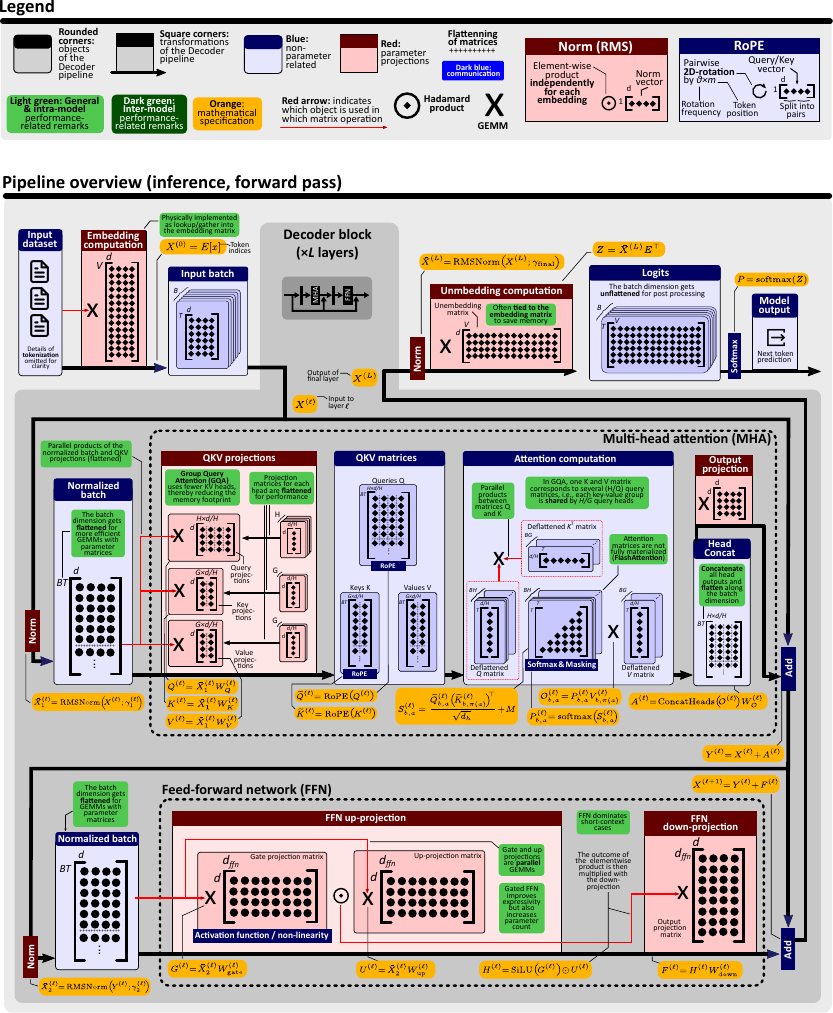}
    \vspace{-1.5em}
    \caption{\textbf{Intra-model execution details (forward pass) for each RLM model execution.} The design details are based on Llama-3. Further details on parallelization and the corresponding taxonomy are provided in Figures~\ref{fig:taxo-intra-details-dpte}-\ref{fig:taxo-intra-details-scpf}. \textit{No AI was used to conceive or to draw the figure.}}
    \label{fig:taxo-intra-inference}
    \vspaceSQ{-1.5em}
\end{figure*}

\begin{figure*}[hbtp]
    \centering
    \includegraphics[width=1.0\textwidth]{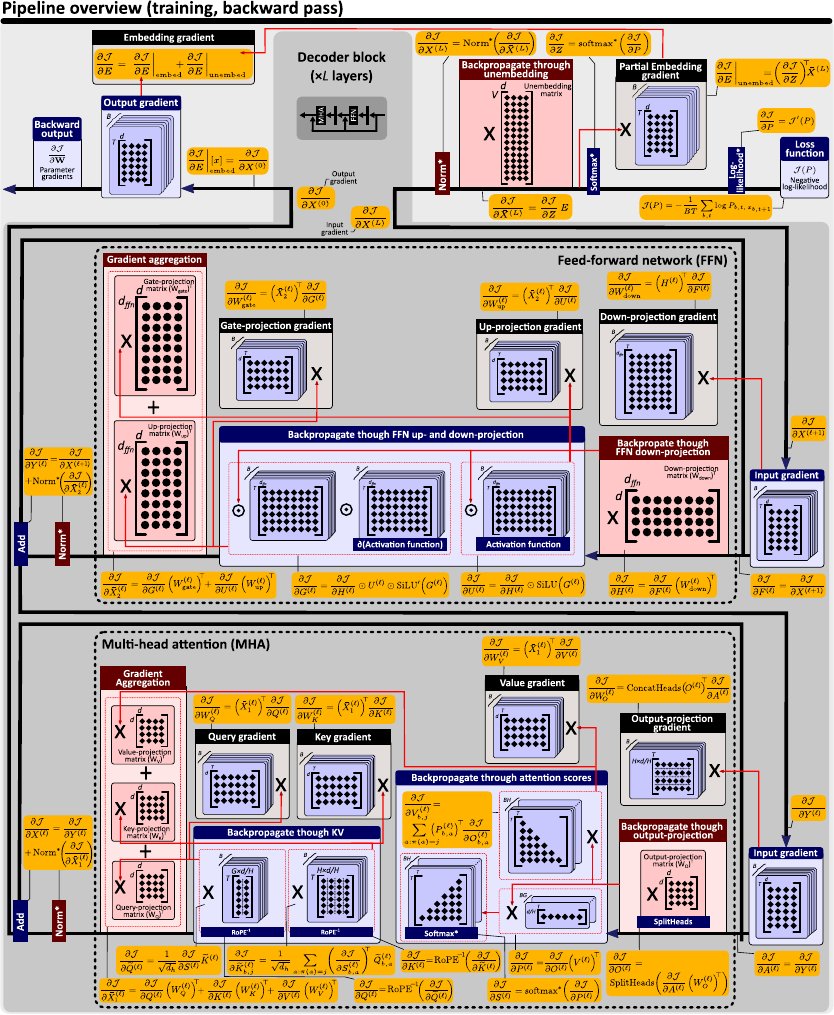}
    \vspace{-1.5em}
    \caption{\textbf{Intra-model execution details (backward pass) for each RLM model execution.} Legend is provided in Figure~\ref{fig:taxo-intra-inference}. The design details are based on Llama-3. Further details on parallelization and the corresponding taxonomy are provided in Figures~\ref{fig:taxo-intra-details-dpte}-\ref{fig:taxo-intra-details-scpf}. \textit{No AI was used to conceive or to draw the figure.}}
    \label{fig:taxo-intra-training}
    \vspaceSQ{-1.5em}
\end{figure*}

\iftr
\begin{figure*}[hbtp]
\vspace{-1.5em}
    \centering
    \includegraphics[width=0.98\textwidth]{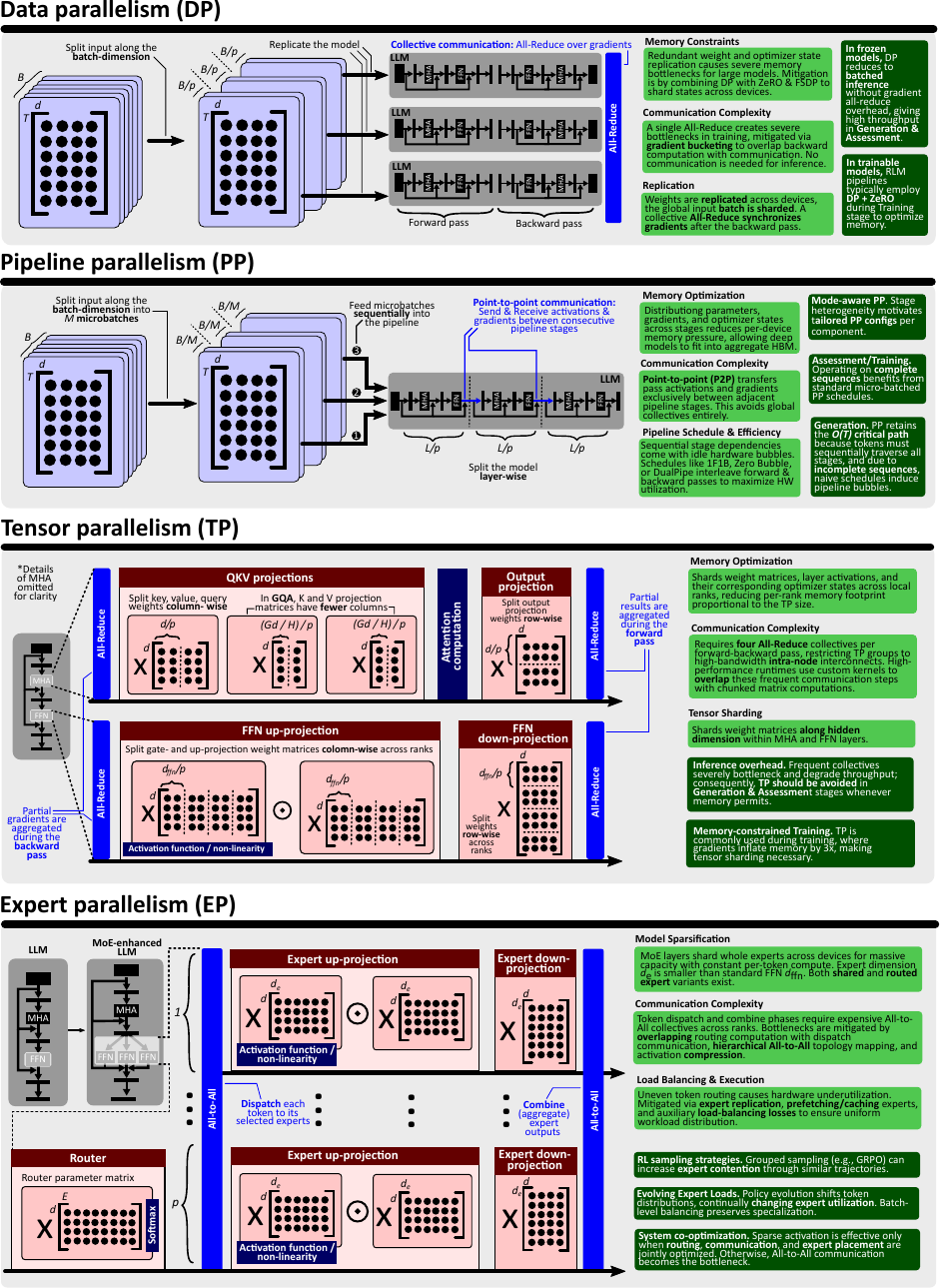}
    \vspace{-0.75em}
    \caption{\textbf{Intra-model parallelism (data, pipeline, tensor, and expert parallelism) details for each RLM model execution; legend is provided in Figure~\ref{fig:taxo-intra-inference}.} \textit{No AI was used to conceive or to draw the figure.}}
    \label{fig:taxo-intra-details-dpte}
\end{figure*}

\begin{figure*}[hbtp]
\vspace{-1.5em}
    \centering
    \includegraphics[width=0.98\textwidth]{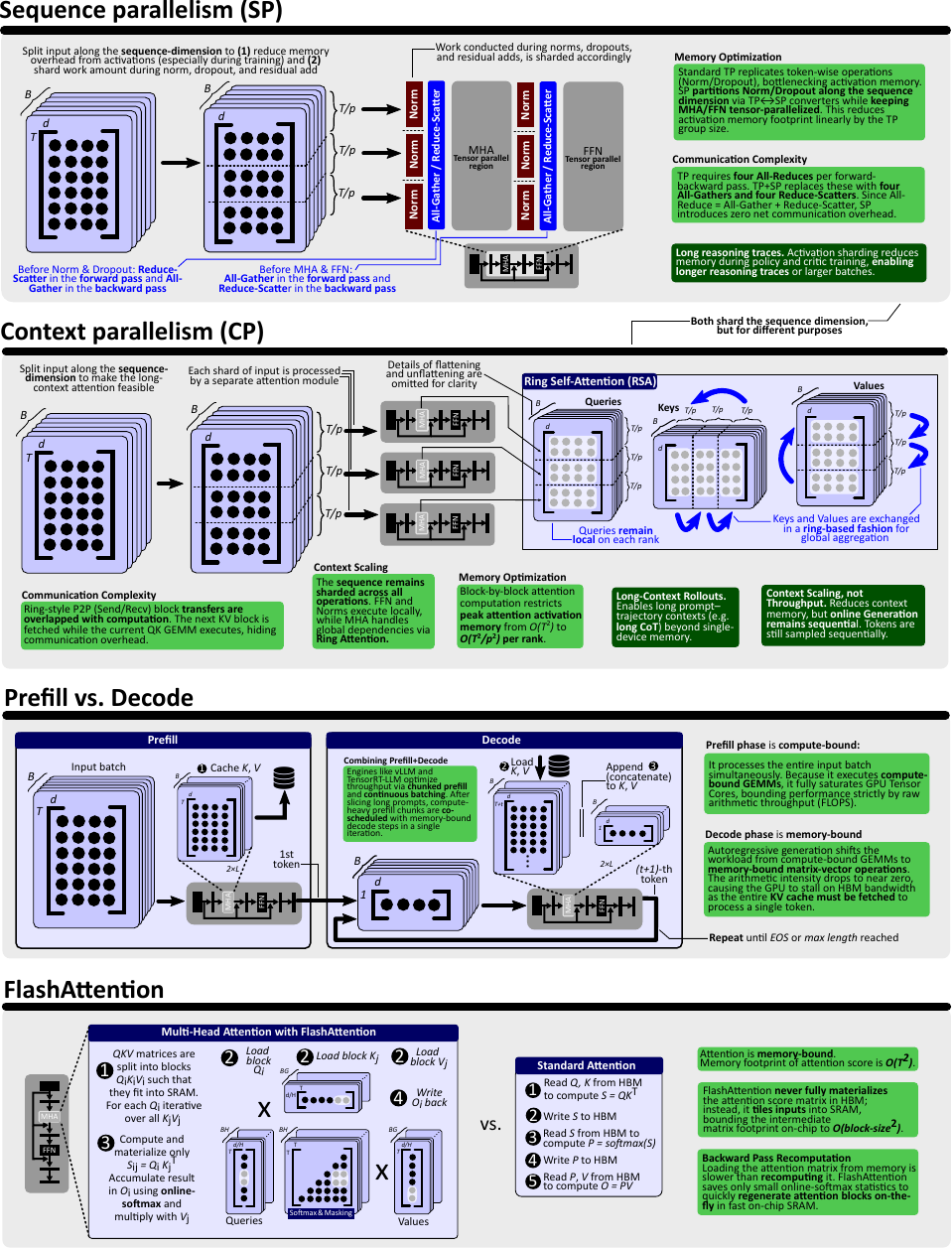}
    \vspace{-0.75em}
    \caption{\textbf{Intra-model parallelism (sequence \& context parallelism, as well as prefill vs.~decode and flash attention) details for each RLM model execution; legend is provided in Figure~\ref{fig:taxo-intra-inference}.} \textit{No AI was used to conceive or to draw the figure.}}
    \label{fig:taxo-intra-details-scpf}
\end{figure*}
\else
\begin{figure*}[hbtp]
\vspace{-1.5em}
    \centering
    \includegraphics[width=0.98\textwidth]{rlms-intra_parallelism_v2.pdf}
    \vspace{-0.75em}
    \caption{\textbf{Intra-model parallelism details for each RLM model execution; legend is provided in Figure~\ref{fig:taxo-intra-inference}.} \textit{No AI was used to conceive or to draw the figure.}}
    \label{fig:taxo-intra-details}
\end{figure*}
\fi

\section{Intra-Model Parallelism}

Efficient training and inference for each individual model in the RL-LLM pipeline is essential. \emph{Intra-model parallelism} denotes established techniques that allow a single model to be trained or served across multiple devices by partitioning its parameters, activations, or computation graph. The main families are: (i) data parallelism, (ii) tensor/operator parallelism, (iii) sequence and context parallelism, (iv) pipeline parallelism, (v) expert parallelism, and (vi) memory-centric optimizations such as optimizer sharding and activation checkpointing. These techniques are typically combined into hybrid schemes (e.g., ZeRO-style sharded data parallelism plus tensor parallelism) to balance memory footprint, communication cost, and compute utilization.
As these techniques are well-known~\cite{li2020pytorch, shoeybi2019megatron, huang2019gpipe, narayanan2021efficient, korthikanti2022reducing, liu2023ring, lepikhin2020gshard}, we summarize them and instead focus on implications for RL--LLM pipelines. 

\ifconf
\else
We detail the computational aspects of the intra-model execution Figure~\ref{fig:taxo-intra-math} (mathematical details and flow diagrams), Figure~\ref{fig:taxo-intra-inference} (forward pass), Figure~\ref{fig:taxo-intra-training} (backward pass),  and in Figures~\ref{fig:taxo-intra-details-dpte}-\ref{fig:taxo-intra-details-scpf} (details on intra-model parallelism and taxonomy).
\fi

\subsection{Data Parallelism}

Data parallelism (DP) replicates the model across $N$ devices and shards the input batch across replicas. Each replica performs a local forward and backward pass on its shard, followed by a gradient synchronization step (typically an all-reduce~\cite{chan2007collective, thakur2005optimization}) to keep parameters identical across devices. This strategy is conceptually simple and scales throughput almost linearly when the model fits into a single device and communication is not a bottleneck.

Naive DP quickly becomes memory-limited for very large models because each device must store a full copy of parameters, gradients, and optimizer state. Hence, DP is most effective when combined with memory-centric optimizations (e.g., ZeRO/FSDP~\cite{rajbhandari2020zero}) so that parameters, gradients, and optimizer state are partially sharded rather than fully replicated.
Moreover, the all-reduce step can become a major source of overhead as the number of devices grows or when interconnect bandwidth is limited. Here, modern systems apply several communication optimizations, such as overlap of computation and communication, gradient bucketing, and gradient accumulation~\cite{li2020pytorch}.

\subsubsection{Implications for RL-LLM Pipelines}

In RL-LLM pipelines, a crucial point for DP is \emph{whether a given model or stage requires gradients or is forward-only}.

\textbf{Forward-only models. }
If a model is not updated in a given stage, DP reduces to independent batched inference with replicated weights and scales almost linearly in the number of devices until limited by input or network I/O. This is the typical regime for the actor during Generation (sampling rollouts), the actor during Assessment when only forward log-probabilities and KL terms are needed, the reward model during Assessment, and the reference model during Assessment.
OpenRLHF, for example, uses Ray and vLLM to run actor, reward, and reference as large-scale batched inference services, separate from training loops~\cite{hu2025openrlhf}. Here, DP is highly effective and typically the first choice.

\textbf{Trainable models. }
Here, DP must all-reduce gradients across workers and thus benefits from sharded data-parallel variants such as ZeRO-2/3 or FSDP. This applies to the actor in the Training stage (PPO/GRPO/DPO updates) and to the critic in actor--critic methods.
Frameworks such as OpenRLHF explicitly apply ZeRO-3/FSDP-style sharded data parallelism to actor and critic training while keeping forward-only models replicated~\cite{hu2025openrlhf,rajbhandari2020zero}. 

In RLM pipelines, a common pattern is therefore: DP + ZeRO for trainable components; pure DP for forward-only components.



\subsection{Tensor Parallelism}

Tensor (or operator) parallelism (TP) partitions individual layers across devices by sharding their parameters along hidden dimensions, so that each device computes only a slice of the layer. Synchronization via all-reduce or all-gather then reconciles partial results.

In Transformer blocks, TP is commonly applied to both {feed-forward networks (FFNs)} and to the multi-head attention (MHA). For FFN, the first linear projection (e.g., $X A$) is split column-wise across devices, and the second projection (e.g., $Y B$) is split row-wise. Each device holds a shard of $A$ or $B$ and computes a partial result. For MHA, query/key/value projections are sharded column-wise, assigning subsets of heads to devices. The output projection is sharded row-wise, mirroring the FFN pattern. TP is typically combined with DP (and sometimes pipeline parallelism) to form 2D, 3D, or 5D parallelism configurations capable of training trillion parameter models~\cite{shoeybi2019megatron,narayanan2021efficient}.

Megatron-LM and related systems implement TP using communication operators that behave differently in forward and backward passes, e.g., an operator $f$ that is identity in the forward pass and all-reduce in the backward, and an operator $g$ that is all-reduce in the forward and identity in the backward~\cite{shoeybi2019megatron}. This enables communication-computation overlap and reduces idle time.

\subsubsection{Implications for RL--LLM pipelines}

Tensor parallelism is essential when a \emph{single} model does not fit on one device even after applying memory optimizations (e.g., ZeRO, mixed precision, quantization). However, for many RLHF workloads, inference dominates runtime. For example, OpenRLHF reports that rollout generation and forward Assessment account for more than 90\% of wall-clock time in RLHF loops~\cite{hu2025openrlhf}. Here, TP can hurt throughput when not strictly necessary. For instance, Megatron-style TP introduces two all-reduces per Transformer block during inference and four during training~\cite{shoeybi2019megatron,narayanan2021efficient}. When a model (actor, reward, or reference) fits on a single device, empirical studies from vLLM and DeepSpeed show that enabling TP reduces total tokens-per-second because collective communication becomes the bottleneck~\cite{vlm_perf,rajbhandari2020zero}. AMD's Qwen2-7B RLHF experiments on MI300X (192\,GB) similarly report that disabling TP gives 2.3--4.3$\times$ higher rollout throughput than when sharding with TP by 2 or 4 for the same model, as communication is negligible in the single-device setting~\cite{amd_qwen2_rlhf}.

The resulting rule-of-thumb for using TP in RLM pipelines is: (1) use TP only when a model does not fit on a single device even with ZeRO/FSDP and mixed precision; (2) prefer no TP for inference-heavy stages (Generation, forward-only Assessment) whenever memory allows; and (3) tolerate TP overhead more readily in training stages, where gradient storage already inflates memory by $\approx 3\times$ and TP may be necessary to fit the model~\cite{shoeybi2019megatron}.
In practice, large RLM deployments often use TP inside high-bandwidth device groups and combine it with pipeline and data parallelism across nodes~\cite{narayanan2021efficient}.

\subsection{Sequence Parallelism}

TP shards heavyweight projections but leaves token-wise operations such as layer normalization and dropout replicated on each TP shard, leading to duplicated activations and higher memory use, especially for long sequences. Sequence parallelism (SP)~\cite{korthikanti2022reducing} addresses this by partitioning activations along the sequence dimension. Because layer norm and dropout operate independently per token, their activations can be split across devices without changing the model's semantics. This reduces per-device activation memory at essentially no additional computation cost.

Sequence parallelism has been adopted in large-scale stacks such as Megatron-LM, DeepSpeed, and ColossalAI for long-context or high-batch training~\cite{korthikanti2022reducing,shoeybi2019megatron,jacobs2023deepspeedulysses,colossalai_sp}, where activation memory is a primary limiter.

\subsubsection{Implications for RL-LLM Pipelines}

Sequence parallelism is primarily beneficial in \emph{training} stages where activations must be stored or recomputed for every token, such as policy/critic updates over long responses and reasoning traces. Sharding activations by sequence length reduces per-GPU memory and allows larger $T$ or batch sizes under fixed memory budgets. Because SP reuses the same collective bandwidth as TP (by switching all-reduce to reduce-scatter/all-gather pairs), its net communication volume stays similar, with modest changes in latency due to additional synchronization points~\cite{korthikanti2022reducing,shoeybi2019megatron}.


\subsection{Context Parallelism}

Context parallelism extends sequence-based partitioning further by maintaining the sequence split throughout the entire Transformer layer, including attention and FFN blocks~\cite{yang2025contextparallel}. Most operations remain token-wise and are naturally compatible; the main challenge is attention, which requires access to keys and values from all sequence partitions. Recent implementations address this using Ring Attention~\cite{liu2023ring}, which pipelines key/value communication in a ring topology to reduce latency and memory pressure.
CP enables training and inference with very long contexts (hundreds of thousands or millions of tokens) by combining tensor, sequence, and context partitioning without replicating activations or weights unnecessarily~\cite{yang2025contextparallel}.

\subsubsection{Implications for RL-LLM Pipelines}

Context parallelism is most useful when RLM training is dominated by long prompt--trajectory contexts (e.g., long CoT traces, tool histories, verifier outputs, etc.). In this regime, the bottleneck is often not only model size but also attention and activation memory over the full context. CP directly targets this axis by partitioning the sequence dimension across devices while preserving exact attention through cross-device KV exchange~\cite{liu2023ring,jacobs2023deepspeedulysses,nvidia2025contextparallelism}.
This yields concrete implications for RLMs. First, CP can make long-CoT rollouts feasible when the actor's context length exceeds single-device memory. Second, in verifier- or tool-augmented RL, CP allows prompts to include larger execution traces, retrieved evidence, and interaction histories without truncating the reasoning state.

However, CP does not remove the autoregressive dependence of online Generation: it reduces per-device context memory and attention work, but tokens are still sampled sequentially. Thus, CP is primarily a feasibility and long-context scaling mechanism; online rollout latency still requires complementary methods such as TP, efficient KV-cache exchange, continuous batching, or stage-level overlap. Recent long-context systems validate this direction: Ring Attention overlaps KV-block communication with blockwise attention, DeepSpeed-Ulysses uses sequence partitioning and all-to-all communication for long-sequence training, and CP-style inference systems report near-linear scaling for million-token prefill workloads~\cite{liu2023ring,jacobs2023deepspeedulysses,yang2025contextparallel}.

\subsection{Pipeline Parallelism}

Pipeline parallelism (PP) partitions the layers of a model into \emph{stages}, each placed on a different device. Mini-batches are further split into \textit{micro-batches} that flow through the pipeline stages. This allows models that are too deep to fit on a single device to be trained, at the cost of pipeline \emph{bubbles} (periods where some stages are idle). PP is often combined with DP and TP, yielding 3D/5D parallelism that supports extremely large models~\cite{narayanan2021efficient}.

\iftr
Numerous pipeline scheduling strategies have been proposed, examples include {AFAB}~\cite{huang2019gpipe} (all-forward-all-backward, all micro-batches complete forward passes before any backward pass starts -- simple but bubble-heavy), {1F1B}~\cite{narayanan2021memory} (forward and backward passes interleave, improving memory efficiency and reducing bubbles), {Interleaved 1F1B}~\cite{narayanan2021efficient} (devices host multiple non-contiguous pipeline stages to further overlap computation and communication), {Zero Bubble}~\cite{qi2023zero} (decomposes the backward pass into $B$-steps (gradients w.r.t.\ activations) and $W$-steps (gradients w.r.t.\ weights) scheduling them separately to eliminate bubbles without increasing peak memory), and {DualPipe}~\cite{liu2025deepseekv3} (runs two pipelines in opposite directions across the same devices, overlapping forward and backward streams to improve utilization).
\fi

\subsubsection{Implications for RL-LLM Pipelines}

The relevance of PP to RL-LLM heavily depends on the considered stage. In Generation, the actor performs autoregressive decoding, so tokens must traverse all pipeline stages sequentially; thus, PP helps fit large actors but does not remove the $O(T)$ critical path, and naive schedules can suffer from large pipeline bubbles and poor utilization~\cite{xiao2025flexrlhf,xu2025piperlhf}. By contrast, Assessment and Training operate on complete sequences and therefore benefit more directly from standard micro-batched PP schedules such as GPipe- or 1F1B-style execution.
This heterogeneity makes \emph{mode-aware} PP especially important in RLHF systems. The best sharding and scheduling strategy for low-latency Generation need not match the best one for high-throughput Training. Recent systems therefore decouple configurations across stages and overlap different RL iterations, e.g., by letting one batch generate while another is being assessed or trained \cite{xu2025piperlhf,hu2025openrlhf,yao2023deepspeed}. Such inter-batch pipelining can substantially reduce idle time. In large reasoning-oriented systems, this design is often combined with explicit communication--computation overlap to reduce pipeline bubbles further~\cite{liu2025deepseekv3}.

PP also interacts with optimization stability. Interleaved or asynchronous schedules may introduce weight staleness, where forward and backward passes for a micro-batch observe different parameter versions. In RLHF, where gradients are already noisy, this can destabilize training unless explicit versioning or consistency mechanisms are used~\cite{xiao2025flexrlhf,xu2025piperlhf}. Finally, PP strengthens the case for disaggregated placement: placing actor, critic, reward, and reference on separate device groups allows each component to use a PP configuration matched to its role, instead of forcing a single compromise configuration for the whole pipeline~\cite{xiao2025flexrlhf,hu2025openrlhf}.

\subsection{Expert Parallelism}

Expert parallelism (EP) is the standard way to scale Mixture-of-Experts (MoE) models by distributing whole experts across devices. Unlike DP, which replicates the full model, or TP, which shards individual operators, EP assigns different experts to different GPUs or nodes. Since only a small subset of experts is activated per token, EP enables very large model capacity with much smaller per-token compute than an equally sized dense model~\cite{dai2024deepseekmoe,liu2025deepseekv3}. Modern MoE designs also separate \emph{shared} and \emph{routed} experts. Shared experts are activated for all tokens and capture common linguistic features, while routed experts specialize in narrower domains such as mathematics or code. In practice, EP is rarely used alone; it is typically combined with TP, PP, and DP in hybrid multi-dimensional layouts~\cite{dai2024deepseekmoe,liu2025deepseekv3}.
 
Formally, an MoE layer contains $E$ experts and a router $G(x)$ that selects the top-$k$ experts for each token representation $x$, with $k \ll E$. The output can be written as $y = \sum_{i=1}^{k} g_i(x) f_i(x)$, where $f_i$ is the transformation of the $i$-th selected expert and $g_i(x)$ is its routing weight. EP exploits this sparsity by storing experts on different devices and executing only the selected ones~\cite{dai2024deepseekmoe}.

EP's main systems cost is communication. Each MoE layer typically requires two all-to-all phases: \emph{dispatch}, which sends token activations to the devices hosting the selected experts, and \emph{combine}, which returns expert outputs to their originating ranks. As expert count and cluster size increase, this communication can dominate runtime, so practical EP deployments rely on hierarchical collectives, locality-aware routing, and communication-computation overlap~\cite{nvidia2026hybridep}.

\subsubsection{Implications for RL-LLM Pipelines}

EP can adapt a very large model while activating and updating only a small subset of parameters per token. This is particularly valuable for long reasoning trajectories, where dense models would make both rollout generation and policy updates prohibitively expensive~\cite{liu2025deepseekv3,dai2024deepseekmoe}.

However, RL post-training makes routing harder. As the policy evolves, the token distribution shifts, so expert loads can become highly imbalanced. Load balancing schemes can help hardware efficiency, but they may also blur expert specialization by forcing artificially uniform routing \cite{guo2025expert_specialization}.
For this reason, recent reasoning-oriented MoE systems increasingly prefer auxiliary-loss-free or bias-based balancing mechanisms that preserve specialization while correcting large load skew at the batch level~\cite{liu2025deepseekv3,guo2025expert_specialization}. More broadly, EP in RLHF requires joint optimization of routing quality, communication cost, and systems balance: if routing is good but overloaded experts are poorly placed, all-to-all communication and straggler effects can erase the theoretical gains from sparse activation \cite{nvidia2026hybridep,nguyen2026llep}.

EP also interacts strongly with RL framework design. Grouped sampling methods such as GRPO may create many structurally similar trajectories for the same prompt, which can stress the same experts simultaneously. Thus, efficient RL-MoE training often requires combining EP with dynamic load-balancing policies~\cite{liu2025deepseekv3,hu2025openrlhf,nguyen2026llep}.

\subsection{Complexity Analysis}
\label{sec:comp-anal-intra}

We now analyze how intra-model parallelism changes the work, depth, and memory of the Transformer invocations inside RL-LLM pipelines; detailed derivations are in Appendix~\ref{sec:app:derivations-intra}. The same local formulas apply to the actor in Generation, the reward/reference/critic models in Assessment, and the policy or critic in Training; the RL-specific consequences come from whether the invocation is autoregressive, forward-only, or trainable.

Generation uses the actor in an autoregressive loop, so reducing the depth of one model invocation helps but cannot remove the outer \(T\)-step dependence. Assessment is teacher-forced and forward-only for frozen models, so it is mainly a batched-inference and memory-placement problem. Training requires backward passes, activations, gradients, and optimizer state, so memory sharding becomes central. Tables~\ref{tab:intra_model_parallelism_local}--\ref{tab:intra_model_parallelism_exact_global} therefore should be read as architecture-independent scaling laws for the individual model calls that compose the RL-LLM loop, not as hardware-calibrated throughput predictions.

A logical model invocation may be distributed across $N$ participating ranks according to one or more parallelism dimensions. Depending on the parallelization strategy, each rank may operate on a partition of the input, parameters, layers, activations, or experts while other components remain replicated. We report both per-rank costs, which characterize local device pressure, and global costs, which aggregate costs across all $N$ ranks participating in the complete logical model invocation. Global work is the sum of arithmetic FLOPs executed across these ranks, including replicated computation on every rank where it occurs. Global memory analogously sums resident state across all participating ranks. Depth denotes the critical path of the complete distributed execution. 

Throughout Tables~\ref{tab:intra_model_parallelism_local}--%
\ref{tab:intra_model_parallelism_exact_global}, we use an idealized arithmetic
work--depth--memory model intended to expose scaling laws rather than predict
hardware runtime. Unless stated otherwise, we exclude communication,
synchronization, kernel-launch and scheduling overheads, finite-device
utilization, load imbalance, and temporary communication workspaces. Pipeline
parallelism assumes an approximately uniform partition of the \(L\) layers and
does not model the number of microbatches, pipeline schedules, fill/drain
bubbles, or inter-stage activation transfers. Training activation memory is
represented by the leading-order term \(B(S+T)Ld\), suppressing constant-factor
storage for Q/K/V tensors, FFN intermediates, normalization 
temporaries, and other implementation-specific buffers; activation
checkpointing is treated separately. The Adam model-state expressions count
parameters, gradients, and first and second moments using a common element-size
abstraction, omitting precision-specific byte factors and transient buffers.
The TP activation terms use an idealized shardable-activation
model; replicated token-wise components are suppressed. Attention is assumed
not to materialize the full attention matrix, as in FlashAttention-style
execution. 
For MoE models, the tables retain the dominant shared-Transformer and expert-FFN
terms; router projection, softmax/top-$k$, auxiliary load-balancing operations,
and their relatively small parameter state are omitted.
Finally, the isolated EP row assumes that only expert FFNs are
partitioned across \(P_e\) ranks while shared dense computation is replicated;
the combined 5D row instead assumes a coupled execution in which EP ranks also
process disjoint token/batch shards for the shared path, so dense arithmetic is
not redundantly executed across the EP dimension.


Each parallelism strategy partitions a different dimension. DP splits the batch, reducing per-group work and activation memory but leaving the critical path unchanged. PP splits layers, reducing per-stage work, memory, and architectural depth by replacing \(L\) with \(L/P_p\), though realized runtime also depends on microbatch bubbles. TP splits hidden-dimensional operators, reducing per-device GEMM work and replacing hidden-dimension depth terms by \(\log(d/P_t)\). CP splits the sequence dimension, reducing context-dependent attention and activation memory, which is crucial for long reasoning traces. EP splits MoE experts, reducing per-device expert memory and compute while leaving dense attention largely unchanged. Finally, 3D/5D parallelism combines data, pipeline, and tensor sharding, giving the strongest per-device memory reduction but also the most communication and scheduling constraints.

For RL-LLMs, this means DP is usually best for throughput-oriented batched Generation or Assessment when models fit per device; TP/PP are needed for large actors, critics, or low-latency single invocations; CP is most useful for long-context reasoning; and 3D/5D/ZeRO-style sharding is most important during Training, where optimizer state and activations dominate memory.

Note that \textbf{pipeline parallelism} does \textit{not} reduce
the \textit{global} depth layer-wise dependency from \(L\) to \(L/P_p\); the \(L/P_p\) factor
appears only in the per-rank depth. This is because, for global end-to-end critical path, the microbatch still passes through all $P_p$ pipeline stages, so it traverses all $L$ layers.

Context parallelism reduces the number of query positions and stored activations per device, but each query still depends on the global key/value context. Therefore, under the work--depth model, CP does not replace the global attention reduction length $S$ by $S/P_c$; its principal benefits are reduced per-rank work and memory. Distributed attention additionally incurs communication-round dependencies, which are outside the present FLOP-based depth abstraction.


\subsubsection{ZeRO \& FSDP}
\label{sec:zero-fsdp}

The memory expressions in Tables~\ref{tab:intra_model_parallelism_local}--\ref{tab:intra_model_parallelism_exact_global} use a simplified Adam model-state
factor consisting of parameters, gradients, and first and second moments. ZeRO~\cite{rajbhandari2020zero}
and FSDP~\cite{zhao2023pytorchfsdp} refine this term without changing the layerwise Transformer
computation. If \(P\) denotes the parameter count of the trainable model and
\(P_z\) the sharding degree, then standard data parallelism stores approximately
\(4P\) model-state elements per replica. ZeRO-1 shards only optimizer states,
giving \(P+P+2P/P_z\); ZeRO-2 shards optimizer states and gradients, giving
\(P+3P/P_z\); and ZeRO-3/FSDP shards parameters, gradients, and optimizer states,
giving approximately \(4P/P_z\), up to transient all-gather buffers. 


\subsubsection{Activation Checkpointing}
\label{sec:ac}

The activation terms in the tables correspond to stored training activations
without checkpointing. Activation checkpointing reduces this term by storing only
a subset of intermediate activations and recomputing the missing ones during
backpropagation. In our notation, this replaces \(M_{\mathrm{Act}}\) by
\(\kappa_{\mathrm{ckpt}}M_{\mathrm{Act}}\) for some
\(0<\kappa_{\mathrm{ckpt}}<1\), while increasing training work and depth by the
extra forward recomputation required during the backward pass. This trade-off is
most relevant for actor and critic training, especially for long reasoning
trajectories.

\input{table-intra-par-complexity}

\input{table-intra-par-exact-complexity}

\subsection{Key Insights \& Takeaways}

The main lesson is \textit{stage-aware hybridization}: no single intra-model strategy is best for the whole RL-LLM loop. Generation,
Assessment, and Training invoke similar Transformer building blocks, but they
stress different dimensions of the system. Generation is autoregressive and
latency-sensitive; Assessment is mostly teacher-forced, forward-only, and
throughput-oriented; Training is memory-intensive as it uses activations, gradients, and optimizer state.

\noindent
\trianbox1{cblue} \textbf{DP scales throughput, not single-sample latency.}
Data parallelism is ideal for increasing the number of prompts, completions, or
preference pairs processed per unit time. In online Generation, it can process
more prompts or candidates in parallel, but each replica still executes the full
autoregressive chain \(D_{\mathrm{gen}}(\pi_\theta;S,T)=O(TD_f(\pi_\theta;S+T))\). Thus, DP does
not shorten the critical path of one rollout trajectory or one model invocation.

\noindent
\trianbox1{cblue} \textbf{TP and PP are capacity/latency tools with communication costs.}
Tensor and pipeline parallelism are useful when a model does not fit on one
device or when the latency of a single invocation must be reduced. In
Generation, they reduce the cost of each actor invocation but do not remove the
outer token-by-token dependence. In Assessment and Training, they help execute
larger reward, reference, actor, or critic models. Their benefit must be
balanced against TP collectives, PP activation transfers, pipeline bubbles, and
resharding overheads.

\noindent
\trianbox1{cblue} \textbf{SP/CP matter increasingly for reasoning.}
Long CoT traces, tool histories, retrieved documents, verifier outputs, and
multi-turn interaction histories increase \(S+T\), making activation and
attention memory central bottlenecks. Sequence and context parallelism directly
target this regime by partitioning the sequence/context dimension. They are
especially useful for long-context Assessment and Training, and for actor
Generation when the rollout context exceeds single-device memory.

\noindent
\trianbox1{cblue} \textbf{Frozen and trainable models prefer different layouts.}
Reward and reference models are usually frozen and forward-only, so they often
scale best as replicated batched-inference services when they fit per device.
The actor and critic are trainable, and therefore dominate memory through
activations, gradients, and optimizer state. Consequently, actor/critic Training
benefits most from ZeRO/FSDP-style sharding, TP/PP, and 3D/5D combinations. PPO is
the most demanding case because it may train both \(\pi_\theta\) and \(V_\psi\),
whereas GRPO and DPO train only \(\pi_\theta\).

\noindent
\trianbox1{cblue} \textbf{EP scales capability but can create systems skew.}
MoE policies can increase reasoning capacity at limited per-token compute by
activating only a subset of experts. However, dense attention remains on the
critical path, and expert routing introduces all-to-all communication. RL
post-training further complicates EP because the evolving policy changes the
token distribution, which can create expert-load imbalance. Efficient EP for
RLMs therefore requires routing-aware load balancing and communication-aware
expert placement.

\noindent
\trianbox1{cblue} \textbf{Practical RLM systems should specialize by model and stage.}
Actor Generation may prioritize KV-cache memory, rollout latency, and batching;
reward/reference Assessment may use cheap replicated inference; actor/critic
Training may require 3D parallelism and optimizer-state sharding. This
intra-model specialization is precisely what motivates the inter-model
placement, stage fusion, hybrid execution, and asynchronous strategies discussed
next.

%% file: table-intra-par-complexity.tex
\newcolumntype{C}[1]{>{\centering\arraybackslash}p{#1}}

\newcommand{\tech}[1]{\textbf{#1}}

\begin{table*}[hbtp]

\centering

\footnotesize

\setlength{\tabcolsep}{1pt}

\renewcommand{\arraystretch}{1.6}

\begin{tabular*}{\textwidth}{@{\extracolsep{\fill}} l c c c @{}}

\toprule

\textbf{Parallelism} & \textbf{Work} & \textbf{Depth} & \textbf{Memory} \\

\midrule

\tech{None\textsuperscript{\dag}}

& $O\!\left(BmL(d^2 + md)\right)$

& $O\!\left(L[\log d+\log m]+\log(Bm)\right)$

& $O\!\left(Ld^2+Vd+BmLd\right)$ \\

\tech{Data}

& $O\!\left(\tfrac{BmL(d^2 + md)}{P_d}\right)$

& $O\!\left(L[\log d+\log m]
+\log\!\left(\tfrac{B}{P_d}m\right)\right)$

& $O\!\left(Ld^2+Vd+\tfrac{BmLd}{P_d}\right)$ \\

\tech{Pipeline$^*$}

& $O\!\left(\tfrac{BmL(d^2 + md)}{P_p}\right)$

& $O\!\left(\tfrac{L}{P_p}[\log d+\log m]
+\log(Bm)\right)$

& $O\!\left(\tfrac{Ld^2}{P_p}+Vd+\tfrac{BmLd}{P_p}\right)$ \\

\tech{Tensor}

& $O\!\left(\tfrac{BmL(d^2 + md)}{P_t}\right)$

& $O\!\left(L\!\left[\log\tfrac{d}{P_t}+\log m\right]
+\log(Bm)\right)$

& $O\!\left(\tfrac{Ld^2+Vd}{P_t}+\tfrac{BmLd}{P_t}\right)$ \\

\tech{Context}

& $O\!\left(\tfrac{BmL(d^2 + md)}{P_c}\right)$

& $O\!\left(L[\log d+\log m]
+\log\!\left(\tfrac{Bm}{P_c}\right)\right)$

& $O\!\left(Ld^2+Vd+\tfrac{BmLd}{P_c}\right)$ \\

\tech{Expert}

& $O\!\left(BmL
\left[(d^2 + md) +\tfrac{E_a d d_e}{P_e}\right]\right)$

& $O\!\left(L[\log d+\log m]+\log(Bm)\right)$

& $O\!\left(L\left[d^2 +\tfrac{E d d_e}{P_e}\right]
+Vd+BmLd\right)$ \\

\tech{3D\textsuperscript{\ddag}}

& $O\!\left(\tfrac{BmL(d^2 + md)}{P_dP_pP_t}\right)$

& $O\!\left(
\tfrac{L}{P_p}
\left[\log\tfrac{d}{P_t}+\log m\right]
+\log\!\left(\tfrac{B}{P_d}m\right)
\right)$

& $O\!\left(
\tfrac{Ld^2+Vd}{P_pP_t}
+\tfrac{BmLd}{P_dP_pP_t}
\right)$ \\

\tech{5D\textsuperscript{\ddag}}

& $O\!\left(
\tfrac{BmL}
{P_dP_eP_cP_pP_t}
\left[d^2+ md + E_a d d_e\right]
\right)$

& $O\!\left(
\tfrac{L}{P_p}
\left[\log\tfrac{d}{P_t}+\log m\right]
+
\log\!\left(
\tfrac{Bm}{P_dP_eP_c}
\right)
\right)$

& $O\!\left(
\tfrac{Ld^2}{P_pP_t}
+
\tfrac{LEdd_e}{P_pP_tP_e}
+
\tfrac{Vd}{P_t}
+
\tfrac{BmLd}{P_dP_eP_cP_pP_t}
\right)$
\\

\bottomrule

\end{tabular*}

\vspace{-0.5em}

\caption{\textbf{Complexity Analysis (Per rank).}
Asymptotic work, depth, and memory for a single Transformer
training iteration (forward and backward pass), expressed in Big-$O$ notation. For clarity, we use $m := S+T$.
The work expressions use the regime where $d_{ff} = O(d)$.
\textsuperscript{\dag}: Baseline configuration without parallelism.
$^*$For PP, the $Vd$ memory term denotes the embedding/output state resident on boundary
pipeline ranks; interior ranks need not store this state. Thus, this term
represents the boundary/peak per-rank footprint rather than replication across
all $P_p$ ranks.
\textsuperscript{\ddag}: 3D and 5D parallelism combine data, pipeline, tensor (3D), context, and expert (5D)
parallelism, with $P_dP_pP_t=N$ (3D) and $P_dP_pP_tP_cP_e=N$ (5D) total devices.
For the combined 5D configuration, we assume the EP dimension also partitions source-token ownership for shared-layer computation.
Training memory assumes Adam without activation checkpointing.}

\label{tab:intra_model_parallelism_local}

\end{table*}

\begin{table*}[hbtp]

\centering

\footnotesize

\setlength{\tabcolsep}{1pt}

\renewcommand{\arraystretch}{1.0}

\begin{tabular*}{\textwidth}{@{\extracolsep{\fill}} l c c c @{}}

\toprule

\textbf{Parallelism} & \textbf{Work} & \textbf{Depth} & \textbf{Memory} \\

\midrule

\tech{None}

& $O\!\left(BmL(d^2 + md)\right)$

& $O\!\left(L[\log d+\log m]+\log(Bm)\right)$

& $O\!\left(Ld^2+Vd+BmLd\right)$ \\

\tech{Data}

& $O\!\left(BmL(d^2 + md)\right)$

& $O\!\left(L[\log d+\log m]+\log(Bm)\right)$

& $O\!\left(P_d[Ld^2+Vd]+BmLd\right)$ \\

\tech{Pipeline}

& $O\!\left(BmL(d^2 + md)\right)$

& $O\!\left(L[\log d+\log m]+\log(Bm)\right)$

& $O\!\left(Ld^2+Vd+BmLd\right)$ \\

\tech{Tensor}

& $O\!\left(BmL(d^2 + md)\right)$

& $O\!\left(L\!\left[\log{d}+\log m\right]
+\log(Bm)\right)$

& $O\!\left(Ld^2+Vd+BmLd\right)$ \\

\tech{Context}

& $O\!\left(BmL(d^2 + md)\right)$

& $O\!\left(L[\log d+\log m]+\log(Bm)\right)$

& $O\!\left(P_c[Ld^2+Vd]+BmLd\right)$ \\

\tech{Expert}

& $O\!\left(
BmL
\left[P_e (d^2 + md)+E_a d d_e\right]
\right)$

& $O\!\left(L[\log d+\log m]+\log(Bm)\right)$

& $O\!\left(
P_e[Ld^2+Vd+BmLd]
+LEdd_e
\right)$ \\

\tech{3D}

& $O\!\left(BmL(d^2 + md)\right)$

& $O\!\left(
L\left[\log{d}+\log m\right]
+\log(Bm)
\right)$

& $O\!\left(P_d[Ld^2+Vd]+BmLd\right)$ \\

\tech{5D}

& $O\!\left(
BmL
\left[d^2+ md +E_a d d_e\right]
\right)$

& $O\!\left(
L\left[\log{d}+\log m\right]
+
\log(Bm)
\right)$

& $O\!\left(
P_dP_cP_e[Ld^2+Vd]
+
P_dP_cLEdd_e
+
BmLd
\right)$
\\

\bottomrule

\end{tabular*}

\vspace{-0.5em}

\caption{\textbf{Complexity Analysis (Global).}
Asymptotic work, depth, and memory for a single Transformer
training iteration (forward \& backward pass). For clarity, we use $m := S+T$.
The work expressions use the regime where $d_{ff} = O(d)$.}

\label{tab:intra_model_parallelism_global}

\end{table*}

%% file: table-intra-par-exact-complexity.tex
\begin{table*}[hbtp]

\centering

\footnotesize

\setlength{\tabcolsep}{1pt}

\renewcommand{\arraystretch}{1.8}

{
\thinmuskip=1.5mu \medmuskip=2mu \thickmuskip=3mu

\begin{tabular*}{\textwidth}{@{\extracolsep{\fill}} l c c c @{}}

\toprule

\textbf{Parallelism}
& \textbf{Work} & \textbf{Depth} & \textbf{Memory} \\

\midrule

\tech{None}

& $6BmL\left(4d^2+2dd_{\text{ff}}+md\right)$

& $2L\log\!\left(\tfrac{d^4d_{\text{ff}}m}{h}\right)
+\log(Bm)$

& $L(16d^2+8dd_{\text{ff}})+4Vd+BmLd$ \\

\tech{Data}

& $6\tfrac{B}{P_d}mL
\left(4d^2+2dd_{\text{ff}}+md\right)$

& $2L\log\!\left(\tfrac{d^4d_{\text{ff}}m}{h}\right)
+\log\!\left(\tfrac{B}{P_d}m\right)$

& $L(16d^2+8dd_{\text{ff}})
+4Vd+\tfrac{B}{P_d}mLd$ \\

\tech{Pipeline}

& $6Bm\tfrac{L}{P_p}
\left(4d^2+2dd_{\text{ff}}+md\right)$

& $2\tfrac{L}{P_p}
\log\!\left(\tfrac{d^4d_{\text{ff}}m}{h}\right)
+\log(Bm)$

& $\tfrac{L}{P_p}(16d^2+8dd_{\text{ff}})
+4Vd+Bm\tfrac{L}{P_p}d$ \\

\tech{Tensor}

& $6BmL
\left(
\tfrac{4d^2+2dd_{\text{ff}}+md}{P_t}
\right)$

& $2L
\log\!\left(
\tfrac{d^4d_{\text{ff}}m}{P_t^2h}
\right)
+\log(Bm)$

& $\tfrac{L(16d^2+8dd_{\text{ff}})}{P_t}
+\tfrac{4Vd}{P_t}
+\tfrac{BmLd}{P_t}$ \\

\tech{Context}

& $6B\tfrac{S+T}{P_c}L
\left(4d^2+2dd_{\text{ff}}+md\right)$

& $2L
\log\!\left(
\tfrac{d^4d_{\text{ff}}m}{h}
\right)
+\log\!\left(\tfrac{Bm}{P_c}\right)$

& $L(16d^2+8dd_{\text{ff}})
+4Vd+B\tfrac{S+T}{P_c}Ld$ \\

\tech{Expert}

& $6BmL
\left(
4d^2+\tfrac{2E_a d d_e}{P_e}+md
\right)$

& $2L
\log\!\left(
\tfrac{d^4d_em}{h}
\right)
+\log(Bm)$

& $L\left(16d^2+\tfrac{8Edd_e}{P_e}\right)
+4Vd+BmLd$ \\

\tech{3D}

& $6\tfrac{B}{P_d}m\tfrac{L}{P_p}
\tfrac{4d^2+2dd_{\text{ff}}+md}{P_t}$

& $2\tfrac{L}{P_p}
\log\!\left(
\tfrac{d^4d_{\text{ff}}m}{P_t^2h}
\right)
+\log\!\left(\tfrac{B}{P_d}m\right)$

& $\tfrac{L(16d^2+8dd_{\text{ff}})}{P_pP_t}
+\tfrac{4Vd}{P_t}
+\tfrac{BmLd}{P_dP_pP_t}$ \\

\tech{5D}

& $6
\tfrac{B}{P_dP_e}
\tfrac{m}{P_c}
\tfrac{L}{P_p}
\left(
\tfrac{
4d^2+2E_a d d_e+md
}{P_t}
\right)$

& $2\tfrac{L}{P_p}
\log\!\left(
\tfrac{d^4d_e m}{P_t^2h}
\right)
+
\log\!\left(
\tfrac{Bm}{P_dP_eP_c}
\right)$

& $\tfrac{16Ld^2}{P_pP_t}
+
\tfrac{8LEdd_e}{P_pP_tP_e}
+
\tfrac{4Vd}{P_t}
+
\tfrac{BmLd}{P_dP_eP_cP_pP_t}$
\\

\bottomrule

\end{tabular*}}

\vspace{-0.5em}

\caption{\textbf{Complexity Analysis (Per rank).}
Explicit leading-order costs under our simplified Transformer cost model,
for a single Transformer training iteration (forward and backward pass),
including explicit constant factors. For clarity, we use $m := S+T$.}

\label{tab:intra_model_parallelism_exact_local}

\end{table*}

\begin{table*}[hbtp]

\centering

\footnotesize

\setlength{\tabcolsep}{1pt}

\renewcommand{\arraystretch}{1.2}

{
\thinmuskip=1.5mu \medmuskip=2mu \thickmuskip=3mu

\begin{tabular*}{\textwidth}{@{\extracolsep{\fill}} l c c c @{}}

\toprule

\textbf{Parallelism}
& \textbf{Work} & \textbf{Depth} & \textbf{Memory} \\

\midrule

\tech{None}

& $6BmL
\left(4d^2+2dd_{\text{ff}}+md\right)$

& $2L
\log\!\left(
{d^4d_{\text{ff}}m}/{h}
\right)
+\log(Bm)$

& $L(16d^2+8dd_{\text{ff}})
+4Vd+BmLd$ \\

\tech{Data}

& $6BmL
\left(4d^2+2dd_{\text{ff}}+md\right)$

& $2L
\log\!\left(
{d^4d_{\text{ff}}m}/{h}
\right)
+\log(Bm)$

& $P_d
\left[L(16d^2+8dd_{\text{ff}})+4Vd\right]
+BmLd$ \\

\tech{Pipeline}

& $6BmL
\left(4d^2+2dd_{\text{ff}}+md\right)$

& $2L
\log\!\left(
{d^4d_{\text{ff}}m}/{h}
\right)
+\log(Bm)$

& $L(16d^2+8dd_{\text{ff}})
+4Vd+BmLd$ \\

\tech{Tensor}

& $6BmL
\left(4d^2+2dd_{\text{ff}}+md\right)$

& $2L
\log\!\left(
{d^4d_{\text{ff}}m}/{h}
\right)
+\log(Bm)$

& $L(16d^2+8dd_{\text{ff}})
+4Vd+BmLd$ \\

\tech{Context}

& $6BmL
\left(4d^2+2dd_{\text{ff}}+md\right)$

& $2L
\log\!\left(
{d^4d_{\text{ff}}m}/{h}
\right)
+\log(Bm)$

& $P_c
\left[L(16d^2+8dd_{\text{ff}})+4Vd\right]
+BmLd$ \\

\tech{Expert}

& $6BmL
\left(
P_e\!\left(4d^2+md\right)
+2E_a d d_e
\right)$

& $2L
\log\!\left(
{d^4d_em}/{h}
\right)
+\log(Bm)$

& $P_e
\left(
16Ld^2+4Vd+BmLd
\right)
+8LEdd_e$ \\

\tech{3D}

& $6BmL
\left(4d^2+2dd_{\text{ff}}+md\right)$

& $2L
\log\!\left(
{d^4d_{\text{ff}}m}/{h}
\right)
+\log(Bm)$

& $P_d
\left[L(16d^2+8dd_{\text{ff}})+4Vd\right]
+BmLd$ \\

\tech{5D}

& $6BmL
\left(
4d^2+2E_a d d_e+md
\right)$

& $2L
\log\!\left(
{d^4d_e m}/{h}
\right)
+
\log(Bm)$

& $P_dP_cP_e
\left(
16Ld^2+4Vd
\right)
+
8P_dP_cLEdd_e
+
BmLd$
\\

\bottomrule

\end{tabular*}}

\vspace{-0.5em}

\caption{\textbf{Complexity Analysis (Global).}
Explicit leading-order costs under our simplified Transformer cost model,
for a single Transformer training iteration (forward and backward pass),
including explicit constant factors. For clarity, we use $m := S+T$.}

\label{tab:intra_model_parallelism_exact_global}

\end{table*}

%% file: inter.tex
\ifconf
\else
\begin{figure*}[hbtp]
    \centering
    \vspace{-1.75em}
    \includegraphics[width=0.95\textwidth]{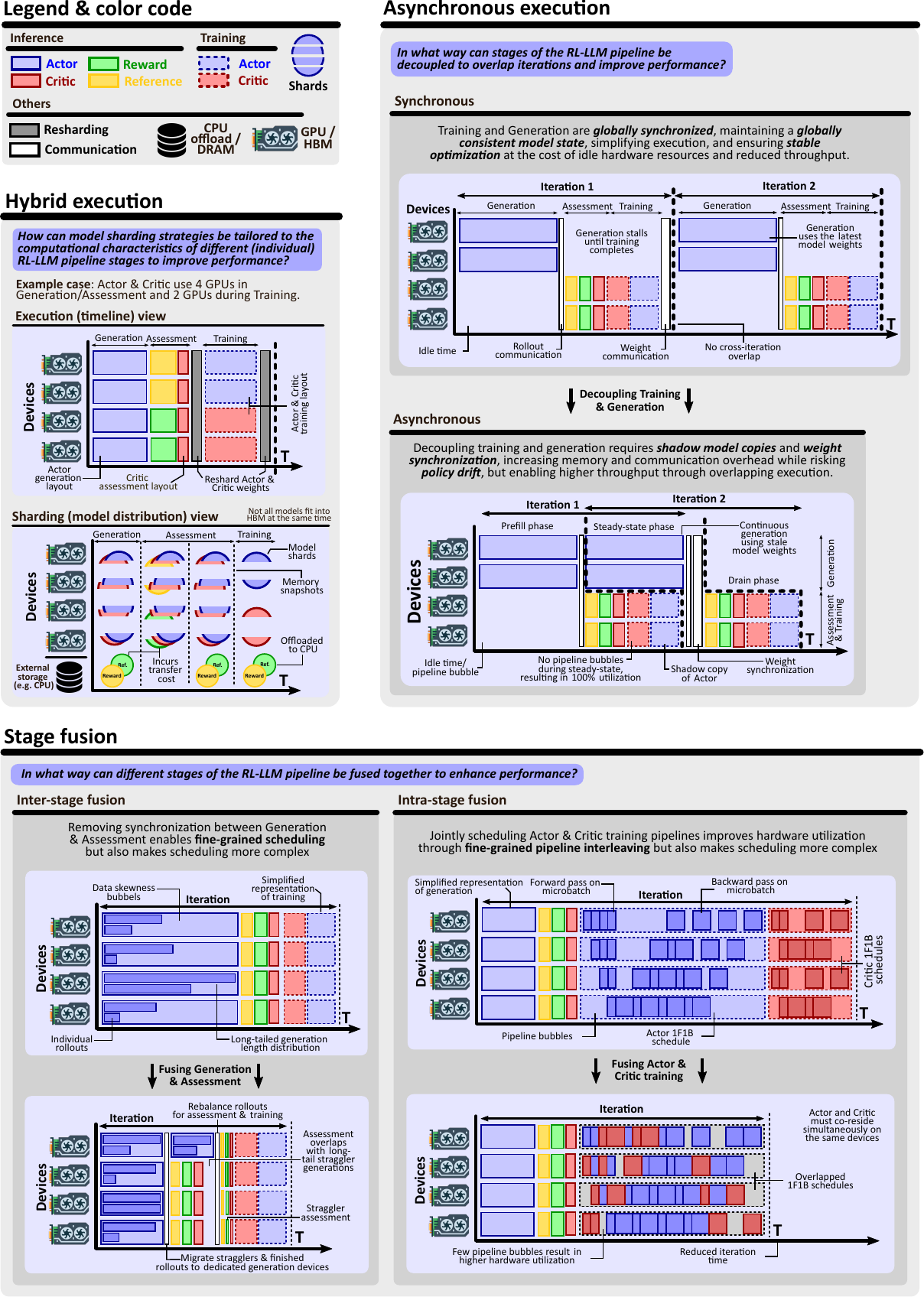}
    \vspace{-0.75em}
    \caption{\textbf{Inter-model parallelism details (asynchronous execution, hybrid execution, stage fusion).} \textit{No AI was used to conceive or to draw the figure.}}
    \label{fig:taxo-inter-async-hybrid-stage}
    \vspaceSQ{-1.5em}
\end{figure*}
\begin{figure*}[hbtp]
    \centering
    \vspace{-1.5em}
    \includegraphics[width=0.7\textwidth]{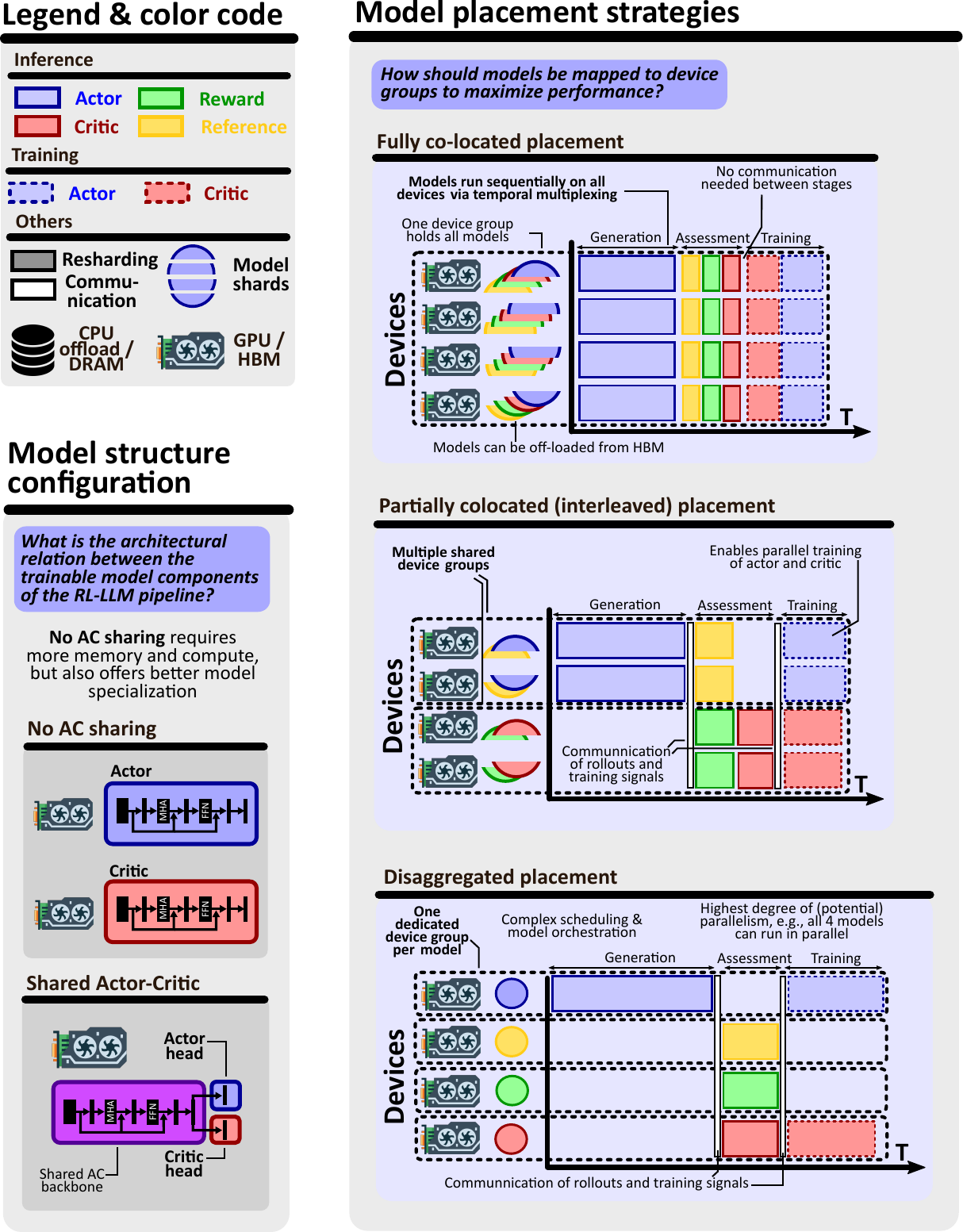}
    \vspace{-0.75em}
    \caption{\textbf{Inter-model parallelism details (placement strategies, model structure configurations).} \textit{No AI was used to conceive or to draw the figure.}}
    \label{fig:taxo-inter-place-struct}
    \vspaceSQ{-1.5em}
\end{figure*}
\fi

\section{Inter-Model Parallelism}


We organize inter-model parallelism into five categories. \textbf{Model structure configurations} decide whether trainable components such as the actor and critic share parameters. \textbf{Model placement strategies} decide whether model roles are co-located, partially co-located, or disaggregated across device groups. \textbf{Stage fusion} removes coarse barriers between stages or sub-stages by streaming partial outputs downstream. \textbf{Hybrid execution} allows different invocations of the same model to use different intra-model layouts, often requiring resharding or automated planning. Finally, \textbf{asynchronous execution} overlaps different RL iterations by allowing bounded-stale parameter reads. These categories are often combined in high-performance systems such as ReaL, RLHFuse, HybridFlow/verl, OpenRLHF, StreamRL, and AReaL~\cite{mei2025real,zhong2024optimizing,sheng2024hybridflow,hu2025openrlhf,zhong2025streamrl,fu2025areal}. 
\ifconf
\else
Figures \ref{fig:taxo-inter-async-hybrid-stage} and \ref{fig:taxo-inter-place-struct} illustrate this taxonomy.
\fi

\subsection{Model Structure Configurations}

Model structure configurations determine the \textit{architectural relation between the trainable model components} in the RL-LLM loop. The central question is whether the actor \(\pi_\theta\) and critic \(V_\psi\) are implemented as independent models, as a shared actor--critic network, or whether the critic is removed by the learning algorithm. PPO-style RLHF uses an actor and a value model, so this choice directly affects work, memory, and optimization stability~\cite{schulman2017proximal, zheng2023secrets}. Critic-free methods such as GRPO remove \(V_\psi\) and replace value-based advantage estimation by group-relative normalization or related baselines, changing both the algorithm and the systems profile~\cite{shao2024deepseekmath, guo2025deepseekr1}.


In a \textbf{shared actor--critic configuration}, a single Transformer backbone \(f_\omega\) produces hidden states that feed two task-specific heads: a policy
head $h_\pi$ for token logits and a value head $h_V$ for scalar value estimates (i.e., $\pi_\theta(\cdot\mid x,y_{<t}) = h_\pi(f_\omega(x,y_{<t}))$ and $V_\psi(x,y_{<t}) = h_V(f_\omega(x,y_{<t}))$). The main systems benefit is that the actor and critic no longer require two separate backbones. This reduces parameter, gradient, optimizer-state, and activation memory, and can replace separate actor/critic forward passes by one shared backbone pass followed by two small heads. In a PPO-style pipeline, this is an inter-model decision: it changes the number of trainable model invocations in Assessment and Training, not merely the internal architecture of one model.

The trade-off is optimization coupling. The policy-gradient loss and value loss both update \(f_\omega\), so their gradients may pull the shared representation in different directions. A single backbone must serve both action selection and return prediction, and a shared optimizer schedule must accommodate losses with different scales and curvature. In practice, this often requires careful value loss weighting, separate learning rates for heads, or partial sharing in which only lower layers are shared and upper layers remain task-specific. Shared actor--critic is therefore most attractive when memory is scarce or when the critic is close in scale to the actor; it is less attractive when value learning requires substantially different representations or optimization dynamics.


With \textbf{independent actor and critic models}, \(\pi_\theta\) and \(V_\psi\) have separate backbones, optimizer states, and hyperparameters. This avoids gradient interference and allows the critic to be sized differently from the actor, for example by using a smaller critic to reduce memory and compute. The cost is a larger model-state footprint and a separate critic forward/backward path. In the work--depth analysis of Section~\ref{sec:comp-anal-frameworks}, this is exactly the additional PPO cost $\Theta\!\left(BK(S+T)[C_f^{\text{tok}}(V_\psi)+C_b^{\text{tok}}(V_\psi)]\right)$ plus the corresponding critic model-state memory.


Finally, \textbf{critic-free methods} move this trade-off from architecture to algorithm design. GRPO removes the learned value model and computes relative advantages from groups of completions sampled for the same prompt, while DPO removes the online actor--critic loop altogether and optimizes over preference pairs.

\subsection{Model Placement Strategies}

Model placement specifies the mapping from model execution to device groups. A device group is a set of GPUs that jointly stores and executes one or more model invocations, possibly using an intra-model strategy internally. The placement decision controls aspects such as memory co-residency, inter-stage communication, and the amount of concurrency available across model roles.


In \textbf{fully co-located placement}, all model roles share one device group and execute through temporal multiplexing. The same pool of GPUs first performs actor generation, then reward/reference/critic inference, and then actor/critic training. This maximizes locality: intermediate tensors can remain within the same device group, and no inter-group communication is needed to pass rollouts, log-probabilities, values, or rewards. It also simplifies orchestration because the runtime schedules one device pool rather than multiple distributed services.

The cost is co-resident memory pressure and poor role specialization. The device group must be provisioned for the largest memory requirement, often Training, yet it is also used for Generation, which has different arithmetic intensity and is often memory-bandwidth or latency dominated. Moreover, ``redundant memory'' in this context means that the same devices must hold the resident state for multiple model roles even when only one role is active. If the system uses data parallelism, this co-residency may be replicated across data-parallel groups: each replica may need actor, critic, reward, and reference state or the ability to materialize them. To mitigate this, one can employ techniques such as offloading, loading and unloading idle models, or weight gathering from sharded states. These techniques reduce memory pressure but introduce context-switching and data-movement overheads. As a result, fully co-located execution is simple and locality-friendly, but it often leaves hardware idle because the stage service times and resource needs are highly imbalanced.


\textbf{Partially co-located, or interleaved}, placement sits between full colocation and full disaggregation. Selected models or stages share a device group;  others are placed separately. For example, actor generation and training may share a group to avoid repeated actor-weight transfers, while reward and reference inference run on smaller inference-oriented groups. Alternatively, reward and reference models may be co-located because both are frozen and forward-only, while the trainable actor and critic are separated.

This design is useful when some model pairs benefit from locality while others benefit from concurrency. It reduces the worst co-residency pressure of full colocation without requiring every stage boundary to become a network boundary. Yet, it also complicates scheduling: groups that share some but not all models must coordinate execution order, memory residency, and data handoff. The best interleaving depends on model sizes, stage service times, cluster topology, and whether a given edge carries small metadata, token IDs, activations, or large parameter shards.


In \textbf{disaggregated placement}, different model roles or stages run on dedicated,
disjoint device groups. The actor group stores and executes the actor, the reward
group stores the reward model, the critic group stores the value model, and the
reference group stores the frozen reference model. This eliminates cross-model
co-residency: a reward-model GPU no longer needs actor or critic weights, and a
training GPU no longer needs to keep frozen reward/reference models resident.
It also enables heterogeneous hardware specialization. Large trainable actors
can be assigned to high-memory GPUs, while smaller frozen reward or reference
models can run on cheaper or inference-optimized devices. StreamRL explicitly
argues for such disaggregation because colocated generation and training couple
resources with very different compute and memory profiles, while disaggregated
stream generation enables flexible resource allocation, heterogeneous training
setups, and even cross-datacenter deployment~\cite{zhong2025streamrl}.

Disaggregation is a prerequisite for many forms of inter-model concurrency, but
it is not by itself a speedup. If the stages still execute strictly
Generation \(\rightarrow\) Assessment \(\rightarrow\) Training, then the
iteration latency remains the sum of stage times and some groups will wait idle.
The benefit appears when disaggregation is combined with stage fusion,
streaming, or asynchronous execution. The price is explicit communication:
rollouts, token IDs, log-probabilities, rewards, values, and sometimes updated
parameters must cross device-group boundaries. Systems such as OpenRLHF,
HybridFlow/verl, ReaL, FlexRLHF, and StreamRL explore different points in this
placement space, using Ray-style orchestration, hierarchical APIs, parameter
reallocation, or streaming data paths to manage the additional complexity~\cite{hu2025openrlhf,sheng2024hybridflow,mei2025real,xiao2025flexrlhf,
zhong2025streamrl}.

\subsection{Stage Fusion}

Stage fusion removes coarse barriers between consecutive stages or between
subtasks within a stage. In the baseline RLHF loop, the Assessment stage waits
until the entire Generation batch has completed, and Training waits until all
assessment outputs are ready. This barrier structure is inefficient for RLMs
because generation lengths are long-tailed: short trajectories may finish much
earlier than the longest ones, yet their reward/reference/value inference is
delayed by stragglers. Stage fusion decomposes stages into finer-grained
subtasks and schedules downstream work as soon as its input is available.

There are two common forms. In \emph{inter-stage fusion}, completed samples are
streamed from Generation into Assessment without waiting for the whole rollout
batch. A short completion can be scored by the reward model and processed by the
reference or critic while the actor is still decoding longer completions. This
turns generation-length skewness into useful overlap. In \emph{intra-stage
fusion}, operators within the same stage interleave microbatches or pipeline
schedules. For example, actor and critic training pipelines may be interleaved so
that one model's bubble time is filled by another model's forward or backward
microbatch.

RLHFuse is a canonical example: it splits generation and assessment into
sample-level subtasks for inter-stage fusion and splits training into
microbatch-level subtasks for intra-stage fusion, enhancing throughput by up to \(3.7\times\)
by mitigating long-tail generation skew and pipeline
bubbles~\cite{zhong2024optimizing}. Related overlap-oriented work such as OPPO
also targets PPO-style serialization and long-tail response lengths by
overlapping pipeline stages~\cite{yan2025oppo}. The key point is that fusion
does not reduce total work; it reduces depth and idle time by changing when
ready subtasks may execute.

Fusion is throughput-oriented and can increase single-sample latency. A sample
that finishes early may wait in a queue until enough samples form an efficient
microbatch, or it may be delayed by backpressure from a downstream stage. Thus,
the steady-state time per batch can improve even if the time from one sample's
generation to its training update increases. Fusion also increases peak memory
because multiple stages are live simultaneously: model weights, KV caches,
activations, queues, and communication buffers may coexist. The design problem is
therefore to choose a granularity that is fine enough to hide skew and bubbles
but not so fine that scheduling overhead, fragmentation, or memory pressure
dominates.

\subsection{Hybrid Parallelism and Adaptive Scheduling}

Hybrid parallelism means that different model invocations in the same
RL-LLM pipeline may use different intra-model strategies, device allocations, or
execution modes. This is distinct from ordinary 3D parallelism inside one model:
the same actor \(\pi_\theta\) may be invoked once for autoregressive generation
and later for training, and these two invocations have different optimal layouts.
Generation favors low-latency inference and efficient KV-cache use while Training favors higher memory
capacity, efficient gradient synchronization, activation checkpointing, or optimizer-state
sharding. A single static layout forces a compromise; hybrid execution allows
each role to use a matched layout.

For example, if the actor is stored in a
ZeRO/FSDP-style sharded training layout, generation may require gathering or
resharding the weights into an inference-centered layout. Conversely, after generation,
the system may need to redistribute weights or optimizer state back into the
training layout. DeepSpeed-Chat's Hybrid Engine is an early example of this
idea, combining training-mode memory optimizations with inference-mode kernel and
parallelism optimizations~\cite{yao2023deepspeed}. ReaL generalizes the idea
as parameter reallocation: an execution plan chooses role-specific allocations
and parallelization strategies, and the runtime redistributes parameters between
them~\cite{mei2025real}. HybridFlow/verl's 3D-HybridEngine similarly targets
efficient actor resharding between training and generation with low redundancy,
while NeMo RL exposes a mode that can train with
pipeline parallelism but run TensorRT-LLM inference in a tensor-parallel
layout~\cite{sheng2024hybridflow,shen2024nemoaligner,nvidia2026nemorl}.

The cost is \textit{resharding}. When two invocations of the same model use incompatible
layouts or different device groups, parameter shards must be communicated and
repartitioned. During the transfer, the system may need extra memory for source
and destination layouts, and the transfer may sit on the critical path unless it
is overlapped with unrelated work. Hybrid execution is profitable only when the
per-stage gains from using specialized layouts exceed the resharding cost.

Pipe-RLHF is an example of computation-mode-aware
parallelism: it 
uses stage-specific parallelization to improve
resource utilization~\cite{xu2025piperlhf}. ReaL and HybridFlow also move in
this direction by representing the RLHF pipeline as a dataflow or execution plan
rather than a fixed sequence of scripts~\cite{mei2025real,sheng2024hybridflow}.
The trade-off is engineering complexity: an adaptive scheduler requires
profiling, cost modeling, memory feasibility checks, and robust orchestration.
It is most valuable at scales where a static hand-designed layout leaves
substantial hardware idle.

\subsection{Asynchronous Execution}

Stage fusion overlaps work within an iteration. Asynchronous execution overlaps
different RL iterations by relaxing cross-iteration parameter freshness. In a
synchronous loop, iteration \(k+1\) cannot generate data until iteration \(k\)
has completed Training and published updated parameters. In an asynchronous
loop, Generation or Assessment may read a bounded-stale snapshot while Training
from the previous iteration has still not completed.


Bounded asynchrony introduces a systems--algorithm trade-off. Systems benefit
because rollout workers and learners no longer wait for one another. This can
substantially improve utilization when generation is long-tailed or much slower
than training. Asynchronous RLHF explicitly studies this off-policy setting and
shows that generation and learning can be decoupled for better efficiency, while
AReaL further develops a fully asynchronous RL system with a
staleness-aware PPO~\cite{noukhovitch2025asynchronous,fu2025areal}.
Laminar pushes the systems side further through trajectory-level asynchrony and
a distributed parameter relay tier, while StaleFlow explicitly coordinates
rollouts under staleness constraints to balance convergence and throughput
\cite{sheng2025laminar,li2026staleflow}.

The algorithmic risk is \textit{policy drift}. A trajectory may have been sampled from a
behavior policy \(\pi_{\mathrm{beh}}\), while the learner updates a newer policy
\(\pi_\theta\). 
%
%
To alleviate this, systems can track statistics such as 
staleness, token-level KL, ratio variance, clip fraction, effective
sample size, and reward/advantage shifts. Standard PPO tolerates small drift
because clipped importance ratios limit the update, but large staleness can move
the trust-region center toward an outdated low-quality policy. AReaL addresses
this by separating the behavior policy used for off-policy correction from the
proximal policy used as the trust-region center.

Asynchrony also changes memory accounting. If generation and training run on
disjoint device groups and must proceed concurrently, generation needs a stable
read-only snapshot while training updates another copy. This creates a
\emph{shadow pair}: a training copy with optimizer state and a generation copy
used for inference. In the worst case, this roughly doubles the parameter memory
for each asynchronously consumed trainable model. The factor is not necessarily
a full \(2\times\) increase in total training memory, because optimizer states
usually remain only on the training side and the shadow copy may be BF16,
quantized, or offloaded; nevertheless, any table or memory model for
asynchronous execution should include an additional persistent inference-side
parameter copy whenever trainable models are disaggregated across stale
readers and writers.

\subsection{Complexity Analysis}
\label{sec:analysis-inter}

\input{table-inter-par-complexity}

We now rigorously quantify the inter-model strategies. 
The analysis is conducted for a PPO-style online RL-LLM pipeline, which is the most demanding common setting; GRPO and DPO
can be recovered by deleting the critic and/or online assessment components.
Detailed derivations are in Appendix~\ref{sec:app:derivations-inter}.

We first define stage-level terms. \textbf{Generation} comes with
\begin{align*}
W_G &= BK\,C_{\mathrm{gen}}(\pi_\theta)& \text{(work)},\\
D_G &= D_{\mathrm{gen}}(\pi_\theta)& \text{(depth)},\\
M_G &= |\pi_\theta|+BK\,M_{\mathrm{KV}}& \text{(memory)}.
\end{align*}

Here \(C_{\mathrm{gen}}\) and \(D_{\mathrm{gen}}\) include the prefill--decode
decomposition from Table~\ref{tab:building-blocks}. 

Next, the work of the \textbf{assessment} stage is
\[
W_A = BK(S+T)\,[C_f^{\text{tok}}(\pi_{\mathrm{ref}})+C_f^{\text{tok}}(R_\phi)+C_f^{\text{tok}}(V_\psi)].
\]
For depth, we distinguish co-located sequential execution
\[
D_A^{\Sigma}
=
D_f(\pi_{\mathrm{ref}})+D_f(R_\phi)+D_f(V_\psi),
\]
from disaggregated execution
\[
D_A^{\max}
=
\max\{D_f(\pi_{\mathrm{ref}}),D_f(R_\phi),D_f(V_\psi)\}.
\]
Similarly, \textbf{training} work is
\[
W_T = BK(S+T)\,[C_{\text{tr}}^{\text{tok}}(\pi_\theta)+C_{\text{tr}}^{\text{tok}}(V_\psi)],
\]
with sequential and disaggregated depths
\[
D_T^\Sigma=D_{\text{tr}}(\pi_\theta)+D_{\text{tr}}(V_\psi),
\quad
D_T^{\max}=\max\{D_{\text{tr}}(\pi_\theta),D_{\text{tr}}(V_\psi)\}.
\]
For memory, we use \(M_A^\Sigma\) for the co-resident assessment footprint,
\(M_T^\Sigma\) for the co-resident actor--critic training footprint, and
\(M_T^{\max}\) for the maximum over disaggregated actor and critic training
groups. When asynchronous execution requires a persistent inference-side copy
of a trainable model, we write this shadow-copy cost as \(M_{\mathrm{sh}}\).

Table~\ref{tab:inter_model_parallelism} summarizes the resulting work, depth,
and peak per-device-group memory. The most important point is that most
inter-model techniques do not reduce total work: they reduce depth by replacing
sequential sums with maxima, or reduce peak memory by eliminating co-residency.
The main exception is shared actor--critic, which also reduces work and model
state by replacing two trainable backbones with one shared backbone.

The baseline has critical path $D_G+D_A^\Sigma+D_T^\Sigma$.
Disaggregation preserves work but changes within-stage composition from sums to
maxima, giving $D_G+D_A^{\max}+D_T^{\max}$.
Stage fusion further overlaps Generation and Assessment, resulting in
$\max\{D_G,D_A^{\max}\}+D_T^{\max}
\approx D_G+D_T^{\max}$
whenever generation dominates assessment. Bounded asynchrony overlaps
generation/assessment of one iteration with training of another, yielding a
steady-state recurrence depth $\max\{D_G+D_A^{\max},D_T^{\max}\}$.
The strongest configuration combines intra-model sharding, disaggregated
placement, stage fusion, and bounded asynchrony. Its idealized depth is $\max\{D_G^{\mathrm{TP}},D_T^{\mathrm{shard}}\}$,
up to any unhidden assessment tail, resharding, and communication overheads.
This expression makes the central limitation explicit: inter-model parallelism
can hide or overlap non-generation work, but it cannot remove the autoregressive
decode chain inside \(D_G\). Reducing that term requires intra-model inference
parallelism, faster decoding kernels, batching, or algorithmic changes that
shorten generated trajectories.

\subsection{Key Insights \& Takeaways}
\label{sec:inter-model-takeaways}

The inter-model analysis yields several lessons.

\noindent
\trianbox1{cblue} \textbf{Most inter-model schemes reduce depth, not work.}
Placement, fusion, hybrid execution, and asynchrony mainly change the critical
path by replacing sequential sums with maxima or by overlapping stages. The total FLOP work remains essentially unchanged. Shared actor--critic is the
main exception because it removes a separate critic backbone and therefore
reduces both work and trainable model-state memory.

\noindent
\trianbox1{cblue} \textbf{Disaggregation is a memory and concurrency enabler, not a standalone speedup.}
Moving actor, critic, reward, and reference models to separate device groups
reduces co-resident memory and turns independent subcomputations into max-depth
terms. However, if the pipeline still executes strict
Generation \(\rightarrow\) Assessment \(\rightarrow\) Training barriers,
disaggregation only moves idle time to different device groups. It must be
combined with fusion, streaming, or asynchrony to reduce iteration depth.

\noindent
\trianbox1{cblue} \textbf{Stage fusion and asynchrony attack different barriers.}
Stage fusion removes within-iteration barriers, especially the barrier between
long-tailed Generation and Assessment. Asynchrony removes cross-iteration
barriers by allowing bounded-stale parameter reads. Fusion gives $\max\{D_G,D_A^{\max}\}+D_T^{\max}$,
while asynchrony gives $\max\{D_G+D_A^{\max},D_T^{\max}\}$.
They can be combined.

\noindent
\trianbox1{cblue} \textbf{The actor remains the dominant bottleneck.}
Even the combined configuration cannot remove the autoregressive decode chain in $D_G=D_{\mathrm{gen}}(\pi_\theta)=O(TD_f(\pi_\theta))$.
Inter-model scheduling can hide reward, reference, critic, and training work
around actor generation, but reducing \(D_G\) itself requires better inference
parallelism (most notably tensor parallelism), batching, KV-cache management, shorter
trajectories, or algorithmic changes.

\noindent
\trianbox1{cblue} \textbf{Asynchrony trades freshness for hardware utilization and memory.}
Bounded staleness can reduce steady-state depth from a sum to a max, but it
introduces policy drift and may require shadow copies.
Thus, the relevant control variables are not only throughput and memory, but also
staleness, KL drift between behavior and current policies, PPO clip fraction, and others.

\noindent
\trianbox1{cblue} \textbf{The best systems are stage- and role-aware.}
A scalable RLM system should not assign one global execution mode to all models.
Actor Generation, reward/reference Assessment, critic inference, actor Training,
and critic Training have different bottlenecks. The strongest designs therefore
combine shared or critic-free structure when appropriate, disaggregated
placement for memory and concurrency, stage fusion for long-tail overlap,
hybrid layouts for stage-specific efficiency, and bounded asynchrony when the
learning rule tolerates stale data.

\noindent
\trianbox1{cblue} \textbf{Shared actor--critic is effective under resource constraints.}
{Shared actor--critic} architectures are particularly effective when device groups and memory are scarce and inter-model concurrency is limited thanks to sharing a single backbone between the policy and
value functions.

%% file: table-inter-par-complexity.tex
\newcommand{\cfg}[1]{\textbf{#1}}

\begin{table*}[hbtp]
\centering
\scriptsize

\setlength{\tabcolsep}{0pt}
\renewcommand{\arraystretch}{1.5}

\begin{tabular*}{\textwidth}{@{\extracolsep{\fill}}
  l @{\hspace{8pt}} l
  >{\centering\arraybackslash}m{5.0cm}
  >{\centering\arraybackslash}m{4.8cm}
  >{\centering\arraybackslash}m{5.4cm}@{}}
\toprule
\textbf{Config} & \textbf{Stage} & \textbf{Global Work} &
\textbf{Global Depth} & \textbf{Peak Memory per-device group} \\
\midrule

\multirow{3}{*}{\cfg{Baseline}}
& Gen.
& $BK\!\cdot\! C_{\text{gen}}(\pi_\theta)$
& $D_{\text{gen}}(\pi_\theta)$
& $|\pi_\theta|+BK\!\cdot\!M_{\text{KV}}$
\\
& Assess.
& \parbox[c]{\linewidth}{%
\centering
$BK(S+T)$\\[1pt]
$\cdot[C_f^{\text{tok}}(\pi_{\text{ref}})
+C_f^{\text{tok}}(R_\varphi)
+C_f^{\text{tok}}(V_\psi)]$
}
& $D_f(\pi_{\text{ref}})+D_f(R_\varphi)+D_f(V_\psi)$
& \parbox[c]{\linewidth}{%
\centering
$|\pi_{\text{ref}}|+|R_\varphi|+|V_\psi|$\\[1pt]
$+ BK(S+T)$\\[1pt]
$\cdot\max\{M_{\mathrm{Inf}}(\pi_{\mathrm{ref}}),M_{\mathrm{Inf}}(R_\varphi),M_{\mathrm{Inf}}(V_\psi)\}$}\vspace{1em}
\\
& Train.
& $BK(S+T)\!\cdot\!
[C_{\text{tr}}^{\text{tok}}(\pi_\theta)
+C_{\text{tr}}^{\text{tok}}(V_\psi)]$
& $D_{\text{tr}}(\pi_\theta)+D_{\text{tr}}(V_\psi)$
& \parbox[c]{\linewidth}{%
\centering
$4\!\cdot\!(|\pi_\theta|+|V_\psi|)$\\[1pt]
$+BK(S+T)(L_\pi d_\pi+L_\psi d_\psi)$
}
\\
\midrule

\multirow{3}{*}{\cfg{AC shared}}
& Gen.
& $BK\!\cdot\!C_{\text{gen}}(\pi_{\text{SH}})$
& $D_{\text{gen}}(\pi_{\text{SH}})$
& $|\pi_{\text{SH}}|+BK\!\cdot\!M_{\text{KV}}$
\\
& Assess.
& \parbox[c]{\linewidth}{%
\centering
$BK(S+T)$\\[1pt]
$\cdot[C_f^{\text{tok}}(\pi_{\text{ref}})
+C_f^{\text{tok}}(R_\varphi)
+C_f^{\text{tok}}(\pi_{\text{SH}})]$
}
& $D_f(\pi_{\text{ref}})
+D_f(R_\varphi)
+D_f(\pi_{\text{SH}})$
& \parbox[c]{\linewidth}{%
\centering
$|\pi_{\text{ref}}|+|R_\varphi|+|\pi_{\text{SH}}|$\\[1pt]
$+ BK(S+T)$\\[1pt]
$\cdot\max\{M_{\mathrm{Inf}}(\pi_{\mathrm{ref}}),M_{\mathrm{Inf}}(R_\varphi),M_{\mathrm{Inf}}(\pi_{\text{SH}})\}$}\vspace{1em}
\\
& Train.
& $BK(S+T)\!\cdot\!
C_{\text{tr}}^{\text{tok}}(\pi_{\text{SH}})$
& $D_{\text{tr}}(\pi_{\text{SH}})$
& $4\!\cdot\!|\pi_{\text{SH}}|
+BK(S+T)L_{\text{SH}}d_{\text{SH}}$
\\
\midrule

\multirow{3}{*}{\cfg{Disagg.}}
& Gen.
& $BK\!\cdot\!C_{\text{gen}}(\pi_\theta)$
& $D_{\text{gen}}(\pi_\theta)$
& $|\pi_\theta|+BK\!\cdot\!M_{\text{KV}}$
\\
& Assess.
& \parbox[c]{\linewidth}{%
\centering
$BK(S+T)$\\[1pt]
$\cdot[C_f^{\text{tok}}(\pi_{\text{ref}})
+C_f^{\text{tok}}(R_\varphi)
+C_f^{\text{tok}}(V_\psi)]$
}
& $\max\{
D_f(\pi_{\text{ref}}),
D_f(R_\varphi),
D_f(V_\psi)\}$
& \parbox[c]{\linewidth}{%
\centering
$\max\{|\pi_{\text{ref}}| + BK(S+T)M_{\mathrm{Inf}}(\pi_{\mathrm{ref}}),$\\[1pt]
$|R_\varphi| + BK(S+T)M_{\mathrm{Inf}}(R_\varphi),$\\[1pt]
$|V_\psi| + BK(S+T)M_{\mathrm{Inf}}(V_\psi)\}$}\vspace{1em}
\\
& Train.
& $BK(S+T)\!\cdot\!
[C_{\text{tr}}^{\text{tok}}(\pi_\theta)
+C_{\text{tr}}^{\text{tok}}(V_\psi)]$
& $\max\{
D_{\text{tr}}(\pi_\theta),
D_{\text{tr}}(V_\psi)\}$
& \parbox[c]{\linewidth}{%
\centering
$\max\{4|\pi_\theta|+BK(S+T)L_\pi d_\pi,$\\[1pt]
$4|V_\psi|+BK(S+T)L_\psi d_\psi\}$
}
\\
\midrule

\multirow{3}{*}{\cfg{Hybrid}}
& Gen.
& $BK\!\cdot\!C_{\text{gen}}(\pi_\theta)$
& $D_{\text{gen}}^{\text{TP}}(\pi_\theta)$
& $|\pi_\theta|+BK\!\cdot\!M_{\text{KV}}$
\\
& Assess.
& \parbox[c]{\linewidth}{%
\centering
$BK(S+T)$\\[1pt]
$\cdot[C_f^{\text{tok}}(\pi_{\text{ref}})
+C_f^{\text{tok}}(R_\varphi)
+C_f^{\text{tok}}(V_\psi)]$
}
& $D_f^{\text{TP}}(\pi_{\text{ref}})
+D_f^{\text{TP}}(R_\varphi)
+D_f^{\text{TP}}(V_\psi)$
& \parbox[c]{\linewidth}{%
\centering
$|\pi_{\text{ref}}|+|R_\varphi|+|V_{\psi}|$\\[1pt]
$+ BK(S+T)$\\[1pt]
$\cdot\max\{M_{\mathrm{Inf}}(\pi_{\mathrm{ref}}),M_{\mathrm{Inf}}(R_\varphi),M_{\mathrm{Inf}}(V_\psi)\}$}\vspace{1em}
\\
& Train.
& $BK(S+T)\!\cdot\!
[C_{\text{tr}}^{\text{tok}}(\pi_\theta)
+C_{\text{tr}}^{\text{tok}}(V_\psi)]$
& $D_{\text{tr}}(\pi_\theta)+D_{\text{tr}}(V_\psi)$
& \parbox[c]{\linewidth}{%
\centering
$4\!\cdot\!(|\pi_\theta|+|V_\psi|)$\\[1pt]
$+BK(S+T)(L_\pi d_\pi+L_\psi d_\psi)$
}
\\
\midrule

\multirow{2}{*}{\cfg{Stage F.}}
& Inter
& \parbox[c]{\linewidth}{%
\centering
$BK\,C_{\text{gen}}(\pi_\theta)+BK(S+T)$\\[1pt]
$\cdot[C_f^{\text{tok}}(\pi_{\text{ref}})
+C_f^{\text{tok}}(R_\varphi)
+C_f^{\text{tok}}(V_\psi)]$
}
& \parbox[c]{\linewidth}{%
\centering
$\max\{D_{\text{gen}}(\pi_\theta),
D_f(\pi_{\text{ref}}),$\\[1pt]
$D_f(R_\varphi),
D_f(V_\psi)\}$
}
& $\max\{
|\pi_\theta|+BK M_{\text{KV}},\,
|\pi_{\text{ref}}|+|R_\varphi|+|V_\psi|\}$
\\
& Intra
& $BK(S+T)\!\cdot\!
[C_{\text{tr}}^{\text{tok}}(\pi_\theta)
+C_{\text{tr}}^{\text{tok}}(V_\psi)]$
& $\max\{
D_{\text{tr}}(\pi_\theta),
D_{\text{tr}}(V_\psi)\}$
& \parbox[c]{\linewidth}{%
\centering
$4\!\cdot\!(|\pi_\theta|+|V_\psi|)$\\[1pt]
$+BK(S+T)(L_\pi d_\pi+L_\psi d_\psi)$
}
\\
\midrule

\cfg{Async}
& Aggr.
& \parbox[c]{\linewidth}{%
\centering
$BK\,C_{\text{gen}}(\pi_\theta)+BK(S+T)$\\[1pt]
$\cdot[C_f^{\text{tok}}(\pi_{\text{ref}})
+C_f^{\text{tok}}(R_\varphi)
+C_f^{\text{tok}}(V_\psi)$\\[1pt]
$\quad+C_{\text{tr}}^{\text{tok}}(\pi_\theta)
+C_{\text{tr}}^{\text{tok}}(V_\psi)]$
}
& \parbox[c]{\linewidth}{%
\centering
$\max\big\{
D_{\text{gen}}(\pi_\theta)
+\max\{D_f(\pi_{\text{ref}}),
D_f(R_\varphi),D_f(V_\psi)\},$\\[1pt]
$\max\{D_{\text{tr}}(\pi_\theta),
D_{\text{tr}}(V_\psi)\}
\big\}$
}
& \parbox[c]{\linewidth}{%
\centering
$\max\{
|\pi_\theta|+BK M_{\text{KV}}+M_{\text{sh}}(\pi_\theta),$\\[1pt]
$|\pi_{\text{ref}}|,
|R_\varphi|,
|V_\psi|,$\\[1pt]
$4|\pi_\theta|+BK(S+T)L_\pi d_\pi,$\\[1pt]
$4|V_\psi|+BK(S+T)L_\psi d_\psi
\}$
}
\\
\midrule

\cfg{Combined}
& Aggr.
& \parbox[c]{\linewidth}{%
\centering
$BK\,C_{\text{gen}}(\pi_\theta)+BK(S+T)$\\[1pt]
$\cdot[C_f^{\text{tok}}(\pi_{\text{ref}})
+C_f^{\text{tok}}(R_\varphi)
+C_f^{\text{tok}}(V_\psi)$\\[1pt]
$\quad+C_{\text{tr}}^{\text{tok}}(\pi_\theta)
+C_{\text{tr}}^{\text{tok}}(V_\psi)]$
}
& \parbox[c]{\linewidth}{%
\centering
$\max\Big\{
\max\{
D_{\text{gen}}^{\text{TP}}(\pi_\theta),
D_f(\pi_{\text{ref}}),$\\[1pt]
$\qquad\quad
D_f(R_\varphi),
D_f(V_\psi)\},$\\[1pt]
$\max\{
D_{\text{tr}}(\pi_\theta),
D_{\text{tr}}(V_\psi)\}
\Big\}$
}
& \parbox[c]{\linewidth}{%
\centering
$\max\{
|\pi_\theta|+BK M_{\text{KV}}+M_{\text{sh}}(\pi_\theta),$\\[1pt]
$|\pi_{\text{ref}}|,
|R_\varphi|,
|V_\psi|,$\\[1pt]
$4|\pi_\theta|+BK(S+T)L_\pi d_\pi,$\\[1pt]
$4|V_\psi|+BK(S+T)L_\psi d_\psi
\}$
}
\\

\bottomrule
\end{tabular*}

\vspace{-0.5em}

\caption{
\textbf{Inter-model parallelism.}
Global work, global depth, and peak memory per device group reported for the
Generation, Assessment, and Training stages of a single RL iteration within
one epoch under different inter-model parallelism techniques.
Global work and depth report total FLOPs and the critical-path length per stage.
Peak memory is measured per device group and taken as the maximum over all
groups active within a stage; in configurations without disaggregation,
all models share the same device group and execute through temporal
multiplexing, where different models become active at different stages.
\textbf{``TP'' superscript \& ``SH'' subscript:}
A $\mathrm{TP}$ superscript denotes tensor-parallel execution;
$\pi_{\mathrm{SH}}$ denotes the shared actor--critic configuration.
\textbf{``tr'' subscript:}
For conciseness, for training-stage model invocations, we define
$C_{\text{tr}}^{\text{tok}}(M)
:=C_f^{\text{tok}}(M)+C_b^{\text{tok}}(M)
\approx3C_f^{\text{tok}}(M)$
and
$D_{\text{tr}}(M):=D_f(M)+D_b(M)$,
because a parameter update requires a forward pass followed by backpropagation.
$M_{\text{sh}}(\pi_\theta)$ denotes the persistent inference-side shadow copy
of the actor required when asynchronous consumers use a bounded-stale snapshot.
Also, sample-time actor log-probabilities are assumed to be computed on the fly
during Generation from the same logits used for sampling; therefore they do not
add a separate actor forward pass.
\textbf{Baselines:}
The \emph{Baseline} configuration sequentially (i.e., one device group)
executes all RL components, resulting in additive depth terms.
\emph{AC shared} uses a shared Transformer backbone for actor and critic.
All models run sequentially in a single device group.
\emph{Disaggregated} execution assigns models to distinct device groups,
allowing concurrent execution.
The \emph{Hybrid} configuration applies TP to Generation, TP/PP to Assessment,
and ZeRO-3 to Training. All models share the same device group.
\emph{Stage F.} applies inter-stage fusion to Generation and Assessment and
intra-stage fusion to Training.
\emph{Asynchronous execution} overlaps Generation and Assessment with Training
by allowing bounded weight staleness.
\emph{Combined execution} combines asynchronous overlap with disaggregated model
placement and hybrid intra-model parallelism, following the RLHFuse execution
model.
Training memory assumes the Adam optimizer without activation checkpointing,
accounting for parameters, gradients, optimizer states, and activations.
}

\label{tab:inter_model_parallelism}

\end{table*}

%% file: appendix-schemes.tex
\section{Parallelism-Focused Specifications}
\label{sec:alg-specs}

\definecolor{parmark}{RGB}{150,150,150}
\definecolor{laneActor}{RGB}{230,240,255}      
\definecolor{laneCritic}{RGB}{255,235,235}     
\definecolor{laneReward}{RGB}{225,245,225}     
\definecolor{laneReference}{RGB}{240,230,250}  
\definecolor{laneAnnot}{RGB}{28,95,115}        
\newcommand{\parannot}[1]{\textcolor{laneAnnot}{\textbf{[#1]}}}

\newcommand{\parlabel}{%
  {\textcolor{blue}{\textbf{[inter-model concurrency]}}}%
}

\newcommand{\fillbox}[2]{%
  \begingroup
    \setlength{\fboxrule}{0pt}%
    \setlength{\fboxsep}{0.55em}%
    \fcolorbox{#1}{#1}{%
      \parbox{\dimexpr\hsize-2\fboxsep\relax}{\raggedright #2}%
    }%
  \endgroup
}

\newcommand{\lanebox}[3]{%
  \fillbox{#1}{%
    {\scriptsize\itshape #2}\par\vspace{0.2ex}%
    #3%
  }%
}

\newcommand{\parpair}[4]{%
  \parlabel\par\vspace{0.35ex}%
  \lanebox{laneActor}{#1}{#2}\par\vspace{0.30ex}%
  \lanebox{laneCritic}{#3}{#4}%
}

\newcommand{\partriple}[6]{%
  \parlabel\par\vspace{0.35ex}%
  \lanebox{laneReward}{#1}{#2}\par\vspace{0.30ex}%
  \lanebox{laneCritic}{#3}{#4}\par\vspace{0.30ex}%
  \lanebox{laneReference}{#5}{#6}%
}

\newcommand{\parpairAC}[4]{%
  \parlabel\par\vspace{0.35ex}%
  \lanebox{laneReward}{#1}{#2}\par\vspace{0.30ex}%
  \lanebox{laneReference}{#3}{#4}%
}

\newcommand{\parpairARef}[4]{%
  \parlabel\par\vspace{0.35ex}%
  \lanebox{laneActor}{#1}{#2}\par\vspace{0.30ex}%
  \lanebox{laneReference}{#3}{#4}%
}

We present a unified algorithmic and mathematical formulation of PPO
(Algorithm~\ref{alg:ppo}) and GRPO (Algorithm~\ref{alg:grpo}), which share a
common rollout-generation routine (Algorithm~\ref{alg:generation-common}),
together with the offline counterpart DPO (Algorithm~\ref{alg:dpo}). The goal is to expose where the RL-LLM
pipeline admits intra- and inter-model parallelism, and this to facilitate the development of more efficient RLM architectures.

We use a notation in which boldface denotes a per-token vector and a non-bold letter with a $t$ subscript denotes one of its per-token components. For example, $\boldsymbol{V}_n \gets V_\psi(\boldsymbol{x}_n, \boldsymbol{y}_n)$ abbreviates $(V_{n,t})_{t=1}^{|\boldsymbol{y}_n|} \gets (V_\psi(\boldsymbol{x}_n, \boldsymbol{y}_{n,<t}))_{t=1}^{|\boldsymbol{y}_n|}$, with all $|\boldsymbol{y}_n|$ entries produced by a single teacher-forced forward pass. A plain letter without a $t$ subscript is a scalar (e.g., $r_n \gets R_\varphi(\boldsymbol{x}_n, \boldsymbol{y}_n)$). The unit basis vector $\boldsymbol{e}_t$ has $1$ at position $t$ and $0$ elsewhere. The all-ones vector $\boldsymbol{1}$ has $1$ at every position, with length inferred from context.

If the actor and critic share a backbone, the PPO
actor/critic two invocations should instead be read as a single shared-model
invocation with two heads. For DPO, the reference path may be omitted from the online training loop if
\(\boldsymbol{\ell}_{\mathrm{ref},w}^{(n)}\) and
\(\boldsymbol{\ell}_{\mathrm{ref},l}^{(n)}\) are precomputed and stored with the
preference dataset.

Background colors indicate model roles. The label
\textcolor{blue}{\textbf{[inter-model concurrency]}} marks following fork--join (colored) regions in which distinct
model invocations have no data dependency and may run concurrently when placed
on separate device groups. The label \textcolor{blue}{\textbf{[DP]}} (data parallelism) marks independent
sample-level work, such as prompts, completions, rollouts, or preference pairs.
The tags \textcolor{blue}{\textbf{[TP]}}, \textcolor{blue}{\textbf{[PP]}}, \textcolor{blue}{\textbf{[SP]}}, \textcolor{blue}{\textbf{[CP]}}, and
\textcolor{blue}{\textbf{[EP]}} indicate intra-model parallel strategies that can be used for the
corresponding invocation. Boldface (e.g., \textcolor{blue}{\textbf{TP}}) indicates a usual use, non-bold font (e.g., \textcolor{blue}{SP}) is a possible but not a necessarily common use case. Each tag describes the next following line or code block.

TP/PP are mainly capacity or single-invocation latency
tools, SP is mainly useful for trainable teacher-forced passes with large
activation memory, CP is useful for long-context invocations, and EP applies
only to MoE models.

\begin{algorithm}[h]
\caption{A common part of Generation, used by PPO and GRPO}
\label{alg:generation-common}
\KwIn{Batch of prompts $X = \{\boldsymbol{x}_b\}_{b=1}^B$, candidates/prompt $K$.}
\KwOut{Set $\mathcal{B}_{\mathrm{roll}}$ of generated samples paired with per-token rollout log-probabilities.}
\BlankLine
Initialize $\mathcal{B}_{\mathrm{roll}} \gets \emptyset$\;

\textcolor{blue}{\textbf{[DP]}} \ForEach{prompt $\boldsymbol{x}_b \in X$}{
    \textcolor{blue}{\textbf{[DP]}} \For{candidate $i = 1$ \KwTo $K$}{
        Initialize $\boldsymbol{y}_b^{(i)} \gets []$ and $\boldsymbol{\ell}_{\mathrm{old},b}^{(i)} \gets []$\;
        \lanebox{laneActor}{actor path}{%
        \For{$t = 1, 2, \dots$ until EOS or max length}{
            \textcolor{blue}{\textbf{[TP, PP, \textmd{SP,} CP, EP]}} \\ Sample next token $y_{b,t}^{(i)} \sim \pi_\theta\left(\cdot \mid \boldsymbol{x}_b,\boldsymbol{y}_{b,<t}^{(i)}\right)$ via stochastic decoding\;
            Actor log-prob $\ell_{\mathrm{old},b,t}^{(i)} \gets \log\pi_\theta\left(y_{b,t}^{(i)} \mid \boldsymbol{x}_b, \boldsymbol{y}_{b,<t}^{(i)}\right)$\;
            Append $y_{b,t}^{(i)}$ to $\boldsymbol{y}_b^{(i)}$ and $\ell_{\mathrm{old},b,t}^{(i)}$ to $\boldsymbol{\ell}_{\mathrm{old},b}^{(i)}$\;
        }%
        }
        Append $\left(\boldsymbol{x}_b, \boldsymbol{y}_b^{(i)}, \boldsymbol{\ell}_{\mathrm{old},b}^{(i)}\right)$ to $\mathcal{B}_{\mathrm{roll}}$\;
    }
}
\end{algorithm}

\input{code-ppo}
\input{code-grpo}
\input{code-dpo}

The intra-model annotations are attached only to operations that invoke a
Transformer model or backpropagate through one. For independent samples, prompts,
candidates, rollouts, or preference pairs, \textbf{DP} is the natural default
because these units have no semantic dependency and can be assigned to different
replicas with only later reductions for losses, statistics, or gradients. For
model invocations, TP and PP are marked whenever the actor, critic, reward, or
reference model may need to be sharded to fit memory or reduce single-invocation
latency; however, they are conditional because their collectives, activation
transfers, and pipeline bubbles can hurt throughput when the model already fits
on one device. CP is marked on long-context forward or training invocations,
including autoregressive decode, because KV/attention memory of RLM rollouts may exceed
single-device capacity. SP is marked only on trainable teacher-forced
forward/backward paths, not on ordinary forward-only Assessment or decode-time
Generation, because its main benefit is reducing stored activation memory during
Training. EP is marked only as an MoE-dependent option: it applies if the
corresponding actor, critic, reward, or reference model contains routed experts,
where it reduces per-device expert storage and compute but introduces routing and
all-to-all communication.

%% file: code-ppo.tex

\begin{algorithm}[hbtp]
\caption{Algorithmic and Mathematical Specification of PPO}
\label{alg:ppo}
\KwIn{
GAE discount $\gamma$ and smoothing $\lambda$, KL coefficient $\beta$, PPO clip range $\epsilon$, entropy coefficient $c_H$;
set of prompts $X$, prompt batch size $B$, candidates/prompt $K$, training epochs $E_{\mathrm{PPO}}$.
}
\KwOut{Trained parameters $(\theta,\psi)$.}
\BlankLine

\textcolor{blue}{\textbf{[DP]}} \ForEach{prompt batch $\{\boldsymbol{x}_b\}_{b=1}^{B}\subset X$}{
    \BlankLine
    \textbf{——————————— Generation (Online) ————————}

    Generate completions $\mathcal{B}_{\mathrm{roll}} \gets \left\{\left(\boldsymbol{x}_b, \boldsymbol{y}_b^{(i)}, \boldsymbol{\ell}_{\mathrm{old},b}^{(i)}\right)\right\}_{b=1,i=1}^{B,K}$ via \textbf{Algorithm~\ref{alg:generation-common}}\;

    Flatten $\mathcal{B}_{\mathrm{roll}} \gets \left\{\left(\boldsymbol{x}_n, \boldsymbol{y}_n, \boldsymbol{\ell}_{\mathrm{old}}^{(n)}\right)\right\}_{n=1}^{BK}$\;

    \BlankLine
    \textbf{——————————— Assessment ————————————}

    \textcolor{blue}{\textbf{[DP]}} \ForEach{$\left(\boldsymbol{x}_n, \boldsymbol{y}_n, \boldsymbol{\ell}_{\mathrm{old}}^{(n)}\right) \in \mathcal{B}_{\mathrm{roll}}$}{
        \partriple{reward path}{
        \textcolor{blue}{\textbf{[TP, PP, \textmd{SP,} CP, EP]}} \\ Compute scalar reward $s_n \gets R_\varphi\left(\boldsymbol{x}_n, \boldsymbol{y}_n\right)$\;
        }{critic path}{
        \textcolor{blue}{\textbf{[TP, PP, \textmd{SP,} CP, EP]}} \\ Compute values $\boldsymbol{V}_n \gets V_\psi(\boldsymbol{x}_n, \boldsymbol{y}_n)$\;
        EOS terminal value $V_{n,|\boldsymbol{y}_n|+1}\gets 0$\;
        }{reference path}{
        \textcolor{blue}{\textbf{[TP, PP, \textmd{SP,} CP, EP]}} \\ Compute reference log-probs $\boldsymbol{\ell}_{\mathrm{ref}}^{(n)} \gets \log\pi_{\mathrm{ref}}(\boldsymbol{y}_n \mid \boldsymbol{x}_n)$\;
        }

        Per-token KL-shaped rewards $\boldsymbol{r}_n \gets -\beta\,\left(\boldsymbol{\ell}_{\mathrm{old}}^{(n)} - \boldsymbol{\ell}_{\mathrm{ref}}^{(n)}\right) + s_n\,\boldsymbol{e}_{|\boldsymbol{y}_n|}$ \textit{(KL penalty enters via reward and propagates through GAE; no KL term in PPO loss)}\\
      
        Initialize $\boldsymbol{A}_n \gets []$, $\hat{\boldsymbol{G}}_n \gets []$, and EOS terminal advantage $A_{n,|\boldsymbol{y}_n|+1} \gets 0$\;
        \For{$t = |\boldsymbol{y}_n|$ \KwTo $1$}{
            Temporal difference residual $\delta_{n,t} \gets r_{n,t} + \gamma V_{n,t+1} - V_{n,t}$\;

            Compute advantage (GAE) $A_{n,t} \gets \delta_{n,t} + \gamma\lambda A_{n,t+1}$\;

            Compute discounted return $\hat G_{n,t} \gets V_{n,t} + A_{n,t}$\;

            Prepend $A_{n,t}$ to $\boldsymbol{A}_n$ and $\hat G_{n,t}$ to $\hat{\boldsymbol{G}}_n$\;
        }

        Augment rollout $n$ in $\mathcal{B}_{\mathrm{roll}}$ with $\left(\boldsymbol{A}_n,\hat{\boldsymbol{G}}_n\right)$\;
    }

    \BlankLine
    \textbf{——————————— Training via PPO —————————---}

    \For{$e=1$ \KwTo $E_{\mathrm{PPO}}$}{
        Shuffle $\mathcal{B}_{\mathrm{roll}}$ and partition into minibatches $\{\mathcal{B}\}$\;

        \textcolor{blue}{\textbf{[DP]}} \ForEach{minibatch $\mathcal{B}$}{
            \textcolor{blue}{\textbf{[DP]}} \ForEach{rollout $n \in \mathcal{B}$}{
                \parpair{actor path}{
                \textcolor{blue}{\textbf{[TP, PP, SP, CP, EP]}} \\ Current log-probs $\boldsymbol{\ell}_\theta^{(n)} \gets \log\pi_\theta(\boldsymbol{y}_n\mid \boldsymbol{x}_n)$\;

                Importance ratios $\boldsymbol{\rho}_n \gets \exp\!\left(\boldsymbol{\ell}_\theta^{(n)} - \boldsymbol{\ell}_{\mathrm{old}}^{(n)}\right)$\;
                }{critic path}{
                \textcolor{blue}{\textbf{[TP, PP, SP, CP, EP]}} \\ Current values $\boldsymbol{V}'_n \gets V_\psi(\boldsymbol{x}_n, \boldsymbol{y}_n)$\;
                }
            }

            Number of tokens in batch $Z \gets \sum_{n \in \mathcal B} |\boldsymbol{y}_n|$\;
            \parpair{actor path}{
            Compute PPO loss:\\
            $\mathcal{L}_{\mathrm{PPO}} =
            -\frac{1}{Z}\sum_{n\in\mathcal{B}}\sum_{t=1}^{|\boldsymbol{y}_n|}
            \min(\rho_{n,t}A_{n,t},$ \\
            $ \text{ \quad \quad \quad \quad \quad \ \ \ \ \ \ \ \ \ \quad \quad \quad \quad \quad clip}(\rho_{n,t},1\pm\epsilon)\,A_{n,t})$\;

            (Optional) Entropy bonus; $H(\cdot)$ is the actor's per-token Shannon entropy ($c_H = 0$ disables) \\
            $\mathcal{L}_{H} =
            -\frac{1}{Z}\sum_{n\in\mathcal{B}}\sum_{t=1}^{|\boldsymbol{y}_n|}
            H\left(\pi_\theta\left(\cdot\mid \boldsymbol{x}_n, \boldsymbol{y}_{n,<t}\right)\right)$\;

            Define actor objective: $\mathcal{L}_{\mathrm{PPO}} \gets \mathcal{L}_{\mathrm{PPO}} + c_H\,\mathcal{L}_{H}$\;

            \textcolor{blue}{\textbf{[TP, PP, SP, CP, EP]}} \\ Backpropagate \& update on $\theta$ to minimize $\mathcal L_\mathrm{PPO}$\;
            }{critic path}{
            Compute critic loss:\\
            $\mathcal{L}_{\mathrm{critic}} =
            \frac{1}{Z}\sum_{n\in\mathcal{B}}\sum_{t=1}^{|\boldsymbol{y}_n|}
            (V'_{n,t}-\hat G_{n,t})^2$\;

            \textcolor{blue}{\textbf{[TP, PP, SP, CP, EP]}} \\ Backpropagate \& update on $\psi$ to minimize $\mathcal L_\mathrm{critic}$\;
            }
        }
    }
}
\end{algorithm}

%% file: code-grpo.tex
\begin{algorithm}[t]
\caption{Algorithmic and Mathematical Specification of GRPO}
\label{alg:grpo}
\KwIn{
KL coefficient $\beta$, GRPO clip range $\epsilon$; 
set of prompts $X$, prompt batch size $B$, candidates/prompt $K$, training epochs $E_{\mathrm{GRPO}}$.
}
\KwOut{Trained parameters $\theta$.}
\BlankLine

\ForEach{prompt batch $\{\boldsymbol{x}_b\}_{b=1}^B \subset X$}{
    \BlankLine
    \textbf{——————————— Generation (Online) ————————}

    Generate completions $\mathcal{B}_{\mathrm{roll}} \gets \left\{\left(\boldsymbol{x}_b, \boldsymbol{y}_b^{(i)}, \boldsymbol{\ell}_{\mathrm{old},b}^{(i)}\right)\right\}_{b=1,i=1}^{B,K}$ via \textbf{Algorithm~\ref{alg:generation-common}}\;

    \BlankLine
    \textbf{——————————— Assessment ————————————}
    
    \textcolor{blue}{\textbf{[DP]}}  \For{prompt $b=1$ \KwTo $B$}{
        \textcolor{blue}{\textbf{[DP]}}  \For{candidate $i=1$ \KwTo $K$}{
            \parpairAC{reward path}{
            \textcolor{blue}{\textbf{[TP, PP, \textmd{SP,} CP, EP]}} \\ Compute scalar reward $r_{b}^{(i)} \gets R_\varphi\left(\boldsymbol{x}_b, \boldsymbol{y}_b^{(i)}\right)$\;
            }{reference path}{
            \textcolor{blue}{\textbf{[TP, PP, \textmd{SP,} CP, EP]}} \\ Compute reference log-probs $\boldsymbol{\ell}_{\mathrm{ref},b}^{(i)} \gets \log\pi_{\mathrm{ref}}\left(\boldsymbol{y}_b^{(i)} \mid \boldsymbol{x}_b\right)$\;
            }
        }
        Group mean $\mu_b \gets \frac{1}{K}\sum_{i=1}^{K} r_b^{(i)}$\;

        Group std $\sigma_b \gets \left(\tfrac{1}{K}\sum_{i=1}^{K} \left(r_b^{(i)} - \mu_b\right)^2\right)^{1/2}$\;
        \textcolor{blue}{\textbf{[DP]}} \For{candidate $i=1$ \KwTo $K$}{
            Per-token advantage (broadcast across tokens) $\boldsymbol{A}_b^{(i)} \gets \frac{r_{b}^{(i)}-\mu_b}{\sigma_b+10^{-8}}\,\boldsymbol{1}$\;

            Augment rollout $(b,i)$ in $\mathcal{B}_{\mathrm{roll}}$ with $\left(\boldsymbol{A}_b^{(i)},\, \boldsymbol{\ell}_{\mathrm{ref},b}^{(i)}\right)$\;
        }
    }
    
    Flatten $\mathcal{B}_{\mathrm{roll}} \gets \left\{\left(\boldsymbol{x}_n, \boldsymbol{y}_n, \boldsymbol{\ell}_{\mathrm{old}}^{(n)}, \boldsymbol{A}_n, \boldsymbol{\ell}_{\mathrm{ref}}^{(n)}\right)\right\}_{n=1}^{BK}$\;

    \BlankLine
    \textbf{——————————— Training via GRPO —————————}

    \For{$e=1$ \KwTo $E_{\mathrm{GRPO}}$}{
        Shuffle $\mathcal{B}_{\mathrm{roll}}$ and partition into minibatches $\{\mathcal{B}\}$\;

        \textcolor{blue}{\textbf{[DP]}} \ForEach{minibatch $\mathcal{B}$}{
            \textcolor{blue}{\textbf{[DP]}}  \ForEach{rollout $n \in \mathcal{B}$}{
                \lanebox{laneActor}{actor path}{%
                \textcolor{blue}{\textbf{[TP, PP, SP, CP, EP]}} \\ Current log-probs $\boldsymbol{\ell}_\theta^{(n)} \gets \log\pi_\theta(\boldsymbol{y}_n \mid \boldsymbol{x}_n)$\;

                Importance ratios $\boldsymbol{\rho}_n \gets \exp\!\left(\boldsymbol{\ell}_\theta^{(n)} - \boldsymbol{\ell}_{\mathrm{old}}^{(n)}\right)$\;
                }

                Per-token KL approximation $\boldsymbol{d}_n \gets \exp\!\left(\boldsymbol{\ell}_{\mathrm{ref}}^{(n)} - \boldsymbol{\ell}_\theta^{(n)}\right) - \left(\boldsymbol{\ell}_{\mathrm{ref}}^{(n)} - \boldsymbol{\ell}_\theta^{(n)}\right) - 1$\;
            }

            Number of tokens in batch $Z \gets \sum_{n \in \mathcal B} |\boldsymbol{y}_n|$\;

            \lanebox{laneActor}{actor path}{%
            Compute GRPO loss:\\
            $\mathcal{L}_{\mathrm{GRPO}} =
            -\frac{1}{Z}\sum_{n\in\mathcal{B}}\sum_{t=1}^{|\boldsymbol{y}_n|}
            \Big[\min(\rho_{n,t}A_{n,t},$ \\
            $\text{ \quad \quad \quad \quad \quad \quad \quad \ \ \ \ \ \ \ \ \ clip}(\rho_{n,t},1\pm\epsilon)\,A_{n,t}) - \beta\,d_{n,t}\Big]$\;

            \textcolor{blue}{\textbf{[TP, PP, SP, CP, EP]}} \\ Backpropagate \& update on $\theta$ to minimize $\mathcal{L}_{\mathrm{GRPO}}$\;
            }
        }
    }
}
\end{algorithm}

%% file: code-dpo.tex
\begin{algorithm}[t]
\caption{Algorithmic and Mathematical Specification of DPO}
\label{alg:dpo}
\KwIn{
KL coefficient $\beta$, training epochs $E_{\mathrm{DPO}}$;
preference dataset $\mathcal D = \!\left\{\left(\boldsymbol{x}_n, \boldsymbol{y}_w^{(n)}, \boldsymbol{y}_l^{(n)}\right)\right\}_{n=1}^{N}$.
}
\KwOut{Trained parameters $\theta$.}
\BlankLine
\textbf{——————————— Training via DPO ————————————}

\For{$e = 1$ \KwTo $E_{\mathrm{DPO}}$}{
    Shuffle $\mathcal D$ and partition into minibatches $\{\mathcal B\}$\;
    \textcolor{blue}{\textbf{[DP]}} \ForEach{minibatch $\mathcal B$}{ 
        \textcolor{blue}{\textbf{[DP]}}  \ForEach{$\left(\boldsymbol{x}_n,\boldsymbol{y}_w^{(n)},\boldsymbol{y}_l^{(n)}\right)\in\mathcal B$}{
            \parpairARef{actor path}{
            \textcolor{blue}{\textbf{[TP, PP, SP, CP, EP]}} Actor log-probs:\\
            $\boldsymbol{\ell}_{\theta,w}^{(n)} \gets \log\pi_\theta\left(\boldsymbol{y}_w^{(n)} \mid \boldsymbol{x}_n\right)$ \textit{(preferred completion)}\;

            $\boldsymbol{\ell}_{\theta,l}^{(n)} \gets \log\pi_\theta\left(\boldsymbol{y}_l^{(n)} \mid \boldsymbol{x}_n\right)$ \textit{(dispreferred completion)}\;}
            {reference path}{
            \textcolor{blue}{\textbf{[TP, PP, \textmd{SP,} CP, EP]}} Reference log-probs:\\
            $\boldsymbol{\ell}_{\mathrm{ref},w}^{(n)} \gets \log\pi_{\mathrm{ref}}\left(\boldsymbol{y}_w^{(n)} \mid \boldsymbol{x}_n\right)$ \textit{(preferred completion)}\;

            $\boldsymbol{\ell}_{\mathrm{ref},l}^{(n)} \gets \log\pi_{\mathrm{ref}}\left(\boldsymbol{y}_l^{(n)} \mid \boldsymbol{x}_n\right)$ \textit{(dispreferred completion)}\;}
            Trajectory-level log-ratio difference: $h_n \gets \boldsymbol{1}^\top\!\left(\boldsymbol{\ell}_{\theta,w}^{(n)} - \boldsymbol{\ell}_{\mathrm{ref},w}^{(n)}\right) - \boldsymbol{1}^\top\!\left(\boldsymbol{\ell}_{\theta,l}^{(n)} - \boldsymbol{\ell}_{\mathrm{ref},l}^{(n)}\right)$\;
        }
        \lanebox{laneActor}{actor path}{%
        Compute DPO loss, where $\sigma(\cdot)$ is the sigmoid:\\
        $\mathcal{L}_{\mathrm{DPO}} = -\frac{1}{|\mathcal B|}
        \sum_{n \in \mathcal B}\log \sigma \big(\beta\,h_n\big)$\;

        \textcolor{blue}{\textbf{[TP, PP, SP, CP, EP]}} \\ Backpropagate \& update on $\theta$ to minimize $\mathcal{L}_{\mathrm{DPO}}$\;
        }
    }
}
\end{algorithm}

%% file: designs.tex
\section{Analysis of Existing Models \& Designs}

We also analyze existing models and frameworks.

\subsection{Reasoning Language Models}

\input{table-models-specific}

We compare representative post-trained LLMs and RLMs in Table~\ref{tab:general-llm-rlm-comparison}. We classify a model as a \emph{general aligned LLM} if post-training primarily improves instruction-following, helpfulness, safety, or general assistant behavior, even when the model can solve reasoning tasks. We classify a model as an \textit{RLM} if its training or inference pipeline explicitly targets reasoning trajectories, verifiable rewards, long CoT, search, tool use, or controllable test-time computation. This distinction is often blurred in practice: recent systems increasingly unify fast response modes and slow reasoning modes inside a single routed or hybrid model family.

Overall, the comparison shows a shift from reference-driven alignment as \emph{preference optimization} to alignment as \emph{reasoning-time computation}. Early RL-enhanced LLMs primarily used PPO-like RLHF or more lightweight DPO-style preference optimization to improve instruction following, safety, and general assistant quality, whereas RLMs increasingly optimize or allocate compute to long-CoT trajectories, RLVR, search, tools, agentic environments, or hybrid fast/thinking modes. The training signal is often more predictive of reasoning gains than parameter count alone: exact-answer rewards, unit tests, process/preference models, MCTS, or environment feedback can make smaller specialized models competitive on narrow math/code tasks, while frontier systems retain broader coverage through scale, data diversity, and tool integration. Architecturally, sparse MoE has become central for open frontier models because it increases total capacity at moderate active-parameter cost, but it shifts systems pressure toward expert parallelism, routing balance, all-to-all communication, and memory placement; dense models remain simpler and more predictable for deployment. 


\subsection{Training \& Inference Frameworks for RLMs}

\input{table-frameworks}

Table~\ref{tab:frameworks} shows that modern RL-LLM frameworks differ primarily by how aggressively they separate,
specialize, and overlap the rollout and training workloads. In practice, wall-clock time is often dominated by rollout/generation under long-output reasoning settings, while gradient updates are comparatively more parallelizable; consequently, the highest-throughput frameworks either (a) maximize rollout throughput via inference engines and stage-specific parallelism, or (b) overlap generation with evaluation and training via stage fusion or asynchronous execution. 

{Library-centric stacks} such as TRL~\cite{vonwerra2022trl}, DeepSpeed-Chat~\cite{yao2023deepspeed},
and ColossalChat~\cite{colossalchat-medium} are easiest to use and rely mainly
on DP/ZeRO/FSDP-style sharding~\cite{rajbhandari2020zero,zhao2023pytorchfsdp}
plus optional fast rollout engines such as vLLM~\cite{kwon2023vllm}, but they
expose limited inter-model scheduling. 
{Orchestrator-centric stacks} such as
OpenRLHF~\cite{hu2025openrlhf}, HybridFlow/verl~\cite{sheng2024hybridflow},
SkyRL~\cite{skyrl2026}, NeMo RL~\cite{shen2024nemoaligner, nvidia2026nemorl}, slime~\cite{slime_github}, ROLL~\cite{wang2025reinforcement} and
ReaL~\cite{mei2025real} treat RLHF as a multi-role dataflow. They provide explicit placement and resource management for actor, rollout, reward, reference, critic, and data-buffer components, and commonly combine training backends such as DeepSpeed ZeRO, PyTorch FSDP/FSDP2, or Megatron-Core that support TP/PP/CP/DP/EP~\cite{shoeybi2019megatron, narayanan2021efficient,liu2023ring,jacobs2023deepspeedulysses}, with 
rollout engines such as vLLM, SGLang, or TensorRT-LLM~\cite{kwon2023vllm,
zheng2023sglang,nvidia2024tensorrtllm}. These systems often support heterogeneous parallelism across roles, weight synchronization, and resharding between training and rollout with explicit weight synchronization or resharding~\cite{mei2025real, sheng2024hybridflow}. 
{Scheduling-centric systems} such as RLHFuse~\cite{
zhong2024optimizing}, Pipe-RLHF~\cite{xu2025piperlhf}, StreamRL~\cite{
zhong2025streamrl}, AReaL~\cite{fu2025areal}, AsyncFlow~\cite{han2025asyncflow},
Laminar~\cite{sheng2025laminar}, StaleFlow~\cite{li2026staleflow}, and
MindSpeed RL~\cite{feng2025mindspeedrl} are best viewed as emphasizing
scheduling policies rather than defining a disjoint framework class. In practice,
many orchestrator-centric frameworks can also manually schedule workloads or
incorporate similar optimizations. These systems target the dominant bottlenecks
of long-tailed autoregressive rollout by improving utilization by
stage fusion~\cite{zhong2024optimizing}, streaming and disaggregation~\cite{
zhong2025streamrl}, dynamic load balancing~\cite{han2025asyncflow}, and
bounded-stale asynchronous execution~\cite{noukhovitch2025asynchronous,
fu2025areal,li2026staleflow}, at the cost of more complex queues, weight-version
management, and convergence validation. 

Practically, small experiments should favor library-centric stacks; memory-bounded large-model runs should favor mature state-sharding backends such as FSDP/ZeRO or Megatron~\cite{rajbhandari2020zero,zhao2023pytorchfsdp,shoeybi2019megatron};
RL post-training workloads whose wall-clock time is dominated by rollout/generation should favor disaggregated vLLM/SGLang-style rollout~\cite{ kwon2023vllm,zheng2023sglang,zhong2025streamrl}; and frontier-scale reasoning RL increasingly requires async or streaming systems that explicitly control staleness and generation-length skew~\cite{fu2025areal,sheng2025laminar,li2026staleflow}.

Across frameworks, data parallelism remains a common outer abstraction, but
modern RL post-training stacks increasingly compose multiple forms of
parallelism across roles and stages. FSDP/ZeRO-style state sharding and
Megatron-FSDP occupy the same broad design space for sharding optimizer states,
gradients, and parameters, while Megatron-style backends can further combine DP
with TP, PP, CP, and EP when model size, sequence length, or MoE require
it. Thus, the main systems issue is not a fixed choice between FSDP/ZeRO and
Megatron, but how the framework places heterogeneous roles, synchronizes or
reshards weights between the trainer and rollout workers, and balances optimizer-state memory, communication, KV-cache pressure, and variable rollout lengths. In this sense, orchestrator-centric and scheduling-centric systems are better viewed as different emphases within the same design space: the former provides the multi-role execution substrate, while the latter emphasizes utilization policies for long-tailed generation workloads.

%% file: table-models-specific.tex
\begin{table*}[hbtp]
\centering
\scriptsize
\setlength{\tabcolsep}{1.5pt}
\renewcommand{\arraystretch}{1.13}
\begin{tabular}{
P{0.14\textwidth}
P{0.04\textwidth}
P{0.17\textwidth}
P{0.24\textwidth}
P{0.38\textwidth}}
\toprule
\textbf{Model / family} &
\textbf{Access} &
\textbf{Scale / arch.} &
\textbf{Post-training signal} &
\textbf{Selected system details}
\\
\midrule
\multicolumn{5}{l}{\textbf{General aligned large language models (LLMs)}} \\
\midrule

InstructGPT~\cite{ouyang2022training} &
Closed &
1.3B, 6B, 175B dense &
PPO-like (SFT + RM + PPO) &
Canonical actor--reward--reference PPO loop.
\\

GPT-4~\cite{openai2023gpt4} &
Closed &
Undisclosed &
PPO-like (rule-based rewards) &
Frontier-scale alignment; limited public systems detail.
\\

Gemini 1.x~\cite{team2023gemini} &
Closed &
Undisclosed &
PPO-like (iterative RM refinement) &
Repeated RM/RL cycles increase assessment cost.
\\

Claude 3~\cite{anthropic2024claude,bai2022constitutional} &
Closed &
Undisclosed &
PPO-like (RLAIF / Constitutional AI) &
AI-generated preference labels from written principles.
\\

Reka~\cite{reka2024} &
Mixed &
7B, 21B dense &
PPO-like (multi-round RLHF) &
Standard PPO-style alignment at moderate scale.
\\

InternLM2~\cite{cai2024internlm2} &
Open &
1.8B, 7B, 20B dense &
PPO-like (conditional RM) &
Multiple domain-specific reward heads.
\\

Zephyr~\cite{tunstall2023zephyr,hong2024orpo} &
Open &
MoE variant reported &
DPO-like (ORPO) &
No explicit reward model or PPO loop.
\\

Phi-3~\cite{abdin2024phi3} &
Open &
3.8B, 7B, 14B dense &
DPO-like &
SFT-like alignment cost.
\\

Phi-4~\cite{abdin2024phi4} &
Open &
7B, 14B dense &
DPO-like (RLAIF data) &
Preference tuning over AI/human feedback data.
\\

ChatGLM~\cite{glm2024chatglm} &
Open &
6B, 9B dense &
Mixed (PPO-like vs. DPO-like) &
Controlled comparison of online vs.~DPO training.
\\

Gemma 2~\cite{team2024gemma} &
Open &
2B, 9B, 27B dense &
PPO-like (Bradley--Terry RM) &
Reward-model alignment.
\\

Starling-7B~\cite{zhu2024starling} &
Open &
7B dense &
PPO-like (RLAIF; Plackett--Luce RM) &
Ranking-based RM; partial model updates reduce memory.
\\

Hermes 3~\cite{teknium2024hermes} &
Open &
8B, 70B, 405B dense &
DPO-like (LoRA-DPO) &
Only LoRA/adapters are trained; optimizer memory is small.
\\

Athene-70B~\cite{Athene2024} &
Open &
70B dense &
PPO-like (details limited) &
A model with preference/safety tuning.
\\

Llama 3 \& 3.1~\cite{dubey2024llama} &
Open &
8B, 70B, 405B dense &
DPO-like (RM + rejection sampling) &
--- 
\\

Qwen2 \& Qwen2.5~\cite{yang2024qwen2} &
Open &
\makecell[l]{0.5B--72B dense; \\ 57B MoE (14B active)} &
DPO-like (offline + online refresh) &
New preference pairs are created from online model samples.
\\

Nemotron-4 340B~\cite{adler2024nemotron} &
Open &
340B-class model &
DPO-like (RPO) &
Quality-aware preference optimization without PPO.
\\

DeepSeek-V3~\cite{liu2025deepseekv3} &
Open &
671B MoE ($\sim$37B active) &
Hybrid alignment &
MoE, MLA, FP8, MTP, and DualPipe.
\\

Llama 4 Scout~\cite{meta2025llama4} &
Open &
$\sim$109B MoE (17B active) &
Hybrid alignment &
Long context.
\\

Llama 4 Maverick~\cite{meta2025llama4} &
Open &
$\sim$400B MoE (17B active) &
Hybrid alignment &
MoE and multimodality.
\\

\midrule
\multicolumn{5}{l}{\textbf{Reasoning language models (RLMs)}} \\
\midrule

OpenAI o1~\cite{openai2024o1} &
Closed &
Undisclosed &
Hybrid RL (CoT) &
``Reasoning tokens'' become inference-time cost.
\\

OpenAI o3~\cite{openai2025o3o4} &
Closed &
Undisclosed &
Hybrid RL (CoT + tools) &
Training-time RL and tool-using inference both dominate.
\\

OpenAI o4-mini~\cite{openai2025o3o4} &
Closed &
Undisclosed &
Hybrid RL (CoT + tools) &
Lower-cost reasoning model; latency/quality trade-off.
\\

GPT-5~\cite{openai2025gpt5system} &
Closed &
Undisclosed &
Hybrid (fast/thinking router) &
Runtime routes between fast and reasoning modes.
\\

GPT-5.4 Thinking~\cite{openai2026gpt54} &
Closed &
\makecell[l]{Undisclosed} &
Hybrid (coding + agents) &
Agentic workflow.
\\

GPT-5.5 Thinking~\cite{openai2026gpt55} &
Closed &
Undisclosed &
Hybrid (agentic reasoning) &
Higher autonomy increases orchestration cost.
\\

Claude 3.7 Sonnet~\cite{anthropic2025claude37} &
Closed &
Undisclosed &
PPO-like/RLAIF + thinking mode &
Same model supports quick and extended reasoning.
\\

Claude 4 family~\cite{anthropic2025extendedthinking} &
Closed &
Undisclosed &
PPO-like/RLAIF + thinking mode &
Reasoning budget is a user/system cost knob.
\\

Gemini 2.5 Pro~\cite{google2025gemini25} &
Closed &
Sparse MoE; multimodal &
Hybrid RL (thinking mode) &
Long context.
\\

Gemini 2.5 Flash~\cite{google2025gemini25} &
Closed &
Sparse MoE with a lower-latency variant &
Hybrid RL (low-cost thinking) &
Cheaper reasoning.
\\

Gemini 2.5 Deep Think~\cite{google2025deepthink} &
Closed &
Gemini 2.5 family &
Search-like, parallel reasoning &
Multiple reasoning paths increase per-query compute.
\\

DeepSeek-R1~\cite{guo2025deepseekr1} &
Open &
671B MoE / $\sim$37B active &
GRPO-like (RLVR; no critic) &
Removes \(V_\psi\); generation remains bottleneck.
\\

DAPO~\cite{yu2025dapo} &
Open &
32B dense &
GRPO-like (DAPO; no critic) &
Critic-free long-CoT RL.
\\

Qwen3~\cite{yang2025qwen3} &
Open &
Dense/MoE, 0.6B--235B &
Hybrid (thinking/non-thinking) &
Variable reasoning length.
\\

Kimi k1.5~\cite{du2025kimi} &
Closed &
Undisclosed; long-context multimodal &
DPO-like/custom RL (KL-DPO) &
Long trajectories dominate training and serving cost.
\\

Kimi K2~\cite{bai2025kimik2} &
Open &
1T MoE (32B active) &
Agentic RL-like (tool environments) &
---
\\

Kimi K2.5~\cite{kimi2026k25} &
Open &
MoE; multimodal agentic extension &
Agentic RL-like (visual tools) &
Adds multimodal environment overheads.
\\

Phi-4-reasoning~\cite{microsoft2025phi4reasoning} &
Open &
14B dense &
DPO-like (RLVR) &
Small strong STEM model at low serving cost.
\\

rStar-Math~\cite{guan2025rstar} &
Open &
7B-scale policy/reward components &
Search/RLVR-like (MCTS + PRM) &
Search improves math but increases inference compute.
\\

Grok 4~\cite{xai2025grok4} &
Closed &
Undisclosed &
Hybrid RL &
---
\\

Grok 4 Heavy~\cite{xai2025grok4} &
Closed &
Undisclosed; multi-agent variant &
Agentic / parallel reasoning &
Parallel agents increase inference compute.
\\

\bottomrule
\end{tabular}
\vspace{-1em}
\caption{\textbf{A comparison of representative general aligned LLMs and RLMs.}
The upper block lists general aligned LLMs, which primarily target instruction
following, safety, preference alignment, and broad assistant quality. The lower
block lists RLMs, which explicitly target reasoning trajectories, long-CoT,
RLVR/verifiable rewards, search, tools, multi-agent inference, or controllable
test-time computation. The ``Post-training signal'' column follows the taxonomy
used in this paper: PPO-like methods use reward-model- or RLAIF-based policy
optimization; GRPO-like methods use critic-free grouped RL/RLVR; DPO-like methods
use direct preference optimization; Hybrid denotes mixed or undisclosed pipelines;
Search/Agentic denotes explicit search, tools, environments, or multi-agent
reasoning. Proprietary details are based on public reports and should be treated
as approximate.}
\label{tab:general-llm-rlm-comparison}
\end{table*}

%% file: table-frameworks.tex
\begin{table*}[t]
\centering
\setlength{\tabcolsep}{1.0pt}
\ifsq\renewcommand{\arraystretch}{0.9}\fi
\scriptsize
\begin{tabular}{lcccccccccccccl}
\toprule
& \multicolumn{8}{c}{\textbf{Intra-Model Parallelism}} & \multicolumn{5}{c}{\textbf{Inter-Model Parallelism}} & \\
\multirow{2}{*}{\textbf{Framework}} & & & & & & & & & & & & & & \multicolumn{1}{c}{\multirow{2}{*}{\textbf{Remarks}}} \\
\cmidrule(lr){2-9} \cmidrule(lr){10-14}
& \textbf{DP} & \textbf{Z1} & \textbf{Z2} & \textbf{Z3} & \textbf{TP} & \textbf{CP} & \textbf{EP} & \textbf{PP} & \makecell[c]{\textbf{AC} \\ \textbf{Share}} & \makecell[c]{\textbf{Dis-} \\ \textbf{agg.}} & \makecell[c]{\textbf{Stage}\\ \textbf{Fusion}} & \textbf{Hybrid} & \textbf{Async} & \\
\midrule

TRL~\cite{vonwerra2022trl} 
& \faY & \faH & \faY & \faY & \faY & \faY & \faN & \faN
& \faY & \faY & \faH & \faY & \faY
& Accelerate; TP/vLLM mostly rollout; CP via FSDP2/Ulysses \\

ColossalChat~\cite{colossalchat-medium}
& \faY & \faY & \faY & \faY & \faY & \faY & \faH & \faY
& \faN & \faN & \faN & \faN & \faN
& Colossal-AI HybridParallel + Gemini; TP/PP/SP, Z1/2 and Z3 \\

DeepSpeed-Chat~\cite{yao2023deepspeed}
& \faY & \faY & \faY & \faY & \faY & \faN & \faN & \faN
& \faN & \faN & \faN & \faY & \faN
& Train ZeRO-2/3; Hybrid Engine uses TP mainly for rollout \\

OpenRLHF~\cite{hu2025openrlhf}
& \faY & \faH & \faY & \faY & \faY & \faY & \faN & \faY
& \faN & \faY & \faH & \faY & \faY
& Train: ZeRO-3/AutoTP; rollout: vLLM TP/PP; RingAttention \\

HybridFlow / verl~\cite{sheng2024hybridflow}
& \faY & \faH & \faH & \faY & \faY & \faY & \faY & \faY
& \faN & \faY & \faH & \faY & \faY
& Train: FSDP or Megatron; rollout: vLLM/SGLang/TensorRT-LLM \\

NeMo RL~\cite{shen2024nemoaligner, nvidia2026nemorl}
& \faY & \faH & \faH & \faY & \faY & \faY & \faY & \faY
& \faN & \faY & \faN & \faY & \faY
& Train: DTensor or Megatron; rollout: vLLM/Megatron/SGLang \\

ReaL~\cite{mei2025real}
& \faY & \faH & \faN & \faN & \faY & \faN & \faN & \faY
& \faN & \faY & \faH & \faY & \faN
& Per-stage DP/TP/PP/SP plans; ZeRO backend not stage-enumerated \\

RLHFuse~\cite{zhong2024optimizing}
& \faH & \faH & \faN & \faN & \faH & \faN & \faN & \faY
& \faN & \faH & \faY & \faH & \faH
& Scheduling layer; inter-stage sample fusion + intra-stage fused PP \\

FlexRLHF~\cite{xiao2025flexrlhf}
& \faY & \faH & \faH & \faH & \faH & \faN & \faN & \faH
& \faY & \faY & \faH & \faY & \faN
& DeepSpeed-based; AC-share/nonshare; interleaving/disaggregated layouts \\

AReaL~\cite{fu2025areal}
& \faY & \faH & \faH & \faY & \faY & \faY & \faY & \faY
& \faN & \faY & \faH & \faY & \faY
& Train: Megatron/FSDP/Archon; infer: SGLang default, vLLM optional \\

MindSpeed RL~\cite{feng2025mindspeedrl}
& \faY & \faH & \faH & \faH & \faY & \faY & \faY & \faY
& \faN & \faY & \faH & \faY & \faY
& Ascend stack; co-card/decoupled deploy, repartition, partial rollout \\

slime~\cite{slime_github}
& \faY & \faH & \faH & \faH & \faY & \faY & \faY & \faY
& \faN & \faY & \faH & \faY & \faY
& Megatron train + SGLang rollout \\

ROLL~\cite{wang2025reinforcement}
& \faY & \faH & \faY & \faY & \faY & \faY & \faY & \faY
& \faN & \faY & \faH & \faY & \faY
& train: DeepSpeed/FSDP2/Megatron; infer: vLLM/SGLang/Megatron \\

\midrule

StreamRL~\cite{zhong2025streamrl}
& \multicolumn{8}{c}{outside system focus}
& \faN & \faY & \faY & \faY & \faY
& Disaggregation-first; stream generation; full overlap in async mode \\

AsyncFlow~\cite{han2025asyncflow}
& \multicolumn{8}{c}{outside system focus}
& \faN & \faY & \faH & \faY & \faY
& Distributed data/param streaming; producer--consumer async \\

Laminar~\cite{sheng2025laminar}
& \multicolumn{8}{c}{outside system focus}
& \faN & \faY & \faH & \faY & \faY
& Relay-worker parameter service; trajectory-level async repack \\

StaleFlow~\cite{li2026staleflow}
& \multicolumn{8}{c}{outside system focus}
& \faN & \faY & \faH & \faY & \faY
& Data+parameter servers; explicit global staleness control \\

\bottomrule
\end{tabular}
\vspace{-1em}
\caption{\textbf{Comparison of public training / inference frameworks for LLM and RLM post-training under the systems taxonomy.}
\textbf{Intra-Model Parallelism}: DP=data parallel; Z1/Z2/Z3=ZeRO/FSDP-like sharding levels (optimizer / optimizer+gradients / parameters+gradients+optimizer; FSDP2 is counted under Z3); TP=tensor parallel; CP=context or long-sequence parallelism; EP=expert parallelism; PP=pipeline parallelism.
\textbf{Inter-Model Parallelism}: AC Share=shared actor/critic parameters or value-head mode; Disagg.=separate services or device groups for rollout, reward, or training; Stage Fusion=explicit overlap or fusion across normally sequential RL stages or subtasks; Hybrid=stage-specific backends/layouts or runtime resharding/refit; Async=asynchronous or bounded-stale execution.
\textbf{Symbols}: \faY~documented support; \faH~indirect, partial, backend-specific, or train-vs-rollout-specific support; \faN~no documented public support; \noAnswer~undocumented or unclear in public materials.}
\label{tab:frameworks}
\end{table*}

%% file: opportunities.tex
\section{Research Opportunities}
\label{sec:research-opportunities}

We briefly outline potential opportunities for future research in the performance aspects of RLMs.

\textbf{Resource-aware test-time compute.}
RLMs increasingly trade inference-time compute for accuracy through long-CoT
generation, self-consistency, search, tool use, and other strategies, exploring different scaling regimes and tradeoffs~\cite{wu2025inference, snell2025scaling, hou2024does, zeng2024skywork, tan2025scaling, kim2025not, hou2024does, openai2025o3o4, google2025gemini25}. Future systems could treat reasoning as a constrained optimization problem: maximize expected utility subject to token,
latency, memory, tool-call, and energy budgets. This opens the door to interesting research into making
reasoning budget a first-class scheduling variable, and into online predictors
of task difficulty, verifier value, branch utility, and stopping time.

\textbf{Efficient RLM execution structures beyond chains.}
RLM reasoning is usually represented as a linear token sequence. Prior works such as
Tree of Thoughts, Graph of Thoughts, and Hypergraph-of-Thoughts work shows that
nonlinear reasoning structures can improve search and aggregation in-context~\cite{yao2023tree,  ning2023skeleton, besta2023graph, yao2023thinking, besta2025reasoning, besta2025demystifyingchains}. Some efforts to distill this behavior into weights have been made into this direction by harnessing aggregation during fine-tuning~\cite{ai2025beyond, li2025drafts, zhao2025majority, li2025llms}.
Efficient and effective integration of such structures and ideas in the RL execution pipeline is an interesting novel direction in the science of RLM performance and parallel design.

\textbf{Automatic selection of parallelization schemes.}
One concrete idea of how to enhance the parallel design of RLM pipelines, is to provide an effective way of \textit{automating} the intra-parallelization of each model invocation based on the available hardware and cluster conditions. There have been works into this direction, for example AutoDDL~\cite{chen2023autoddl} and PyTorch/XLA SPMD~\cite{google2024introducing}, but they focus on an individual Transformer invocations and not whole pipelines~\cite{chen2023autoddl, google2024introducing}.

\textbf{Automatic derivation of reasoning topologies and schedules.}
In the RLM designs that harness explicit structures such as MCTS, most methods still use hand-designed chains, trees, beams,
and others~\cite{besta2025reasoning}. A potential direction is to automatically derive both the reasoning
topology and the execution schedule from aspects such as task difficulty, model uncertainty,
verifier availability, hardware availability and performance properties, and others. Dynamic graph representations and
algorithms could offer useful
templates for such planners in terms of performance-focused aspects such as algorithm design~\cite{besta2021practice}, effective task graph decompositions and partitioning~\cite{gianinazzi2018communication, bulucc2016recent}, scheduling~\cite{dandashi2010graph, besta2020highc}, and others. Example questions to pursuit would be how many
branches to explore, when to prune a branch, when to call a verifier, when to merge
thoughts, how to map the resulting DAG to device groups, and others.

\textbf{RLVR beyond final-answer rewards.}
RLVR scales reasoning post-training through exact-answer checks, unit tests, and
task-specific verifiers, but its mechanisms remain incompletely understood; for example,
weak or spurious rewards can sometimes still improve reasoning, and outcomes are
model-family dependent~\cite{shao2025spurious}. Future work could focus on developing
reliable and efficient process-level rewards, building upon existing work~\cite{ma2023let, zhang2024entropy, choudhury2025process, wang2025visualprm, li2025process, zhang2025lessons, khalifa2025process}, and potentially even extend them to consider the \textit{structural}
aspects of the reasoning process beyond chains. Since process rewards create many assessment nodes,
their usefulness will depend on performance-oriented policies that involve caching, batching, placement, and overlap with
generation.

\textbf{Harnessing data analytics for enhanced reasoning.}
A rich landscape of research opportunities exist at the overlap of reasoning and data analytics. One could delegate certain parts of reasoning tasks that are hard to instill into model weights to existing algorithms and frameworks; examples are graph mining and analytics~\cite{atluri2018spatio, papakyriakou2022data, besta2022motif, gianinazzi2021parallel, besta2022probgraph}.
One could also harness graph learning for reasoning supervision, i.e., graph neural networks (GNNs)~\cite{kipf2016semi, wu2020comprehensive,
zhang2020deep, zhou2020graph, besta2025demystifyinghigher} and broader graph representation learning methods~\cite{bronstein2017geometric, chami2020machine, besta2023hot,
hamilton2017representation, gianinazzi2021learning} may help score branches, 
help constructing process rewards, or predict useful tool calls, based on learning useful reasoning patterns. The systems challenge is to integrate such designs efficiently~\cite{besta2023parallel, besta2023high} without adding bottlenecks to the Generation--Assessment--Training loop or broader MCTS.

\textbf{Enhancing data analytics pipelines.}
On the other hand, data analytics frameworks~\cite{wang2020deep, gheisari2023data, birjandi2021survey, lan2018survey, gupta2020comprehensive, besta2021practice, besta2021graphminesuite, besta2020substream} as well as databases~\cite{angles2008survey,
angles2018introduction,besta2023graph,davoudian2018survey,
besta2023demystifying,besta2022neural} could also integrate reasoning LLMs and agents for more effective data processing, especially at the user--framework interface. Different designs have already been proposed to integrate LLMs into analytics architectures~\cite{weng2024insightlens, besta2026graphseek, lee2025semantic, chang2025approximating, zeighami2025llm, biswal2024text2sql, chen2023data, qiu2024tqa, dorbani2025beyond}. This includes document analysis (e.g., Aryn~\cite{anderson2024design}, DocETL~\cite{shankar2024docetl}, Palimpzest~\cite{liu2025palimpzest}, PalimpChat~\cite{liu2025palimpchat}), tabular data (e.g., InsightPilot~\cite{ding2023insightpilot}, CoddLLM~\cite{zhang2025coddllm}, Pneuma~\cite{balaka2025pneuma}, Chat2data~\cite{zhao2024chat2data}, LOTUS~\cite{patel2024semanticoperators}, DB-GPT~\cite{xue2024demonstration}), video processing~\cite{10.14778/3685800.3685916}, and others~\cite{devunuri2024transitgpt}.
Here, harnessing RLMs and their efficient integration into such frameworks is an interesting novel research direction.

\textbf{Asynchronous and stale-data RL.}
Online RLM training is limited by long-tailed autoregressive generation.
Asynchronous systems such as AReaL, AsyncFlow, Laminar, StreamRL, and StaleFlow
show that bounded staleness and streaming rollouts
can improve utilization~\cite{fu2025areal,han2025asyncflow,sheng2025laminar,
zhong2025streamrl,li2026staleflow}. The open problem is to characterize in a more rigorous way when and to what degree involve staleness -- for example, when stale rollouts remain useful, how to correct policy drift, and how to trade staleness against throughput without degrading reasoning quality.

\textbf{Retrieval, tools, and external state as operators.} Effective and efficient integration of tools into the reasoning process is another interesting research opportunity~\cite{shen2024llm, liu2025toolace}. Various tool calls (e.g., retrieval~\cite{salemi2024evaluating, gao2023retrieval, jiang2023active, lewis2020retrieval, besta2024multi}, simulators, verifiers~\cite{gu2024survey, besta2024checkembed}, agent calls~\cite{wolflein2025llm, huang2024understanding, chu2025llm, besta2025affordable, besta2025psychologically}, etc.) could be modeled as operators in the same execution graph
as actor generation, reward evaluation, and training. Retrieval-augmented and
graph-based LLM methods such as Topologies of Reasoning provide starting points for such integration~\cite{besta2025demystifyingchains}.

\textbf{Harnessing HPC architectures as well as emerging hardware.}
RLMs stress hardware through aspects such as -- among others -- its long outputs~\cite{nayab2024concise}, large KV caches~\cite{li2025survey, shi2024keep, liu2024minicache}, repeated verifier
calls, tool use, and many samples per prompt. Beyond established forms of parallelism, 
promising directions include harnessing emerging hardware such as processing-in-memory~\cite{ahn2015pim,
ghose2019processing, mutlu2022modern, seshadri2017ambit, besta2021sisa}, RDMA and SmartNICs~\cite{gerstenberger2013enabling, besta2014fault, di2019network, di2022building, besta2015active, schmid2016high, strausz2022asynchronous}, serverless architectures~\cite{copik2021sebs, jonas2019cloud, li2022serverless, mcgrath2017serverless, shafiei2022serverless, hassan2021survey, li2022serverless}, next-generation interconnects~\cite{lakhotia2022polarfly, besta2020fatpaths, besta2020highr, lakhotia2024polarstar}, chiplets~\cite{iff2023hexamesh, feng2022chiplet, mounce2016chiplet, loh2021understanding, ma2022survey, li2020chiplet, iff2025placeit}, and others~\cite{besta2018slim, besta2024hardware, iff2023sparse, iff2026network, gianinazzi2022spatial}.

\textbf{Reasoning for discovering novel optimizations.}
Inspired by the schemes such as AlphaTensor~\cite{fawzi2022discovering} and enabled by the emergence of RLMs with their innate coding capabilities~\cite{rando2025longcodebench, wang2023review, chen2026nlperturbator, bairi2024codeplan, robeyns2025self, dong2025survey}, such models could also help discover novel kernels and other performance-centric schemes, including collectives, placement heuristics, and
routing policies, following recent LLM-driven algorithm-discovery systems such
as AlphaEvolve~\cite{novikov2025alphaevolve}.

\textbf{Evaluation and reproducibility.}
The evaluation of RLMs involves a plethora of aspects such as reasoning budget, sampling strategy, rollout length, staleness, and metrics associated with whole RLM components and subsystems such as tools, retrieval, and verification. It poses an opportunity for designing
effective evaluation pipelines, reusing and extending evaluation methodologies for other domains such as parallel programming~\cite{hoefler2015scientific, ben2019modular}.

\textbf{Shared-weight RLHF and snapshot-consistent rollout workers.}
Another example concrete systems opportunity is to reduce the cost of synchronizing actor
weights in disaggregated or asynchronous RLHF pipelines. Current systems
often push full policy snapshots from the learner to rollout workers, which
creates latency, bandwidth pressure, and synchronization stalls at large model
scale. A promising alternative is a shared-weight design in which rollout
actors access versioned trainer-resident parameter slots, switching only at
rollout boundaries to preserve snapshot consistency. Such a system could use
read-copy-update-style double or triple buffering, version counters, and
publication fences so that actors never observe partially written parameters.
On a single node, this suggests CUDA IPC~\cite{potluri2012optimizing} or peer-to-peer refresh over NVLink
or PCIe~\cite{li2019evaluating}; across nodes, one can harness GPUDirect RDMA~\cite{potluri2013efficient} or NVSHMEM-style one-sided
communication~\cite{hsu2020initial}.

%% file: conclusion.tex
\section{Conclusion}

Reasoning Language Models (RLMs) and their underlying paradigms such as Reinforcement Learning with Verifiable Rewards (RLVR) are not only an algorithmic advance, but also an enormous
parallel and distributed systems challenge due to their compute requirements. Our work systematizes the
RL-for-LLM pipeline, analyzes PPO-, GRPO-, and DPO-like frameworks through
work--depth--memory complexity, and develops a taxonomy of intra- and
inter-model parallelism strategies for scalable RLM training and inference. The
result is a \textbf{unified performance vocabulary} for understanding where computation,
memory, and critical-path bottlenecks arise, and for offering opportunities for performance optimizations while considering performance-critical aspects such as autoregressive rollout generation, auxiliary-model assessment, trainable actor/critic updates, long-context memory,
model-state sharding, placement, fusion, and bounded-stale execution. By making
these trade-offs explicit, our analysis provides a foundation for designing the
next generation of RLM systems. 


%% file: appendix-pipeline.tex
\clearpage
\section{RLM Pipeline: Functional Description}
\label{sec:app:pipeline}

In the RLM pipeline, each stage can be understood as a function acting on batches, with clearly defined inputs and outputs. We now describe these stages in more detail.

\subsection{Generation Stage}

The {Generation stage} takes as input a batch of prompts \(X=\{x_1,\ldots,x_B\} \subset \mathcal{X} \) and the behavior policy \(\pi_{\theta_{\mathrm{old}}}\), which either comes from the previous RL iteration, or -- for the 1st iteration -- is the base model. Here, $\mathcal{X}$ is the space of input prompts. For each input prompt $x_b \in \mathcal{X}$, the policy model $\pi_\theta(y \mid x_b)$ generates a set of $K$ candidate completions $\mathrm{Gen}_\theta(x_b) = \left\{ y_b^{(i)} \right\}_{i=1}^K.$ The $i$-th candidate response to a prompt $x_b$ is denoted as $y_b^{(i)} \in \mathcal{Y}$ and it is a sequence of $T$ tokens  $\left(y_{b,1}^{(i)}, \dots, y_{b,T}^{(i)}\right)$, where $y_{b,t}^{(i)}$ represents a token at time-step $t$; $\mathcal{Y}$ is the space of token sequences. The output is thus the rollout batch \(\mathcal{B}_{\mathrm{roll}} = \left\{\left(x_b,y_b^{(i)}\right)\right\}_{b,i}\). Along with the sampled outputs, PPO-like methods often also compute per-token actor log-probabilities $\log \pi_\theta\left(y_{b,t}^{(i)} \mid x_b, y_{b,<t}^{(i)}\right)$ for each step $t \in \{1, \dots, T\}$, for use in the training stage.


\subsection{Assessment Stage}

The {Assessment stage} takes the rollout batch \(\mathcal{B}_{\mathrm{roll}}\) and evaluates it with the auxiliary models required by the algorithm. In PPO-like methods, this usually includes the reward model \(R_\varphi\), reference policy \(\pi_{\mathrm{ref}}\), and critic \(V_\psi\) that are used to transform Generation's output into numerical learning signals. The most important such signals are \textbf{sequence-level scalar rewards} $r_b^{(i)}$: a single quality score assigned to each complete sequence $y_b^{(i)}$, typically computed by the frozen reward model $R_\varphi \left(x_b, y_b^{(i)}\right)$. These scores quantify alignment with human preferences, such as helpfulness or safety.
Moreover, the stage also produces \textbf{token-level training targets} $A_{b}^{(i)} = \left( A_{b,1}^{(i)} A_{b,2}^{(i)} ... A_{b,T}^{(i)} \right)$; each such target is a dense vector of values for each token position $t \in \{1, \dots, T\}$. These values, commonly referred to as \emph{advantages}, enable granular credit assignment by comparing the empirical return $\hat{G}_{b,t}^{(i)}$ against the estimate return by the critic model $V_\psi \left(x_b, y_{b,<t}^{(i)} \right)$. They are constructed from returns/rewards and, for PPO-like actor--critic methods, and critic value estimates, e.g., \(A_{b,t}^{(i)}=\hat G_{b,t}^{(i)}-v_{b,t,i}\). In critic-free methods such as GRPO, no \(V_\psi\) pass is performed; advantages are computed from group-normalized rewards instead.
Finally, the Assessment stage also involves computing \textbf{reference log-probabilities} \(\log\pi_{\mathrm{ref}}\left(y_{b,t}^{(i)}\mid x_b,y_{b,<t}^{(i)}\right)\).

\subsection{Training Stage}

The \emph{Training stage} takes the rollout batch and the learning signals from
Assessment: rewards, reference log-probabilities, stored sample-time actor
log-probabilities, and, when applicable, advantages and value targets. It updates
the trainable model components required by the algorithm. In PPO, the actor uses
policy-ratio terms such as \(\pi_\theta\left(y_{b,t}^{(i)}\mid x_b,y_{b,<t}^{(i)}\right)/
\pi_{\theta_{\mathrm{old}}}\left(y_{b,t}^{(i)}\mid x_b,y_{b,<t}^{(i)}\right)\), while the
critic is updated with a value-regression loss. In GRPO, only the actor is
updated because no learned critic is used. In DPO, Training operates on
preference pairs rather than online rollouts and updates only the policy using
teacher-forced likelihoods of preferred and dispreferred responses. Thus,
Training outputs the next policy \(\pi_{\theta'}\), and updated critic
parameters \(\psi'\) only for algorithms that train a critic.


\subsection{Iterative Loop}

This process is repeated in an iterative loop: after the policy update, the new parameters $\theta' \gets \theta$ are used in the next Generation stage. The full loop therefore evolves as $\theta \rightarrow \mathrm{Gen}_\theta(X) \rightarrow \mathrm{Assess}(X, Y) \rightarrow \mathrm{Train}(\cdot) \rightarrow \theta' \rightarrow \dots$ and continues until convergence or until a predetermined number of iterations is reached.

%% file: appendix-derivations.tex
\clearpage
\section{Details of Mathematical Derivations}
\label{sec:app:derivations}

We offer details of mathematical derivations. To simplify equations, we use $C_f^{\text{tok}} \equiv C_f$, $C_b^{\text{tok}} \equiv C_b$, and $C_{\text{gen}}^{\text{roll}} \equiv C_{\text{gen}}$.

\subsection{Results for Building Blocks}
\label{sec:app:derivations-building-blocks}

We now derive the model-specific building blocks used in Table~\ref{tab:building-blocks} for forward and backward passes in terms of architectural parameters such as the number of layers $L$, hidden size $d$, feed-forward width $d_{\mathrm{ff}}$, vocabulary size $V$, and sequence lengths $S,T$. We parameterize these values with a given model $M$ in question, i.e., hidden size $d_M$, feed-forward width $d_{M,\mathrm{ff}}$, and $L_M$ layers.

\subsubsection{Forward-Pass FLOPs}

We begin with a standard dense Transformer layer. For a model $M$, one layer contains \textbf{four dense attention projections}: $W_Q,W_K,W_V,W_O \in \mathbb{R}^{d_M \times d_M}$, contributing $4d_M^2$ parameters, and \textbf{two FFN projections}, $W_{\mathrm{up}} \in \mathbb{R}^{d_M \times d_{M,\mathrm{ff}}}$ and $W_{\mathrm{down}} \in \mathbb{R}^{d_{M,\mathrm{ff}} \times d_M}$, contributing $2d_M d_{M,\mathrm{ff}}$ parameters.
Thus, one layer contains
\[
4d_M^2 + 2d_M d_{M,\mathrm{ff}}
\]
trainable projection parameters.

Under the standard FLOP accounting for matrix multiplication, each parameter contributes approximately two FLOPs per processed token in the forward pass (one multiply and one add). Hence, the parameter-dependent forward FLOPs per token per layer are
\[
2\bigl(4d_M^2 + 2d_M d_{M,\mathrm{ff}}\bigr).
\]
Across $L_M$ layers, this becomes
\[
2L_M\bigl(4d_M^2 + 2d_M d_{M,\mathrm{ff}}\bigr).
\]

In addition, self-attention performs parameter-free token-mixing operations, namely score computation and value aggregation. Under the same coarse cost model used throughout the paper, these contribute
\[
2L_M (S+T)d_M
\]
FLOPs per token on a sequence of total length $S+T$.

Therefore, the base forward-pass cost per token for model $M$ is
\[
C_f(M)
=
2L_M\bigl(4d_M^2 + 2d_M d_{M,\mathrm{ff}} + (S+T)d_M\bigr),
\]
up to model-specific output heads.

\paragraph{\textbf{Policy and reference model}}
We parameterize $M \in \{\pi, \text{ref}\}$. For the policy model $\pi_\theta$, one additionally computes logits over the vocabulary, i.e. an unembedding/output projection of size $d_\pi \to V$. Under the same multiply--add accounting, this contributes
\[
2Vd_\pi
\]
FLOPs per token. Hence
\[
C_f(\pi_\theta)
=
2L_\pi\bigl(4d_\pi^2 + 2d_\pi d^\pi_{\mathrm{ff}} + (S+T)d_\pi\bigr)
+ 2Vd_\pi.
\]
Because the reference model $\pi_{\mathrm{ref}}$ is architecturally identical to the actor,
\[
C_f(\pi_{\mathrm{ref}})=C_f(\pi_\theta).
\]

\paragraph{\textbf{Reward model}}
Now, $M = R$. The reward model $R_\phi$ processes the full sequence but produces a single scalar score from the final hidden state. Its transformer backbone therefore costs
\[
2L_R\bigl(4d_R^2 + 2d_R d^R_{\mathrm{ff}} + (S+T)d_R\bigr).
\]
The scalar reward head is a projection $d_R \to 1$ applied once per sequence, i.e. $2d_R$ FLOPs per sequence, or equivalently
\[
\frac{2d_R}{S+T}
\]
FLOPs per token when normalized by sequence length. Thus,
\[
C_f(R_\phi)
=
2L_R\bigl(4d_R^2 + 2d_R d^R_{\mathrm{ff}} + (S+T)d_R\bigr)
+ \frac{2d_R}{S+T}.
\]

\paragraph{\textbf{Value model / critic}}
Finally, $M = V$. The critic $V_\psi$ outputs one scalar value per token. Its transformer backbone contributes
\[
2L_V\bigl(4d_V^2 + 2d_V d^V_{\mathrm{ff}} + (S+T)d_V\bigr),
\]
and its value head $d_V \to 1$ is applied at every token position, contributing
\[
2d_V
\]
FLOPs per token. Hence
\[
C_f(V_\psi)
=
2L_V\bigl(4d_V^2 + 2d_V d^V_{\mathrm{ff}} + (S+T)d_V\bigr)
+ 2d_V.
\]

\subsubsection{Backward-pass FLOPs}

For dense Transformer training, a standard approximation is that the backward pass costs about twice the forward pass, because gradients must be propagated both with respect to activations and with respect to weights. Accordingly, for any trainable model $M$ we use
\[
C_b(M)\approx 2\,C_f(M).
\]
This is the approximation used throughout the paper when deriving training-stage costs.

\subsubsection{Autoregressive generation: prefill and decode}

For policy generation, it is important to distinguish prefill from decode.

\paragraph{\textbf{Prefill}}
Given a prompt of length $S$, the prefill pass processes the prompt once, initializes the KV cache, and produces the logits needed to begin generation. Therefore, its cost is simply the forward cost evaluated at sequence length $S$:
\begin{align*}
C_{\mathrm{prefill}}(\pi_\theta;S)
=
2L_\pi S \bigl(4d_\pi^2 + 2d_\pi d^\pi_{\mathrm{ff}} + S d_\pi\bigr)
+ 2Vd_\pi \\
L_\pi \left[ O(S^2) + O(S d_\pi) \right].
\end{align*}

\paragraph{\textbf{Decode}}
After prefill, the model generates tokens autoregressively. At decode step $t$, the current token attends to a context of length $S+t$, consisting of the prompt plus the $t$ previously generated tokens. Hence the cost of the full decode phase is
\small
\[
C_{\mathrm{dec}}(\pi_\theta;S,T)
=
\sum_{t=1}^{T-1}
\left[
2L_\pi\bigl(4d_\pi^2 + 2d_\pi d^\pi_{\mathrm{ff}} + (S+t)d_\pi\bigr)
+ 2Vd_\pi
\right].
\]
\normalsize
Using
\[
\sum_{t=1}^{T-1}(S+t)=(T-1)S+\frac{(T-1)T}{2},
\]
this can be written as
\begin{align*}
C_{\mathrm{dec}}(\pi_\theta;S,T)
= &\ 2(T-1)L_\pi\bigl(4d_\pi^2 + 2d_\pi d^\pi_{\mathrm{ff}}\bigr)\\
+ &\ 4L_\pi d_\pi\!\left((T-1)S+\frac{(T-1)T}{2}\right)\\
+ &\ 2(T-1)Vd_\pi.
\end{align*}

\paragraph{\textbf{Total generation cost}}
The total generation cost is therefore
\[
C_{\mathrm{gen}}(\pi_\theta;S,T)
=
C_{\mathrm{prefill}}(\pi_\theta;S)
+
C_{\mathrm{dec}}(\pi_\theta;S,T).
\]
This is the quantity that should be used in the most detailed generation accounting. For more compact asymptotic comparisons, one may further summarize it as
\[
C_{\mathrm{gen}}(\pi_\theta;S,T)=O\!\bigl((S+T)\,C_f(\pi_\theta;S+T)\bigr),
\]
but this coarser form should be understood as an upper-level abstraction of the explicit prefill+decode decomposition above.

\subsubsection{Depth}

We now derive the depth terms used in Table~\ref{tab:building-blocks}. Here, depth means the critical-path length under unbounded parallelism.

\paragraph{\textbf{Single forward pass}}
Within one Transformer layer, the dense projections in the FFN contribute reduction depth
\[
\log d_M + \log d_{M,\mathrm{ff}}.
\]

For the multi-head attention block, the depth comes from the sequential dependencies among the projection, attention, aggregation, and output-projection subcomputations. The query, key, and value projections are independent and can be computed in parallel; therefore, they contribute only $\log d_M$ to the critical path. Once \(Q\) and \(K\) are available, the attention-score computation within each head reduces over the per-head hidden dimension \(d_M/h_M\), contributing
\[
\log(d_M/h_M)=\log d_M-\log h_M.
\]
The subsequent weighted aggregation of values reduces over the sequence length, contributing \(\log(S+T)\). Finally, the attention output projection maps the concatenated heads back to the hidden dimension and contributes another \(\log d_M\). Thus, suppressing constant factors, the multi-head attention depth is
\begin{gather*}
    \log d_M + \log(d_M/h_M) + \log(S+T) + \log d_M
= \\
3\log d_M+\log(S+T)-\log h_M.
\end{gather*}

Since layers are sequential, the total forward depth scales linearly with $L_M$. In the simplified cost model used in Table~\ref{tab:building-blocks}, this is summarized as
\[
D_f(M)=O\!\bigl(L_M[\log d_M + \log(S+T)]\bigr),
\]
plus model-specific output-head terms.

For the \textbf{policy/reference model}, the additional vocabulary projection contributes one more logarithmic reduction, yielding
\begin{align*}
D_f(\pi_\theta)&=O\!\bigl(L_\pi[\log d_\pi+\log(S+T)] + \log d_\pi\bigr),\\
D_f(\pi_{\mathrm{ref}})&=D_f(\pi_\theta).
\end{align*}
For the \textbf{reward} and \textbf{value models}, the scalar head contributes an additional $\log d$ term:
\[
D_f(R_\phi)=O\!\bigl(L_R[\log d_R+\log(S+T)] + \log d_R\bigr),
\]
\[
D_f(V_\psi)=O\!\bigl(L_V[\log d_V+\log(S+T)] + \log d_V\bigr).
\]

\paragraph{\textbf{Tensor-parallel forward depth}}
If tensor parallelism of degree $P_t$ shards the hidden dimension, the corresponding reduction depth decreases accordingly, giving
\[
D_f^{\mathrm{TP}}(M)
=
O\!\bigl(L_M[\log(d_M/P_t)+\log(S+T)] + \log(d_M/P_t)\bigr).
\]

\paragraph{\textbf{Backward depth}}
Under the same coarse-grained model, the backward pass has a different dependency structure to that of the forward pass. Namely, for any product of activations with model weights, one must consider computing gradients with respect to both the activations and the weights. Computing the gradients with respect to activations has the same asymptotic dependency structure as in the forward pass. Computing the gradients with respect to model weights, however, has to accumulate gradients across the whole batch. Thus
\begin{gather*}
D_b(M)=D_f(M) + O(\log (B (S+T))) \\ = O(L ( \log d_M + \log (S+T)) + \log (B (S+T))).
\end{gather*}

As gradients with respect to weights do not need to be backpropagated to the previous layer, $\log (B(S+T))$ does not have to be multiplied with $L$. 

\paragraph{\textbf{Generation depth}}
Autoregressive generation has two components: one prefill pass plus a sequence of sequential decode steps. Therefore,
\[
D_{\mathrm{gen}}(\pi_\theta;S,T)
=
D_f(\pi_\theta;S)
+
\sum_{t=1}^{T-1} D_f(\pi_\theta;S+t).
\]
In compact asymptotic form, this is
\[
D_{\mathrm{gen}}(\pi_\theta;S,T)
=
O\!\bigl(T\cdot D_f(\pi_\theta;S+T)\bigr),
\]
which makes explicit that the dominant sequential dependence comes from the $T$ decode steps.

\subsubsection{Parameter counts}

We finally derive the parameter-count and memory expressions used in Table~\ref{tab:building-blocks}.

\paragraph{\textbf{Policy/reference model}}
Ignoring biases and layer-norm parameters, the actor contains
\[
|\pi_\theta|
=
L_\pi(4d_\pi^2 + 2d_\pi d^\pi_{\mathrm{ff}}) + Vd_\pi
\]
parameters: the first term comes from the stacked Transformer blocks, and the second from token embeddings / output vocabulary projection. Since the reference model shares the same architecture,
\[
|\pi_{\mathrm{ref}}|=|\pi_\theta|.
\]

\paragraph{\textbf{Shared actor--critic backbone}}
For a shared actor--critic model with backbone width $d_{\mathrm{ac}}$ and $L_{\mathrm{ac}}$ layers, plus a policy head of size $Vd_{\mathrm{ac}}$ and a scalar value head of size $d_{\mathrm{ac}}$, we obtain
\[
|\pi_{\mathrm{ac}}|
=
L_{\mathrm{ac}}(4d_{\mathrm{ac}}^2 + 2d_{\mathrm{ac}}d^{\mathrm{ac}}_{\mathrm{ff}})
+ Vd_{\mathrm{ac}} + d_{\mathrm{ac}}.
\]

\paragraph{\textbf{Reward model}}
The reward model consists of a Transformer backbone plus a scalar reward head:
\[
|R_\phi|
=
L_R(4d_R^2 + 2d_R d^R_{\mathrm{ff}}) + Vd_R + d_R.
\]

\paragraph{\textbf{Value model / critic}}
Similarly, the value model consists of a Transformer backbone plus a scalar head:
\[
|V_\psi|
=
L_V(4d_V^2 + 2d_V d^V_{\mathrm{ff}}) + Vd_V + d_V.
\]

\paragraph{\textbf{KV-cache memory}}
During generation, each layer stores both keys and values for every processed token. For one rollout of total length $S+T$, this yields
\[
M_{\mathrm{KV}} = 2(S+T)L_\pi d_\pi,
\]
where the factor $2$ accounts for keys and values. If all $BK$ rollouts are generated concurrently, the corresponding peak KV-cache memory is
\[
BK \cdot M_{\mathrm{KV}}.
\]
If fewer rollouts are generated concurrently, the peak memory decreases proportionally, at the cost of reduced concurrency.

\paragraph{\textbf{Optimizer state}}
For trainable models, optimizer-state memory is approximated by a factor of four times the parameter count, corresponding to parameters, gradients, and Adam first and second moments. 

\paragraph{\textbf{Training activations and inference buffers}}

For a trainable model \(M\), backpropagation requires intermediate activations
from the forward pass. Under the leading-order activation model used throughout
the paper, we retain one hidden-state-sized contribution per layer and processed
token:
\[
M_{\mathrm{Act}}(M)=\Theta(L_M d_M)
\]
per token. Hence, for a batch of \(Q\) sequences of length \(S+T\),
\[
M_{\mathrm{Act,total}}(M)
=
\Theta\!\left(Q(S+T)L_Md_M\right).
\]
This expression suppresses constant-factor storage for Q/K/V tensors, FFN
intermediates, normalization and residual temporaries, and other
implementation-specific buffers, and assumes no activation checkpointing.

For a frozen forward-only model, intermediate layer buffers need not be retained
for backpropagation and can be reused across layers. Assuming the attention
matrix is not fully materialized, the leading-order inference buffer is
\[
M_{\mathrm{Inf}}(M)=\Theta(d_M)
\]
per processed token, or
\[
M_{\mathrm{Inf,total}}(M)
=
\Theta\!\left(Q(S+T)d_M\right)
\]
for \(Q\) sequences. Model parameters and autoregressive KV-cache storage are
accounted for separately.


\subsection{Results for RL-LLM Frameworks}

We start with derivations for results in Section~\ref{sec:comp-anal-frameworks}.

\subsubsection{Online Frameworks}

Online methods such as PPO and GRPO repeatedly execute the full
Generation $\rightarrow$ Assessment $\rightarrow$ Training loop, so their
per-iteration cost is the sum of these three stages.

\paragraph{\textbf{Generation}}
Generation is dominated by autoregressive sampling with the current
policy $\pi_\theta$. For each of the $B$ prompts, the policy generates
$K$ candidate responses, yielding $BK$ rollouts in total. The cost of one
rollout should be decomposed into a \emph{prefill} pass over the prompt of
length $S$ and a sequence of $T$ \emph{decode} steps:
\[
C_{\mathrm{gen}}(\pi_\theta)
=
C_{\mathrm{prefill}}(\pi_\theta; S)
+
\sum_{t=1}^{T-1} C_{\mathrm{decode}}(\pi_\theta; S+t).
\]
Here, the prefill initializes the KV cache from the prompt, while decode
step $t$ produces the next token while attending to a context of length
$S+t$. Thus, the total generation work over the whole batch is
\[
W_{\mathrm{gen}}
=
\Theta\!\bigl(BK \cdot C_{\mathrm{gen}}(\pi_\theta)\bigr).
\]

The critical path is determined by the autoregressive dependence across
generated tokens: token $t+1$ cannot be produced before token $t$ has been
generated. Therefore, the generation depth scales linearly in $T$:
\[
D_{\mathrm{gen}}
=
\Theta\!\bigl(T \cdot D_f(\pi_\theta)\bigr).
\]
This sequential dependence makes generation the primary wall-clock
bottleneck in online RL-LLM pipelines.

\paragraph{\textbf{Assessment}}
After generation, the completed prompt--response pairs are evaluated by
auxiliary models such as the reward model $R_\phi$, the critic $V_\psi$,
and the reference policy $\pi_{\mathrm{ref}}$. For PPO, all three are used;
for GRPO, the critic is omitted. Each model processes the full sampled
sequences in teacher-forced, non-autoregressive mode, so the work per model
scales as
\[
W_{\mathrm{assess}}(M)=\Theta\!\bigl(BK(S+T)\,C_f(M)\bigr).
\]
Unlike generation, this stage does \emph{not} incur a factor-$T$
sequential dependence across output tokens. Its critical path is therefore
the depth of a single forward pass through the corresponding model:
\[
D_{\mathrm{assess}}(M)=\Theta\!\bigl(D_f(M)\bigr).
\]
If the assessment models are placed on disjoint device groups and executed
concurrently, the stage depth is the maximum of their forward-pass depths;
if they are co-located and executed sequentially, these depths add.

\paragraph{\textbf{Training}}
In the training stage, the policy and, for actor--critic methods, the
critic are updated by backpropagation over the sampled sequences. Since
training is performed in teacher-forced mode on the complete
prompt--response sequences, the work per updated model and per optimization epoch scales as
\[
W_{\mathrm{train}}(M)=\Theta\!\bigl(BK(S+T)\,C_{tr}(M)\bigr).
\]
Again, there is no autoregressive factor-$T$ dependence across tokens in the
critical path; instead, the depth is that of a single backward pass:
\[
D_{\mathrm{train}}(M)=\Theta\!\bigl(D_{tr}(M)\bigr).
\]
Thus, PPO incurs updates for both $\pi_\theta$ and $V_\psi$, whereas GRPO
updates only $\pi_\theta$.

\subsubsection{Offline Frameworks}

Offline frameworks such as DPO decouple generation from optimization.
Instead of sampling fresh rollouts inside the training loop, they optimize
the policy $\pi_\theta$ on a static dataset of preference pairs
$(x, y_w, y_l)$. As a result, the inner loop contains only batched forward
and backward passes over fixed data.

For a batch of $B_{\mathrm{pairs}}$ preference pairs, the forward work is
\[
W_{\mathrm{forward}}
=
\Theta\!\bigl(B_{\mathrm{pairs}}(S+T)\,[C_f(\pi_\theta)+C_f(\pi_{\mathrm{ref}})]\bigr),
\]
while the backward work is
\[
W_{\mathrm{backward}}
=
\Theta\!\bigl(B_{\mathrm{pairs}}(S+T)\,C_b(\pi_\theta)\bigr).
\]
The critical path is therefore the depth of a single batched forward/backward
evaluation rather than an autoregressive decode chain:
\[
D_{\mathrm{DPO}}
=
\Theta\!\bigl(\max\{D_f(\pi_\theta), D_f(\pi_{\mathrm{ref}}), D_b(\pi_\theta)\}\bigr).
\]

\subsection{Results for Intra-Model Parallelism}
\label{sec:app:derivations-intra}

We now derive the results for
Section~\ref{sec:comp-anal-intra}
(Tables~\ref{tab:intra_model_parallelism_local},
\ref{tab:intra_model_parallelism_global},
\ref{tab:intra_model_parallelism_exact_local}, and
\ref{tab:intra_model_parallelism_exact_global}).
The goal is to expose how standard intra-model parallelism strategies transform
the work, depth, and memory terms derived in
Appendix~\ref{sec:app:derivations-building-blocks}.

\subsubsection{Baseline Cost Model}

Appendix~\ref{sec:app:derivations-building-blocks} derives the forward and
backward costs of Transformer-based models in terms of layer count, hidden
dimension, FFN width, vocabulary size, and sequence length. For the present
intra-model analysis, we consider a dense Transformer training iteration with
\(L\) layers, hidden dimension \(d\), FFN dimension \(d_{\mathrm{ff}}\),
vocabulary size \(V\), prompt length \(S\), response length \(T\), and batch
size \(B\). The complete teacher-forced training sequence therefore has length
\(S+T\). Since
Tables~\ref{tab:intra_model_parallelism_local}--%
\ref{tab:intra_model_parallelism_exact_global}
analyze a generic Transformer training step rather than a role-specific actor,
critic, reward, or reference model, we ignore small model-specific head
differences and use the common dense Transformer body.

From Appendix~\ref{sec:app:derivations-building-blocks}, the forward-pass cost
per token of the dense Transformer body is
\[
C_f
=
2L\left(4d^2+2dd_{\mathrm{ff}}+(S+T)d\right),
\]
where \(4d^2\) comes from the query, key, value, and output projections,
\(2dd_{\mathrm{ff}}\) from the two FFN projections, and \((S+T)d\) from the
sequence-dependent attention score/value operations. Using the same
approximation as in Table~\ref{tab:building-blocks},
\[
C_b\approx2C_f.
\]
Thus, the forward-plus-backward work per token is
\[
C_f+C_b
\approx
3C_f
=
6L\left(4d^2+2dd_{\mathrm{ff}}+(S+T)d\right).
\]
For \(B(S+T)\) processed tokens, the baseline work is
\[
W_{\mathrm{base}}
=
6B(S+T)L
\left(4d^2+2dd_{\mathrm{ff}}+(S+T)d\right),
\]
which gives the asymptotic summary
\[
W_{\mathrm{base}}
=
O\!\left(B(S+T)L(d^2 + (S+T)d)\right)
\]
in the parameter-dominated regime.

We next refine the depth expression. A single forward Transformer layer has
reduction depth
\[
4\log d+\log d_{\mathrm{ff}}+\log(S+T)-\log h,
\]
where \(h\) is the number of attention heads. Hence,
\[
D_f
=
L\left(
4\log d+\log d_{\mathrm{ff}}+\log(S+T)-\log h
\right).
\]

The activation-gradient path in the backward pass has a comparable layerwise
dependency structure. However, trainable projections also require parameter
gradients. For a projection \(Y=XW\), with the batch and sequence dimensions
flattened into \(B(S+T)\) rows,
\[
\nabla_X\mathcal{L}
=
\nabla_Y\mathcal{L}\,W^\top,
\qquad
\nabla_W\mathcal{L}
=
X^\top\nabla_Y\mathcal{L}.
\]
The second product accumulates contributions to each shared parameter over the
\(B(S+T)\) batch-token positions and therefore introduces a reduction of depth
\[
\log(B(S+T)).
\]
These parameter-gradient reductions branch from the activation-gradient
backward path and can overlap with propagation through preceding layers;
therefore they are not multiplied by \(L\). We conservatively account for this
additional critical-path contribution as
\[
D_b
=
O\!\left(
L[\log d+\log(S+T)]
+
\log(B(S+T))
\right).
\]
Under the explicit reduction-depth model used in
Tables~\ref{tab:intra_model_parallelism_exact_local}--%
\ref{tab:intra_model_parallelism_exact_global}, we therefore use
\begin{align*}
D_{\mathrm{base}}
&=
2L\left(
4\log d+\log d_{\mathrm{ff}}
+\log(S+T)-\log h
\right)\\
& \quad +\log(B(S+T))
\\
&=
2L\log\!\left(
\frac{d^4d_{\mathrm{ff}}(S+T)}{h}
\right)
+\log(B(S+T)).
\end{align*}
The corresponding asymptotic form is
\[
D_{\mathrm{base}}
=
O\!\left(
L[\log d+\log(S+T)]
+\log(B(S+T))
\right).
\]

Finally, the memory model follows the same parameter-counting convention as
Appendix~\ref{sec:app:derivations-building-blocks}. The dominant model-state
terms are parameters, gradients, and Adam first and second moments, giving a
factor of four over the parameter count:
\[
4L(4d^2+2dd_{\mathrm{ff}})+4Vd
=
L(16d^2+8dd_{\mathrm{ff}})+4Vd.
\]
Stored activations contribute
\[
B(S+T)Ld.
\]
Therefore,
\[
M_{\mathrm{base}}
=
L(16d^2+8dd_{\mathrm{ff}})
+4Vd
+B(S+T)Ld,
\]
or asymptotically
\[
M_{\mathrm{base}}
=
O\!\left(
Ld^2+Vd+B(S+T)Ld
\right).
\]

The remaining derivations apply each intra-model parallelism strategy as a
partitioning transformation of these baseline work, depth, and memory terms.

\subsubsection{Data Parallelism}

Data parallelism of degree \(P_d\) partitions the batch dimension. Each rank
processes \(B/P_d\) sequences but stores a full model replica. Therefore, the
per-rank work is
\begin{align*}
W_{\mathrm{DP,rank}}
&=
\frac{1}{P_d}W_{\mathrm{base}}\\
&=
6\frac{B}{P_d}(S+T)L
\left(4d^2+2dd_{\mathrm{ff}}+(S+T)d\right),
\end{align*}
while the global work remains
\[
W_{\mathrm{DP,global}}=W_{\mathrm{base}}.
\]

The layerwise forward/backward dependency is unchanged. Locally, however, each
rank forms parameter gradients from only
\(\frac{B}{P_d}(S+T)\) batch-token positions. Hence,
\[
D_{\mathrm{DP,rank}}
=
2L\log\!\left(
\frac{d^4d_{\mathrm{ff}}(S+T)}{h}
\right)
+
\log\!\left(
\frac{B}{P_d}(S+T)
\right).
\]
The globally aggregated parameter gradient still depends on all \(B(S+T)\)
positions, so, ignoring communication latency while retaining the logical
gradient-reduction dependency,
\[
D_{\mathrm{DP,global}}
=
D_{\mathrm{base}}.
\]

The per-rank activation memory is reduced by \(P_d\), but model-state memory is
replicated:
\[
M_{\mathrm{DP,rank}}
=
L(16d^2+8dd_{\mathrm{ff}})
+4Vd
+\frac{B}{P_d}(S+T)Ld.
\]
At the global level, the activation term sums back to \(B(S+T)Ld\), while the
model-state term is replicated \(P_d\) times:
\[
M_{\mathrm{DP,global}}
=
P_d\!\left[
L(16d^2+8dd_{\mathrm{ff}})+4Vd
\right]
+B(S+T)Ld.
\]
Thus, DP reduces per-rank work and activation memory, while the full logical
model update retains the global batch-gradient dependency.

\subsubsection{Pipeline Parallelism}

Pipeline parallelism of degree \(P_p\) partitions the layer dimension. Each
pipeline stage stores and computes approximately \(L/P_p\) layers. Applying
\[
L\mapsto\frac{L}{P_p}
\]
to the baseline per-rank formulas gives
\[
W_{\mathrm{PP,rank}}
=
6B(S+T)\frac{L}{P_p}
\left(
4d^2+2dd_{\mathrm{ff}}+(S+T)d
\right),
\]
and
\[
D_{\mathrm{PP,rank}}
=
2\frac{L}{P_p}
\log\!\left(
\frac{d^4d_{\mathrm{ff}}(S+T)}{h}
\right)
+
\log(B(S+T)).
\]
The layer-dependent memory terms are similarly divided across pipeline stages:
\[
M_{\mathrm{PP,rank}}
=
\frac{L}{P_p}(16d^2+8dd_{\mathrm{ff}})
+
4Vd
+
B(S+T)\frac{L}{P_p}d.
\]
The vocabulary term \(4Vd\) is written separately because embeddings and output
heads are commonly pinned to boundary stages rather than evenly partitioned
across all stages.

Globally, pipeline parallelism preserves total work and total model-state
memory up to boundary-stage effects:
\[
W_{\mathrm{PP,global}}=W_{\mathrm{base}},
\qquad
M_{\mathrm{PP,global}}\approx M_{\mathrm{base}}.
\]
For global depth, a microbatch still traverses all \(P_p\) stages and hence all
\(L\) layers. Therefore,
\[
D_{\mathrm{PP,global}}
=
D_{\mathrm{base}}.
\]
Realized runtime additionally depends on the microbatch schedule, pipeline
fill/drain bubbles, and whether one counts the latency of a single microbatch
or steady-state throughput. Those hardware-schedule effects are outside the
present work--depth abstraction.

\subsubsection{Tensor Parallelism}

Tensor parallelism of degree \(P_t\) partitions hidden-dimensional operators
within each layer. In the simplified work--depth model, this shards the dominant
GEMM work, model-state memory, and activation memory by \(P_t\). Thus,
\[
W_{\mathrm{TP,rank}}
=
6B(S+T)L
\left(
\frac{4d^2+2dd_{\mathrm{ff}}+(S+T)d}{P_t}
\right),
\]
and
\[
M_{\mathrm{TP,rank}}
=
\frac{L(16d^2+8dd_{\mathrm{ff}})}{P_t}
+
\frac{4Vd}{P_t}
+
\frac{B(S+T)Ld}{P_t}.
\]

The hidden-dimensional reductions are correspondingly shortened on each tensor
shard, while the parameter-gradient accumulation still spans all
\(B(S+T)\) batch-token positions. Under the explicit model used in the tables,
\[
D_{\mathrm{TP}}
=
2L
\log\!\left(
\frac{d^4d_{\mathrm{ff}}(S+T)}{P_t^2h}
\right)
+
\log(B(S+T)).
\]
Asymptotically,
\[
D_{\mathrm{TP}}
=
O\!\left(
L[\log(d/P_t)+\log(S+T)]
+\log(B(S+T))
\right).
\]

Globally, TP preserves total work and total memory order:
\[
W_{\mathrm{TP,global}}=W_{\mathrm{base}},
\qquad
M_{\mathrm{TP,global}}\approx M_{\mathrm{base}},
\]
ignoring communication buffers and collective overheads. Thus, TP primarily
reduces per-device pressure and hidden-dimensional critical-path reductions,
at the cost of layer-wise communication not modeled in these tables.

\subsubsection{Context Parallelism}

Context parallelism of degree \(P_c\) partitions the sequence dimension. Each
rank owns a local query/context shard of length \((S+T)/P_c\). Each local query,
however, still depends on keys and values from the global context of length
\(S+T\). Thus, CP partitions sequence-local work and activations without
replacing the global attention-reduction length \(S+T\) by
\((S+T)/P_c\).

The per-rank work is
\[
W_{\mathrm{CP,rank}}
=
6B\frac{S+T}{P_c}L
\left(
4d^2+2dd_{\mathrm{ff}}+(S+T)d
\right).
\]
The per-rank memory is
\[
M_{\mathrm{CP,rank}}
=
L(16d^2+8dd_{\mathrm{ff}})
+
4Vd
+
B\frac{S+T}{P_c}Ld.
\]

Under the computation-only depth abstraction, the attention part retains the
global \(S+T\) reduction dependency. The local parameter-gradient reduction is
over \(B(S+T)/P_c\) token instances, giving
\[
D_{\mathrm{CP,rank}}
=
2L
\log\!\left(
\frac{d^4d_{\mathrm{ff}}(S+T)}{h}
\right)
+
\log\!\left(
\frac{B(S+T)}{P_c}
\right).
\]
Globally, the partial parameter gradients span all sequence partitions, so
\[
D_{\mathrm{CP,global}}
=
D_{\mathrm{base}}.
\]
Distributed attention additionally introduces communication-round
dependencies, which are outside the present FLOP-based depth abstraction.

Globally, model parameters are replicated across context partitions, while the
activation term sums back to the original order:
\[
W_{\mathrm{CP,global}}=W_{\mathrm{base}},
\]
\[
M_{\mathrm{CP,global}}
=
P_c\!\left[
L(16d^2+8dd_{\mathrm{ff}})+4Vd
\right]
+
B(S+T)Ld.
\]
Thus, CP is primarily a sequence-work, memory, and long-context feasibility
technique; it does not reduce the global exact-attention reduction from
\(S+T\) to \((S+T)/P_c\).

\subsubsection{Expert Parallelism}

Expert parallelism applies to MoE layers. Suppose the model has \(E\) experts,
\(E_a\) active experts per token, expert hidden dimension \(d_e\), and
expert-parallel degree \(P_e\).

For the isolated EP results in
Tables~\ref{tab:intra_model_parallelism_local}--%
\ref{tab:intra_model_parallelism_exact_global},
we assume that EP is the only partitioning mechanism: the non-expert
Transformer computation is replicated across the \(P_e\) ranks, while expert
FFN parameters and routed expert computation are partitioned across them.
Thus, no additional DP, TP, PP, or CP partitioning is assumed outside the
expert FFN.

The per-rank work is therefore
\[
W_{\mathrm{EP,rank}}
=
6B(S+T)L
\left(
4d^2
+
\frac{2E_a d d_e}{P_e}
+
(S+T)d
\right).
\]
Because the dense/shared computation is replicated, while expert computation
is partitioned, global executed work is
\[
W_{\mathrm{EP,global}}
=
6B(S+T)L
\left(
P_e[4d^2+(S+T)d]
+
2E_a d d_e
\right).
\]

Since experts are selected inside the same layer position, EP does not remove
the dense Transformer critical path. The dense parameter-gradient reduction
also spans all \(B(S+T)\) token instances. Thus,
\[
D_{\mathrm{EP}}
=
2L
\log\!\left(
\frac{d^4d_e(S+T)}{h}
\right)
+
\log(B(S+T)),
\]
or asymptotically,
\[
D_{\mathrm{EP}}
=
O\!\left(
L[\log d+\log(S+T)]
+\log(B(S+T))
\right).
\]

The per-rank memory is
\[
M_{\mathrm{EP,rank}}
=
L\left(
16d^2+\frac{8Edd_e}{P_e}
\right)
+
4Vd
+
B(S+T)Ld.
\]
Globally, the dense model state and dense activations are replicated across the
\(P_e\) ranks, while expert parameters sum across expert partitions:
\[
M_{\mathrm{EP,global}}
=
P_e
\left(
16Ld^2+4Vd+B(S+T)Ld
\right)
+
8LEdd_e,
\]
up to the same optimizer-state convention used in the baseline memory model.

\subsubsection{3D Parallelism}

3D parallelism combines data, pipeline, and tensor parallelism with degrees
\((P_d,P_p,P_t)\). It applies the substitutions
\[
B\mapsto\frac{B}{P_d},
\qquad
L\mapsto\frac{L}{P_p},
\qquad
d\text{-sharded GEMM terms}\mapsto\frac{1}{P_t}
\]
to the corresponding baseline terms. Therefore, the per-rank work is
\[
W_{\mathrm{3D,rank}}
=
6\frac{B}{P_d}(S+T)\frac{L}{P_p}
\left(
\frac{4d^2+2dd_{\mathrm{ff}}+(S+T)d}{P_t}
\right).
\]

The per-rank depth combines pipeline and tensor reductions, while the local
parameter-gradient reduction spans the
\(\frac{B}{P_d}(S+T)\) batch-token positions assigned to the data-parallel
replica:
\[
D_{\mathrm{3D,rank}}
=
2\frac{L}{P_p}
\log\!\left(
\frac{d^4d_{\mathrm{ff}}(S+T)}{P_t^2h}
\right)
+
\log\!\left(
\frac{B}{P_d}(S+T)
\right),
\]
or asymptotically,
\begin{align*}
D_{\mathrm{3D,rank}}
&=
O\!\Biggl(
\frac{L}{P_p}
[\log(d/P_t)+\log(S+T)]\\
&\quad+
\log\!\left(
\frac{B}{P_d}(S+T)
\right)
\Biggr).
\end{align*}

The memory expression combines layer partitioning, tensor partitioning, and
batch partitioning:
\[
M_{\mathrm{3D,rank}}
=
\frac{L(16d^2+8dd_{\mathrm{ff}})}{P_pP_t}
+
\frac{4Vd}{P_t}
+
\frac{B(S+T)Ld}{P_dP_pP_t}.
\]

Globally, the dominant work remains the baseline work, while model-state memory
is replicated across data-parallel groups:
\[
W_{\mathrm{3D,global}}
=
W_{\mathrm{base}},
\]
\[
M_{\mathrm{3D,global}}
=
P_d
\left[
L(16d^2+8dd_{\mathrm{ff}})+4Vd
\right]
+
B(S+T)Ld.
\]
The global depth restores the full \(L\)-layer pipeline path and the full
batch-gradient dependency, while retaining the tensor-parallel hidden reduction:
\[
D_{\mathrm{3D,global}}
=
2L
\log\!\left(
\frac{d^4d_{\mathrm{ff}}(S+T)}{P_t^2h}
\right)
+
\log(B(S+T)),
\]
or
\begin{align*}
D_{\mathrm{3D,global}}
&=
O\!\bigl(
L[\log(d/P_t)+\log(S+T)]\\
&\quad+\log(B(S+T))
\bigr).
\end{align*}

This explains the main role of 3D parallelism in the tables: it gives strong
per-device memory reduction because it simultaneously partitions batch, layers,
and hidden-dimensional computation. Its realized runtime, however, depends on
communication, pipeline scheduling, and microbatching, which are deliberately
outside the work--depth--memory abstraction used here.

\subsubsection{5D Parallelism}

We finally consider a combined configuration using data, pipeline, tensor,
context, and expert parallelism with degrees
\((P_d,P_p,P_t,P_c,P_e)\), where
\[
P_dP_pP_tP_cP_e=N.
\]
An important distinction from the isolated EP analysis above is required.
The isolated EP row is intentionally a conceptual abstraction designed to
expose the effect of expert partitioning alone: only expert FFNs are partitioned
across the \(P_e\) ranks, while the same shared dense Transformer computation
is executed on the same token set by every EP rank. Consequently, its global
work contains \(P_e\) copies of the shared dense arithmetic. This should not be
interpreted as the execution strategy necessarily used in a practical
multidimensional MoE system.

For the 5D configuration, we instead model a practically motivated coupled
execution in which the ranks participating in the EP dimension also own
disjoint \emph{source-token} shards for the shared Transformer path. Expert
parallelism itself determines where expert parameters reside; the additional
source-token partition specifies where tokens reside before expert dispatch.
These are distinct notions. Formally, if
\(\mathcal U\) denotes the \(B(S+T)\) batch-token positions of the global
training invocation, then we assume
\[
\mathcal U
=
\bigsqcup_{r_d=1}^{P_d}
\bigsqcup_{r_c=1}^{P_c}
\bigsqcup_{r_e=1}^{P_e}
\mathcal U_{r_d,r_c,r_e},
\]
with approximately balanced partitions
\[
|\mathcal U_{r_d,r_c,r_e}|
\approx
\frac{B(S+T)}{P_dP_cP_e}.
\]
Thus, a shared dense parameter may be replicated across the EP dimension, but
each source token is processed by the shared path on only one EP rank. After
routing, token representations are redistributed by the expert-parallel
all-to-all to the ranks hosting the selected experts, and expert outputs are
subsequently returned to the corresponding source-token ranks. We assume
balanced routing, so the resulting token--expert assignments are approximately
uniformly distributed across the \(P_e\) expert ranks. Tensor parallelism is
applied to both shared dense and expert GEMMs.

Under these assumptions, the per-rank work is
\[
W_{\mathrm{5D,rank}}
=
6\frac{B}{P_dP_e}
\frac{S+T}{P_c}
\frac{L}{P_p}
\left(
\frac{
4d^2+2E_a d d_e+(S+T)d
}{P_t}
\right).
\]
The factors have distinct origins: \(P_d\) partitions the batch, \(P_c\)
partitions source sequence positions, \(P_e\) further partitions source-token
ownership across the expert-parallel ranks, \(P_p\) partitions layers, and
\(P_t\) partitions intra-layer matrix operations. The \(1/P_e\) factor is
therefore applied to the source-token workload rather than separately to the
expert term; under balanced routing, each EP rank already receives
approximately \(1/P_e\) of the global token--expert assignments.

Summing executed arithmetic across all
\(N=P_dP_pP_tP_cP_e\) ranks gives
\[
W_{\mathrm{5D,global}}
=
6B(S+T)L
\left(
4d^2+2E_a d d_e+(S+T)d
\right).
\]
In particular, the shared dense term is not multiplied by \(P_e\), unlike in
the isolated EP abstraction. The reason is that the execution
mapping assigns disjoint source-token subsets to the EP ranks, so the same
dense token computation is not redundantly executed on every EP rank. This
illustrates the distinction between \emph{parameter replication} and
\emph{arithmetic replication}: shared dense parameters may remain replicated
across the EP dimension even though the corresponding FLOPs are evaluated on
disjoint token subsets.

The per-rank depth is
\[
D_{\mathrm{5D,rank}}
=
2\frac{L}{P_p}
\log\!\left(
\frac{d^4d_e(S+T)}{P_t^2h}
\right)
+
\log\!\left(
\frac{B(S+T)}{P_dP_eP_c}
\right).
\]
The first term combines pipeline and tensor partitioning of the layerwise
critical path while retaining the global attention span \(S+T\). The second
term is the local parameter-gradient reduction over the source batch-token
positions assigned to the rank. As in the other tables, communication,
including the expert dispatch/combine all-to-all and distributed-attention
collectives, is outside the arithmetic-depth abstraction.

At the global level, the complete invocation traverses all \(L\) pipeline
layers and its parameter gradients depend on all \(B(S+T)\) batch-token
positions. Hence,
\[
D_{\mathrm{5D,global}}
=
2L
\log\!\left(
\frac{d^4d_e(S+T)}{P_t^2h}
\right)
+
\log(B(S+T)).
\]

The corresponding per-rank memory is
\[
M_{\mathrm{5D,rank}}
=
\frac{16Ld^2}{P_pP_t}
+
\frac{8LEdd_e}{P_pP_tP_e}
+
\frac{4Vd}{P_t}
+
\frac{B(S+T)Ld}{P_dP_eP_cP_pP_t}.
\]
Shared dense model state is partitioned by PP and TP but remains replicated
across the DP, CP, and EP dimensions. Expert model state is additionally
partitioned across \(P_e\), while the leading-order activation term follows the
combined source-token, layer, and tensor partitioning assumed above.
Consequently, aggregate global memory is
\[\begin{aligned}
M_{\mathrm{5D,global}} = 
&P_dP_cP_e \left(16Ld^2+4Vd \right) \\ 
&+ 8P_dP_cLEdd_e + B(S+T)Ld,
\end{aligned}\]
up to the same boundary-stage and embedding-placement conventions used for PP
and TP.

Isolated EP provides a
clean limiting case in which only expert FFNs are partitioned, making the cost
of replicated shared computation explicit. The 5D configuration instead
captures a more practically relevant multidimensional MoE execution in which
expert placement is combined with source-token distribution and other
intra-model parallelism dimensions. As a result, shared dense arithmetic is
partitioned rather than replicated across the EP dimension, whereas replicated
shared \emph{state} can still contribute a factor \(P_e\) to global memory.

\subsection{Results for Inter-Model Parallelism}
\label{sec:app:derivations-inter}

We provide derivations for results in Section~\ref{sec:analysis-inter} and in Table~\ref{tab:inter_model_parallelism}.
The analysis uses the building blocks from Table~\ref{tab:building-blocks}: forward
costs \(C_f(\cdot)\), backward costs \(C_b(\cdot)\), autoregressive generation
cost \(C_{\mathrm{gen}}(\cdot)\), forward/backward depths \(D_f(\cdot)\),
\(D_b(\cdot)\), and generation depth \(D_{\mathrm{gen}}(\cdot)\). Communication,
kernel efficiency, and scheduling overheads are outside the work--depth
abstraction; they are represented only indirectly through memory buffers such as
\(M_{\mathrm{reshard}}\) and shadow copies \(M_{\mathrm{sh}}\).

\subsubsection{Stage Primitives}

For a PPO-style iteration with \(B\) prompts, \(K\) rollouts per prompt, prompt
length \(S\), and response length \(T\), Generation invokes the actor
autoregressively. Using the prefill--decode decomposition from
Table~\ref{tab:building-blocks},
\begin{align*}
W_G &= BK\,C_{\mathrm{gen}}(\pi_\theta),\\
D_G &= D_{\mathrm{gen}}(\pi_\theta),\\
M_G &= |\pi_\theta|+BK\,M_{\mathrm{KV}}.
\end{align*}
This assumes the rollout-time actor log-probabilities are stored during
generation. If an implementation recomputes old actor log-probabilities in a
teacher-forced pass, an additional \(BK(S+T)C_f(\pi_\theta)\) work term and
\(D_f(\pi_\theta)\) depth term should be added to Assessment or Training,
depending on where the recomputation is performed.

Assessment consists of teacher-forced forward passes through
\(\pi_{\mathrm{ref}}\), \(R_\phi\), and \(V_\psi\):
\[
W_A
=
BK(S+T)[C_f(\pi_{\mathrm{ref}})+C_f(R_\phi)+C_f(V_\psi)].
\]
If these models are co-located and executed sequentially, the depth is
\[
D_A^\Sigma
=
D_f(\pi_{\mathrm{ref}})+D_f(R_\phi)+D_f(V_\psi).
\]
If they are placed on disjoint device groups and executed concurrently, the
depth becomes
\[
D_A^{\max}
=
\max\{D_f(\pi_{\mathrm{ref}}),D_f(R_\phi),D_f(V_\psi)\}.
\]
The corresponding co-located memory footprint is
\[
M_A^\Sigma
=
|\pi_{\mathrm{ref}}|+|R_\phi|+|V_\psi|+BK(S+T)M_{\mathrm{Inf}}.
\]

Training updates the actor and critic. For a trainable model \(M\), we define
\begin{align*}
C_{\mathrm{tr}}(M)
&:=
C_f(M)+C_b(M)
\approx
3C_f(M),
\\
D_{\mathrm{tr}}(M)
&:=
D_f(M)+D_b(M),
\end{align*}
since a parameter update requires a forward pass followed by backpropagation.
Thus,
\[
W_T
=
BK(S+T)[C_{\mathrm{tr}}(\pi_\theta)+C_{\mathrm{tr}}(V_\psi)].
\]
Sequential co-located training has depth
\[
D_T^\Sigma
=
D_{\mathrm{tr}}(\pi_\theta)+D_{\mathrm{tr}}(V_\psi),
\]
whereas disaggregated actor/critic training has depth
\[
D_T^{\max}
=
\max\{D_{\mathrm{tr}}(\pi_\theta),D_{\mathrm{tr}}(V_\psi)\}.
\]
Under the Adam memory model without activation checkpointing, the co-located
training footprint is
\[
M_T^\Sigma
=
4(|\pi_\theta|+|V_\psi|)
+
BK(S+T)(L_\pi d_\pi+L_Vd_V).
\]
For disaggregated training, the peak per device group is
\begin{align*}
M_T^{\max}
&=
\max\{4|\pi_\theta|+BK(S+T)L_\pi d_\pi,\,\\
&\quad\quad4|V_\psi|+BK(S+T)L_Vd_V\}.
\end{align*}

\subsubsection{Baseline Co-Located Execution}

The baseline uses separate actor and critic models and a fully co-located
device group. All stages execute sequentially. Therefore,
\[
W_{\mathrm{base}}=W_G+W_A+W_T,
\]
\[
D_{\mathrm{base}}=D_G+D_A^\Sigma+D_T^\Sigma,
\]
and
\[
M_{\mathrm{base}}
=
\max\{M_G,M_A^\Sigma,M_T^\Sigma\}.
\]
The maximum appears because the stages are temporally multiplexed on the same
device group: the peak is the largest stage footprint, not the sum over all
stages. If an implementation keeps all models resident simultaneously, then the
persistent part of the memory expression should instead sum the resident model
states.

\subsubsection{Shared Actor--Critic}

With a shared actor--critic model \(\pi_{\mathrm{ac}}\), the actor and value
heads share a Transformer backbone. Generation uses \(\pi_{\mathrm{ac}}\) as the
policy:
\[
W_G^{\mathrm{ac}}=BK\,C_{\mathrm{gen}}(\pi_{\mathrm{ac}}),
\qquad
D_G^{\mathrm{ac}}=D_{\mathrm{gen}}(\pi_{\mathrm{ac}}).
\]
Assessment no longer requires an independent critic model; instead, value
estimates are produced by the shared model:
\[
W_A^{\mathrm{ac}}
=
BK(S+T)[C_f(\pi_{\mathrm{ref}})+C_f(R_\phi)+C_f(\pi_{\mathrm{ac}})].
\]
Training performs one forward pass followed by backpropagation through the
shared backbone:
\[
W_T^{\mathrm{ac}}
=
BK(S+T)C_{\mathrm{tr}}(\pi_{\mathrm{ac}}).
\]
Thus, shared actor--critic is the only inter-model configuration in
Table~\ref{tab:inter_model_parallelism} that reduces global work relative to the
separate-backbone PPO baseline. Its training memory becomes
\[
M_T^{\mathrm{ac}}
=
4|\pi_{\mathrm{ac}}|+BK(S+T)L_{\mathrm{ac}}d_{\mathrm{ac}},
\]
up to the small policy/value head activations. The benefit is largest when
\(|V_\psi|\) is comparable to \(|\pi_\theta|\); it is smaller when the critic is
already much smaller than the actor.

\subsubsection{Disaggregated Placement}

Disaggregated placement assigns actor, reference, reward, and critic operators to
separate device groups. It does not change total work:
\[
W_{\mathrm{disagg}}=W_G+W_A+W_T.
\]
It changes depth by replacing independent sequential subcomputations with
parallel fork--join regions:
\[
D_{\mathrm{disagg}}
=
D_G+D_A^{\max}+D_T^{\max}.
\]
It also changes memory from co-resident memory to per-group memory:
\[
M_{\mathrm{disagg}}
=
\max\{M_G,M_{\pi_{\mathrm{ref}}},M_{R_\phi},M_{V_\psi},M_{T,\pi},M_{T,V}\}.
\]
This captures the main memory advantage of disaggregation: reward-model devices
need not hold actor or critic state, and actor-training devices need not hold
frozen reward/reference models. Disaggregation alone does not remove the
Generation \(\rightarrow\) Assessment \(\rightarrow\) Training dependency; it
only enables concurrency where the dataflow graph has independent branches.

\subsubsection{Hybrid Execution and Resharding}

Hybrid execution allows the same model to use different intra-model layouts in
different stages. Let \(D_G^{\mathrm{inf}}\) and \(M_G^{\mathrm{inf}}\) denote
Generation depth and memory under an inference-oriented layout, and let
\(D_T^{\mathrm{train}}\) and \(M_T^{\mathrm{train}}\) denote Training depth and
memory under a training-oriented layout such as ZeRO/FSDP or 3D parallelism.
The FLOP work is unchanged:
\[
W_{\mathrm{hyb}}=W_G+W_A+W_T,
\]
but the stage depths become layout-specific:
\[
D_{\mathrm{hyb}}
=
D_G^{\mathrm{inf}}+D_A^{\mathrm{inf},\Sigma}+D_T^{\mathrm{train}},
\]
assuming the same device group is temporally multiplexed and therefore provides
no inter-model concurrency by itself.

The cost of hybrid execution is resharding. If two invocations of the same model
use incompatible layouts, weights or optimizer state must be redistributed. This
does not change FLOP work in the work--depth table, but it affects realized
runtime and may require an additional transient buffer:
\[
M_{\mathrm{hyb}}
=
\max\{M_G^{\mathrm{inf}},M_A^{\mathrm{inf}},M_T^{\mathrm{train}}\}
+
M_{\mathrm{reshard}}.
\]
Hybrid execution is useful only when the per-stage gains from specialized
layouts exceed the resharding and orchestration costs.

\subsubsection{Stage Fusion}

Stage fusion preserves total work:
\[
W_{\mathrm{fusion}}=W_G+W_A+W_T.
\]
Its effect is on depth. Inter-stage fusion streams completed generations into
Assessment, so the fused Generation--Assessment depth is
\[
D_{G/A}^{\mathrm{fused}}
=
\max\{D_G,D_A^{\max}\}
\]
under ideal overlap. Training remains downstream, so
\[
D_{\mathrm{fusion}}
=
\max\{D_G,D_A^{\max}\}+D_T^{\max}.
\]
When generation dominates assessment, this simplifies to
\[
D_{\mathrm{fusion}}\approx D_G+D_T^{\max}.
\]
Intra-stage fusion similarly replaces sequential actor/critic training pipelines
with a max term when the two pipelines can be overlapped:
\[
D_T^\Sigma \rightarrow D_T^{\max}.
\]
The memory footprint may increase because multiple stage fragments are live
simultaneously:
\[
M_{\mathrm{fusion}}
=
\max\{M_{G/A}^{\mathrm{fused}},M_T^{\max}\}.
\]
Here \(M_{G/A}^{\mathrm{fused}}\) includes the live generation KV cache, the
assessment model states, and any queues or communication buffers needed to
stream completed samples.

\subsubsection{Asynchronous Execution}

Bounded asynchrony overlaps generation/assessment of one iteration with training
of another. Total work is unchanged:
\[
W_{\mathrm{async}}=W_G+W_A+W_T.
\]
The steady-state recurrence depth becomes
\[
D_{\mathrm{async}}
=
\max\{D_G+D_A^{\max},D_T^{\max}\},
\]
assuming disaggregated generation/assessment and training groups. If assessment
is also fused with generation, then the first term can be replaced by
\(\max\{D_G,D_A^{\max}\}\).

Asynchrony requires parameter snapshots. If a trainable model is read by a
stale consumer on a disjoint device group, the consumer needs a stable
inference-side copy while the training side updates another copy. We denote this
additional persistent copy by
\[
M_{\mathrm{sh}}(m).
\]
For actor asynchrony, \(M_{\mathrm{sh}}(\pi_\theta)\) is typically a BF16,
possibly quantized, inference copy of the actor. Thus,
\[
M_{\mathrm{async}}
=
\max\{M_{G/A}^{\mathrm{async}}+M_{\mathrm{sh}}(\pi_\theta),M_T^{\max}\}.
\]
This exposes the memory trade-off: asynchrony reduces depth by overlapping
iterations, but it may require roughly one extra parameter copy for each
asynchronously consumed trainable model. This is not necessarily a full \(2\times\)
increase in total memory, because optimizer states remain on the training side
and the shadow copy may be lower precision or sharded, but it must be accounted
for.

\subsubsection{Combined Configuration}

The combined configuration composes the strongest ingredients: stage-specific
intra-model sharding, disaggregated model placement, stage fusion, and bounded
asynchrony. The total FLOP work is still
\[
W_{\mathrm{comb}}=W_G+W_A+W_T,
\]
unless the model structure is also changed, for example by using a shared
actor--critic backbone. With ideal fusion and bounded asynchrony, the recurrence
depth is
\[
D_{\mathrm{comb}}
=
\max\{D_G^{\mathrm{TP}},D_T^{\mathrm{shard}}\},
\]
where \(D_G^{\mathrm{TP}}\) denotes actor generation depth under an
inference-oriented sharded layout, and \(D_T^{\mathrm{shard}}\) denotes the
depth of the sharded training layout. A more conservative expression keeps the
unhidden assessment tail:
\[
D_{\mathrm{comb}}
=
\max\{\max(D_G^{\mathrm{TP}},D_A^{\max}),D_T^{\mathrm{shard}}\}.
\]
The memory expression is
\[
M_{\mathrm{comb}}
=
\max\{M_G^{\mathrm{TP}}+M_{\mathrm{sh}}(\pi_\theta),
M_{\pi_{\mathrm{ref}}},
M_{R_\phi},
M_{V_\psi},
M_T^{\mathrm{shard}}\}.
\]
Thus, the combined strategy gives the smallest idealized depth and peak
per-device-group memory in Table~\ref{tab:inter_model_parallelism}, but only
under sufficient hardware, careful scheduling, and acceptable staleness.